\PassOptionsToPackage{table}{xcolor}
\documentclass{article}
\usepackage{hyperref}

\usepackage[preprint]{icml2026}

\usepackage{amsmath,amsfonts,bm}

\def\eqref#1{equation~\ref{#1}}

\def\1{\bm{1}}

\DeclareMathAlphabet{\mathsfit}{\encodingdefault}{\sfdefault}{m}{sl}
\SetMathAlphabet{\mathsfit}{bold}{\encodingdefault}{\sfdefault}{bx}{n}

\usepackage{url}
\usepackage{graphicx}
\usepackage{booktabs}
\usepackage{longtable}
\usepackage{adjustbox}
\usepackage[most]{tcolorbox}
\usepackage{fvextra}
\usepackage{multirow}
\usepackage{amsmath,amssymb}
\usepackage{xcolor}
\usepackage{subcaption}
\usepackage{tikz}
\usepackage{pdflscape}
\usepackage{soul}
\usepackage{etoc}
\usepackage{float}
\usepackage{array}
\usepackage{tabularx}
\usepackage[normalem]{ulem}
\usepackage{enumitem}
\usepackage{xspace}
\usetikzlibrary{positioning,arrows.meta,fit,backgrounds,calc}

\definecolor{tabkey}{HTML}{EFEBF5}
\definecolor{tabgroup}{HTML}{EFEBF5}

\definecolor{ours}{HTML}{1F4E8C}
\definecolor{twin}{HTML}{C4622D}
\definecolor{pangram}{HTML}{8A8A8A}
\definecolor{muted}{HTML}{6B6B6B}
\definecolor{rule}{HTML}{CCCCCC}
\definecolor{cellgrey}{HTML}{EDEDED}
\definecolor{target}{HTML}{B03A2E}
\definecolor{gap}{HTML}{7B3F8C}
\definecolor{base}{HTML}{2E7D4F}
\definecolor{ok}{HTML}{2E7D4F}
\definecolor{ink}{HTML}{222222}

\definecolor{inkFable}{HTML}{B4501C}   %
\definecolor{inkOpus}{HTML}{C85A82}
\definecolor{inkCarbon}{HTML}{32122F}
\definecolor{inkArgon}{HTML}{6E4B78}
\definecolor{inkSol}{HTML}{437E8E}
\definecolor{inkAstra}{HTML}{0F3640}
\newlength{\labiconht}
\newcommand{\labicon}[1]{\begingroup\setlength{\labiconht}{1.15\fontcharht\font`H}%
  \raisebox{\dimexpr(\fontcharht\font`H-\labiconht)/2\relax}{\includegraphics[height=\labiconht]{figures/assets/rq2-lab-icons/mono/#1}}\endgroup}
\newcommand{\labiconc}[2]{\begingroup\setlength{\labiconht}{1.15\fontcharht\font`H}%
  \raisebox{\dimexpr(\fontcharht\font`H-\labiconht)/2\relax}{\includegraphics[height=\labiconht]{figures/assets/rq2-lab-icons/color/#2-#1}}\endgroup}
\newcommand{\iconFable}{\labiconc{B4501C}{anthropic}}
\newcommand{\iconOpus}{\labiconc{C85A82}{anthropic}}
\newcommand{\iconCarbon}{\labiconc{32122F}{google}}
\newcommand{\iconArgon}{\labiconc{6E4B78}{google}}
\newcommand{\iconSol}{\labiconc{437E8E}{openai}}
\newcommand{\iconAstra}{\labiconc{0F3640}{openai}}

\usepackage[LGR,TS1,T1,OT1]{fontenc}
\usepackage{etoolbox}
\DeclareFontFamily{T1}{InterTab}{}
\DeclareFontShape{T1}{InterTab}{m}{n}{<-> s*[0.9] InterTab-Regular-t1}{}
\DeclareFontShape{T1}{InterTab}{m}{it}{<-> s*[0.9] InterTab-Italic-t1}{}
\DeclareFontShape{T1}{InterTab}{m}{sl}{<-> ssub * InterTab/m/it}{}
\DeclareFontShape{T1}{InterTab}{sb}{n}{<-> s*[0.9] InterTab-SemiBold-t1}{}
\DeclareFontShape{T1}{InterTab}{sb}{it}{<-> s*[0.9] InterTab-SemiBoldItalic-t1}{}
\DeclareFontShape{T1}{InterTab}{sb}{sl}{<-> ssub * InterTab/sb/it}{}
\DeclareFontShape{T1}{InterTab}{b}{n}{<-> ssub * InterTab/sb/n}{}
\DeclareFontShape{T1}{InterTab}{b}{it}{<-> ssub * InterTab/sb/it}{}
\DeclareFontShape{T1}{InterTab}{bx}{n}{<-> ssub * InterTab/sb/n}{}
\DeclareFontShape{T1}{InterTab}{bx}{it}{<-> ssub * InterTab/sb/it}{}
\newcommand{\tablefont}{\fontencoding{T1}\fontfamily{InterTab}\renewcommand{\bfdefault}{sb}\selectfont}
\DeclareFontFamily{TS1}{InterTab}{}
\DeclareFontShape{TS1}{InterTab}{m}{n}{<-> s*[0.9] Inter-Regular-lf-ts1}{}
\DeclareFontShape{TS1}{InterTab}{sb}{n}{<-> s*[0.9] Inter-SemiBold-lf-ts1}{}
\DeclareFontShape{TS1}{InterTab}{m}{it}{<-> ssub * InterTab/m/n}{}
\DeclareFontShape{TS1}{InterTab}{sb}{it}{<-> ssub * InterTab/sb/n}{}
\DeclareFontFamily{LGR}{InterTab}{}
\DeclareFontShape{LGR}{InterTab}{m}{n}{<-> s*[0.9] Inter-Regular-lf-lgr}{}
\DeclareFontShape{LGR}{InterTab}{sb}{n}{<-> s*[0.9] Inter-SemiBold-lf-lgr}{}
\newcommand{\tabminus}{\UseTextSymbol{TS1}{\textminus}}
\newcommand{\tabdagger}{\UseTextSymbol{TS1}{\textdagger}}

\newcommand{\tabDelta}{{\fontencoding{LGR}\selectfont D}}
\newif\ifinfloat
\AtBeginEnvironment{table}{\infloattrue}
\AtBeginEnvironment{table*}{\infloattrue}
\AtBeginEnvironment{figure}{\infloattrue}
\AtBeginEnvironment{figure*}{\infloattrue}
\AtBeginEnvironment{tabular}{\ifinfloat\tablefont\fi}
\AtBeginEnvironment{tabular*}{\ifinfloat\tablefont\fi}
\AtBeginEnvironment{tabularx}{\ifinfloat\tablefont\fi}
\AtBeginEnvironment{longtable}{\tablefont}
\newcommand{\jFable}[1]{\iconFable\,\textcolor{inkFable}{#1}}
\newcommand{\jOpus}[1]{\iconOpus\,\textcolor{inkOpus}{#1}}
\newcommand{\jSol}[1]{\iconSol\,\textcolor{inkSol}{#1}}
\newcommand{\jArgon}[1]{\iconArgon\,\textcolor{inkArgon}{#1}}
\newcommand{\jCarbon}[1]{\iconCarbon\,\textcolor{inkCarbon}{#1}}

\newcommand{\jDeepSeek}[1]{\labicon{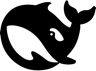}\,#1}

\newcommand{\repourl}{https://github.com/denizbt/unmasking_authors}
\newcommand{\repolink}[1]{\href{\repourl}{#1}}
\definecolor{repoicon}{HTML}{461E47}
\newif\ifthreelinks
\threelinkstrue
\newcommand{\blogurl}{https://atreydesai.com/blog/who-wrote-this/}
\newcommand{\profilesurl}{https://model-profiles.github.io/}
\newcommand{\abslink}[3]{\mbox{\textcolor{repoicon}{#1}~\href{#2}{\textbf{#3}}}}
\newcommand{\abstractlinks}{{\fontsize{8.5}{10.2}\selectfont
  \abslink{\faGithub}{\repourl}{Code}\hspace{1.4em}%
  \abslink{\faGlobe}{\blogurl}{Blog}%
  \ifthreelinks\hspace{1.4em}\abslink{\faIdCard}{\profilesurl}{Model profiles}\fi}}
\usepackage{placeins}
\usepackage{xurl}
\usepackage{longtable}
\usepackage{fontawesome5}
\hypersetup{pdftitle={Frontier models need only one text to unmask its human or AI author}}
\hypersetup{pdfborder={0 0 0}}

\icmltitlerunning{Frontier models need only one text to unmask its human or AI author}

\begin{document}

\twocolumn[
  \icmltitle{Frontier models need only one text to unmask its human or AI author}

  \icmlsetsymbol{equal}{*}

  \begin{icmlauthorlist}\fontsize{9.5}{12}\selectfont
    \icmlauthor{Atrey Desai}{equal,umd}
    \icmlauthor{Deniz B\"{o}l\"{o}ni-Turgut}{equal,umd}
    \icmlauthor{Ty Brassington}{equal,sfu}
    \icmlauthor{Minjoon Choi}{equal,umd}
    \icmlauthor{Jenna Russell}{equal,umd}
    \icmlauthor{Rishanth Rajendhran}{equal,umd}\\
    \icmlauthor{Bradley Emi}{pangram}
    \icmlauthor{Hila Gonen}{ubc}
    \icmlauthor{Hal Daum\'{e} III}{umd}
    \icmlauthor{John Wieting}{gdm}
    \icmlauthor{Marzena Karpinska}{sfu}
    \icmlauthor{Mohit Iyyer}{umd}
  \end{icmlauthorlist}

  \icmlaffiliation{umd}{University of Maryland}
  \icmlaffiliation{sfu}{Simon Fraser University}
  \icmlaffiliation{pangram}{Pangram Labs}
  \icmlaffiliation{ubc}{University of British Columbia and Canada CIFAR AI Chair, Amii}
  \icmlaffiliation{gdm}{Google DeepMind}

  \icmlcorrespondingauthor{Deniz B\"{o}l\"{o}ni-Turgut}{denizbt@umd.edu}
  \icmlcorrespondingauthor{Mohit Iyyer}{miyyer@umd.edu}

  \icmlkeywords{authorship attribution, AI-generated text detection, large language models, explanations}

  \vskip 0.3in
]
\printAffiliationsAndNotice{\icmlEqualContribution}

\etocdepthtag.toc{mainpaper}
\begin{abstract}
Given a single text and a simple zero-shot prompt (no web access, tools, candidate lists, or writing samples), frontier models can answer \textit{``who wrote this?''} at three increasingly specific levels: (1) whether an AI or human wrote it, (2) if an AI, which lab, and (3) if a human, often exactly who. The latest models (Claude Fable 5.2, GPT-6.1 Sol, Gemini 4 Carbon) match trained detectors like Pangram while also explaining their predictions in natural language, and Opus 5.5 names the correct lab for 98\% of texts generated by models from Anthropic, OpenAI, and Google. Most strikingly, Fable 5.2 correctly names the author of 55\% of 15K anonymous ICML 2020 reviews that were never made public, and identifies the username behind 69\% of comments from active Redditors (with $>$500 posts over the last year). Accuracy depends on exposure to an author's writing in pretraining data and on reasoning effort, and simply asking a model how familiar it is with someone predicts whether it can identify their writing. For anyone who has published enough writing online, frontier models present a greater threat to anonymity than commonly assumed, and defenses are limited.
\par\vspace{1pt}
{\centering\abstractlinks\par}
\par\kern-11pt

\end{abstract}

\section{Introduction}
\label{sec:intro}
\definecolor{introhead}{HTML}{6E2042}

\begin{figure*}[!t]
\centering
\includegraphics{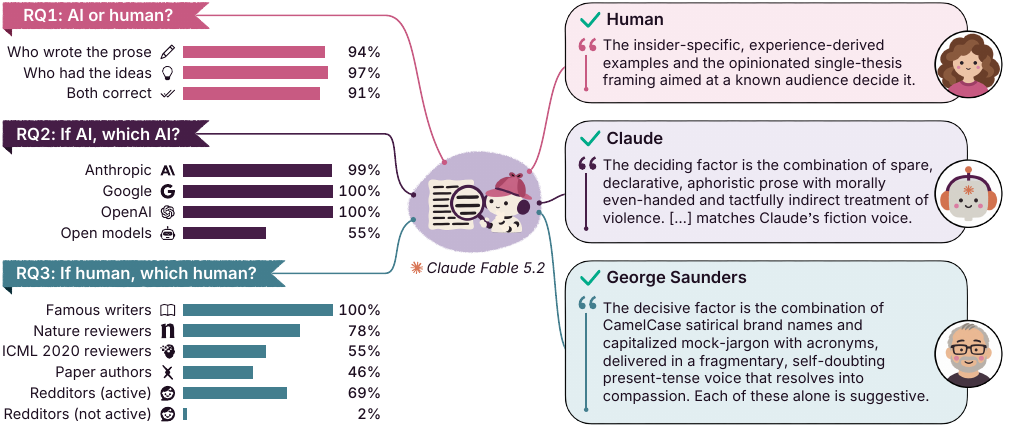}
\caption{Given a text sample with no candidates or external aids, Claude Fable 5.2 can readily identify AI authorship, source labs, and even human writers. Frontier models still face challenges in identifying open-weight models and less-known authors. \emph{Open models}: percent of texts by open-weight writers that Fable~5.2 attributes to any open-weight model; naming the exact lab is harder (\autoref{tab:ai_vs_ai_levels}).}
\label{fig:main-results-overview}
\end{figure*}

In May 2026, the eventual winner of the Commonwealth Short Story Prize was accused of using AI to write their story. Granta, the literary magazine that published the story, was widely mocked for the naivety of its investigation. It simply asked Claude whether AI had written the story.\footnote{For the reaction, see \href{https://x.com/krishnanrohit/status/2056804945698103581}{this post on X}. 
The author was cleared of misconduct, and the story went on to win the overall prize.} In this paper, we show that Granta was not naive, but just a few months early. Frontier models now identify AI text as well as specialized detectors like Pangram, and they go much further. Given a single text, with no web search or list of candidates, they can often name the lab behind an AI-generated text, or even \emph{the specific person} who wrote a human one (\autoref{fig:main-results-overview}).

\textbf{\textcolor{introhead}{Human or AI?}} \textbf{Frontier models can detect AI prose and ideas.} Using a simple zero-shot prompt, we show that frontier models are competitive with state-of-the-art detectors of AI text (Pangram~4.1; \citealp{glickenhaus2026pangram4technicalreport}) and AI ideas (IdeaLens; \citealp{rajendhran2026idealensdetectingaiideas}). On 36K documents from 24 benchmark datasets, Gemini 4 Carbon detects 97\% of fully AI-generated texts vs. 97\% for Pangram 4.1, at the same false positive rate (0.01\% of English human texts), while GPT-6.1 Sol reaches 90\% accuracy on idea detection compared to 87\% for IdeaLens. Unlike most detectors, frontier models can also explain their predictions in natural language, and we find that these explanations agree with features of AI text and ideas discovered by prior work. 

\textbf{\textcolor{introhead}{If AI, which AI?}} \textbf{Frontier models can name the lab behind an AI-generated text.} We create a dataset of short stories, news articles, and op-eds generated by 34 models from 11 labs, and prompt frontier models (with a minimally modified version of the same zero-shot prompt) to name the lab and model behind each text. Despite recent studies reporting increasing homogeneity in AI writing~\cite{jiang2026artificial, russell2026storyscope, kim2026argumentcollapsellmsflatten, gill2026llms}, Opus 5.5 and Gemini 4 Argon both name the correct lab for 98\% of texts generated by Anthropic, OpenAI, and Google. They are much less accurate on texts produced by the other eight labs (at most 35\%), perhaps due to distillation~\citep{anthropic2026distillation, openai2026deepseek}. Moonshot's Kimi K3 is usually mistaken for Claude, and Alibaba's Qwen3.8 models for GPT. The explanations reveal how frontier models view each lab's style (e.g., associating Gemini with clichés and invented quotes).
 
\textbf{\textcolor{introhead}{If human, which human?}} \textbf{Frontier models can name the person who wrote a text.} Identifying the author of a human-written text is a far more open-ended task than detecting which lab wrote an AI text. Nevertheless, given just a text and the same prompt, Claude Fable 5.2 names the reviewer of 55\% of 15K anonymous ICML 2020 reviews whose content and authorship were never made public (studied with IRB approval).\footnote{This study was reviewed by the University of Maryland Institutional Review Board (IRB \#4656-1).} On texts published after its training cutoff, it also names the reviewer of 78\% of \emph{Nature} Portfolio journal reviews, the writer of 15 stories by authors with $>$1M Goodreads ratings, a co-author of 46\% of arXiv papers, and the username behind 69\% of comments made by active Redditors. A model's stated confidence is informative too. When Fable 5.2 reports 25\% or higher confidence, it predicts the right author 87\% of the time. Unlike recent work on deanonymization with LLMs~\citep{lermen2026deanonymization}, none of these results required a pool of candidates, web access, or writing samples---the model relies solely on its parametric knowledge. %

\paragraph{Familiarity with the writer predicts accuracy.}
Accuracy depends on how much of a writer's text a model has likely seen during pretraining \citep{razeghi2022impact, kandpal2023large}. For example, Opus 5.5 identifies only 20\% of stories by authors it rates moderately familiar, %
none of the 91 stories by unknown authors, and none of the Reddit accounts without prior activity in the subreddit. Pretraining corpora for frontier models are inaccessible, so corpus lookups like Infini-gram~\citep{liu2024infinigram} cannot be run on them. However, simply asking a frontier model how well it knows someone (given only their name or username) predicts its accuracy in advance. 
Opus~5.5 names the writer of 92\% of stories and 78\% of public reviews by authors it says it knows well, but $<2$\% of texts from authors it says it does not know at all. Increasing reasoning effort helps most when the model says it is only mildly familiar with a person, as thinking does for recalling rare facts~\citep{calderon2026emptyshelves, gekhman2026thinkingtorecall}. Fable~5.2's accuracy on Redditors it rates very familiar barely changes between low and maximum effort (88\% to 89\%), while on those it rates a little familiar it rises from 8\% to 38\% (\autoref{tab:rq3_reddit_effort_familiarity}).

\paragraph{Implications for anonymity.}
Prompting frontier models is a simple and effective strategy for authorship detection, one that shows no signs of plateauing as more powerful models trained on larger datasets continue to emerge. Thus, maintaining anonymity (e.g., in peer review or online forums) will continue to get harder, especially if our deliberately minimal setup is combined with web access and cross-platform matching~\citep{lermen2026deanonymization}. Defenses are limited. Models rarely refuse these queries, deleting one's online writing does not remove it from the weights of existing models, and rewriting human-authored text with AI only helps partially.

\section{Human or AI: who wrote the prose, who had the ideas?}
\label{sec:ai_vs_human}
\begin{figure}[t]
  \centering
  \includegraphics{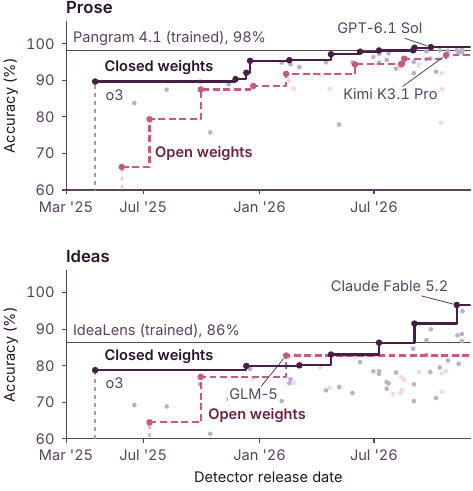}
  \caption{How well each model, by release date, labels who wrote the prose and who had the ideas.
  Accuracy uses an 860-document sample: for prose, the mean over fully human and fully AI documents; for ideas, the mean over all four combinations of human or AI prose and ideas.
  Lines step up at each new best model. Horizontal lines mark the trained detectors. Closed-weight models (solid purple) consistently outperform open-weight models (dashed pink). A complete timeline appears in \autoref{fig:rq1_frontier_full_views}.}
  \label{fig:atrey_rq1_frontier}
\end{figure}

Zero-shot prompting has historically been dismissed as a viable method for detecting AI text, as prior work finds that models have label biases and underperform trained detectors~\citep{bhattacharjee2024fighting, russell-etal-2025-people}. However, an evaluation published by Vals~AI in September 2026 shows that Claude Opus~5.5 and GPT-6 Astra are competitive with commercially available detectors on detecting AI-generated prose~\citep{fein2026vals}. We extend this evaluation by (1) asking whether frontier models can also tell who came up with a text's \emph{ideas}, not just who wrote its prose, and (2) measuring how this capability has improved across 72 models released over the past two years (\autoref{fig:atrey_rq1_frontier}). On 36,427 documents from 24 benchmarks, frontier models are competitive with trained detectors such as Pangram~4.1~\citep{glickenhaus2026pangram4technicalreport} at prose detection, in terms of both accuracy and false positive rate. On idea detection, they exceed the accuracy of IdeaLens, the only existing state-of-the-art idea detector~\citep{rajendhran2026idealensdetectingaiideas}. 
Contamination is an unlikely explanation: though many of these benchmarks were published before the training cutoffs of recently released models, accuracy remains high on texts published after models' training cutoffs and release dates. Unlike Pangram or IdeaLens, frontier models can also justify their predictions with  explanations and quotations, and these explanations are consistent with findings from existing analyses of AI writing.\footnote{Claude Fable 5.2 had the highest accuracy on the 860-document sample (94\% prose and 97\% ideas; \autoref{tab:atrey_rq1_quadrants}), but was not run fully due to cost constraints.}

\subsection{Experimental Setup}
\label{sec:experiments}\label{sec:method_data}
Following IdeaLens~\citep{rajendhran2026idealensdetectingaiideas} and the cognitive process model of writing~\citep{FlowerHayes1981, hayes2012}, we define a document's \emph{ideas} as what its author decided to say (e.g., its claims, events, and organization), as opposed to the words used to say it (its \emph{prose}). A document's prose and ideas can each come from a person or an AI model, giving the following four cases: 
\newcommand{\caseicons}[2]{\makebox[0.6em][c]{\scriptsize\faLightbulb}\makebox[1.3em][c]{#1}\hspace{0.3em}%
  \makebox[0.8em][c]{\scriptsize\faPen}\makebox[1.3em][c]{#2}}
\begin{itemize}[leftmargin=4.8em, labelwidth=4.3em, labelsep=0.5em, topsep=2pt, itemsep=1.5pt, parsep=0pt, partopsep=0pt]
    \item[\caseicons{\faUser}{\faUser}] \textbf{Fully human}: a person came up with the ideas and wrote the prose.
    \item[\caseicons{\faRobot}{\faUser}] \textbf{AI ideas, human prose}: from a model's ideas, a person wrote in their own words.
    \item[\caseicons{\faRobot}{\faRobot}] \textbf{Fully AI}: a model came up with the ideas and wrote the prose.
    \item[\caseicons{\faUser}{\faRobot}] \textbf{Human ideas, AI prose}: a person came up with the ideas, and a model wrote, rewrote, or polished the prose.
\end{itemize}

Each model that we evaluate is given the same zero-shot prompt and a single document, with no tools or reference texts. The prompt asks for two judgments on both ideas and prose, along with accompanying information such as key quotes from the document, reasoning for a decision, decisive factors, and an estimate of the share that AI produced from 0 (all human) to 1 (all AI). The exact prompt is in our \repolink{repository}.

\begin{figure*}[t]
  \centering
  \includegraphics{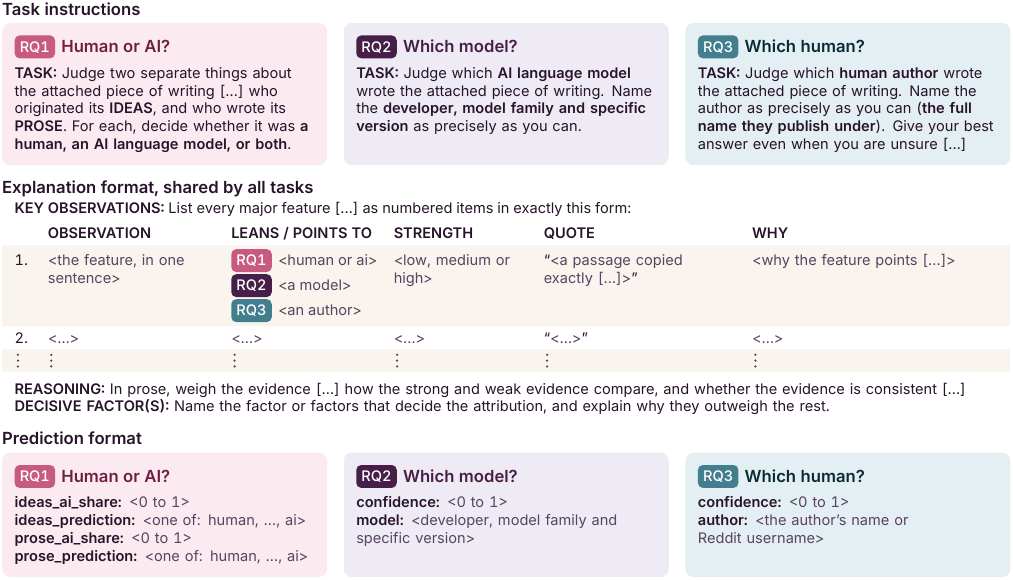}
  \caption{\textbf{Prompt excerpts across the three tasks.}
  For each task, the task instructions, explanation format, and prediction
  format are combined into one prompt, sent to the detector in a single call.
  Task instructions and prediction fields vary by task, while the explanation
  format is shared. The first task (\textit{Human or AI?}) judges ideas and prose
  separately, while other tasks name a single model or human author.
  Ellipses mark omissions and placeholders are abbreviated.
  }
  \label{fig:prompt-overview}
\end{figure*}

\textbf{Models.}
We evaluate 72 models from 13 labs, released between July 2024 and November 2026.\footnote{We especially focus on Claude Fable~5.2 and Opus~5.5~\citep{anthropic2026fable52, anthropic2026opus55}, GPT-6 Astra~\citep{openai2026gpt6astra} and GPT-6.1 Sol~\citep{openai2026gpt61sol}, Gemini~4 Argon and Carbon~\citep{gemini2023, google2026gemini4argon, google2026gemini4carbon}, and DeepSeek~V4.1 Flash~\citep{deepseek2026v41flash}; the full list of models is in \autoref{tab:atrey_detectors}.}
We choose leading closed- and open-weight models released since July 2024 using the Epoch Capabilities Index (ECI) \citep{ho2025rosetta} and Artificial Analysis,\footnote{\url{https://artificialanalysis.ai/}} two standardized aggregated measures of general capabilities.

\textbf{Data.}
We combine 37,575 documents from 24 public datasets (\S\ref{app:rq1_data}) of peer reviews, student essays, news, fiction, Wikipedia and web text, and academic writing. Most pair human documents with fully AI ones, such as MELD~\citep{li2026meldmultitaskequilibratedlearning} and StoryScope~\citep{russell2026storyscope}. Some also have models rewrite or polish human documents, like ARB~\citep{perrone2026arbmatchedauthorshiprewritingbenchmark}, OpAI-Bench~\citep{Bsharat2026OperationGuidedPH} and GEDE~\citep{Gehring2025LimitsOL}. Finally, the TwiceTold dataset~\citep{rajendhran2026idealensdetectingaiideas}  has people write stories in their own words from a model's ideas.
We run a subset of ten models on the full corpus, and evaluate accuracy based on 36,427 documents that all models answered, removing refusals and ambiguous labels (\S\ref{app:atrey_rq1_full}).
To measure capabilities over time, we run all models on an 860-document sample. \textbf{Since public datasets may appear in training data, we also evaluate on 200 stories and 1,000 news articles published after models' knowledge cutoffs (\S\ref{sec:robustness}).}

\textbf{Evaluation metrics and baselines.} Unless otherwise specified, we follow IdeaLens and report \emph{accuracy} macro-averaged over
the four cases, so that each case counts equally regardless of how many documents it contains. An attribute counts as AI when its share is at least 0.5 (shares range from 0 to 1), our default threshold.\footnote{Pangram~4.1 labels text as human, mixed, or AI; however, since it is unclear how mixed text should be treated, we apply the same 0.5 threshold to its AI share, counting AI-assisted text as half. Unweighted micro-averages are described in \S\ref{app:rq1_results}.} Recall is also evaluated at a 1\% false-positive rate (FPR) on English human
text.\footnote{False-positive rates use the 9,687 English fully human documents whose human authorship we can trust (\S\ref{app:rq1_data}).}
We also compare models against three trained detectors: ProseLens \citep{rajendhran2026idealensdetectingaiideas} and Pangram~4.1 \citep{glickenhaus2026pangram4technicalreport} for prose, and IdeaLens \citep{rajendhran2026idealensdetectingaiideas} for ideas.

\begin{table*}[t]
\centering\small
\setlength{\tabcolsep}{4.5pt}
\newcommand{\quadIcon}[1]{\labicon{#1}\,}
\newcommand{\quadNoIcon}{\hphantom{\quadIcon{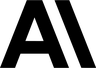}}}
\caption{
Frontier closed-weight models now match trained detectors and significantly lead open-weight LLMs in overall detection abilities. We report accuracy (\%) on the 860-document sample of the full corpus. \textsuperscript{\dag}Pangram~4.1 is not scored on human ideas in AI prose, where its definition of AI-edited text differs from the benchmarks' (\S\ref{sec:ai_vs_human}); its average covers the other three cases.}
\label{tab:atrey_rq1_quadrants}
\begin{tabular}{l rrrr>{\columncolor{tabkey}}r rrrr>{\columncolor{tabkey}}r}
\toprule
& \multicolumn{5}{c}{\textbf{\faPen\ Prose accuracy (\%)}} & \multicolumn{5}{c}{\textbf{\faLightbulb\ Ideas accuracy (\%)}} \\
\cmidrule(lr){2-6}\cmidrule(lr){7-11}
\faPen\ \textit{Prose by} & Human & Human & AI & AI & \multicolumn{1}{r}{} & Human & Human & AI & AI & \multicolumn{1}{r}{} \\
\faLightbulb\ \textit{Ideas by} & Human & AI & AI & Human & \multicolumn{1}{r}{Avg.} & Human & AI & AI & Human & \multicolumn{1}{r}{Avg.} \\
\midrule
\multicolumn{11}{l}{\textbf{Closed-weight LLMs}} \\
\quad \iconFable\,\textcolor{inkFable}{Claude Fable 5.2} & \textbf{100.0} & \textbf{100.0} & 96.5 & \textbf{81.2} & \textbf{94.4} & \textbf{100.0} & \textbf{98.0} & 89.2 & \textbf{99.3} & \textbf{96.6} \\
\quad \iconOpus\,\textcolor{inkOpus}{Claude Opus 5.5} & \textbf{100.0} & \textbf{100.0} & 95.8 & 77.5 & 93.3 & 99.6 & 64.0 & 88.5 & 95.9 & 87.0 \\
\quad \iconAstra\,\textcolor{inkAstra}{GPT-6 Astra} & 99.6 & 98.0 & 98.3 & 80.6 & 94.1 & \textbf{100.0} & 80.0 & 90.2 & 95.9 & 91.5 \\
\quad \iconSol\,\textcolor{inkSol}{GPT-6.1 Sol} & 99.6 & 96.0 & \textbf{98.6} & 79.6 & 93.4 & \textbf{100.0} & 72.0 & 91.3 & 96.6 & 90.0 \\
\quad \iconCarbon\,\textcolor{inkCarbon}{Gemini 4 Carbon} & 99.6 & \textbf{100.0} & 96.5 & 78.6 & 93.7 & 99.6 & 62.0 & \textbf{93.7} & 98.0 & 88.3 \\
\quad \iconArgon\,\textcolor{inkArgon}{Gemini 4 Argon} & 99.6 & \textbf{100.0} & 96.9 & 76.9 & 93.3 & 99.6 & 30.0 & 90.6 & 98.0 & 79.5 \\
\midrule
\multicolumn{11}{l}{\textbf{Open-weight LLMs}} \\
\quad \quadIcon{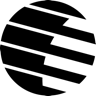}Kimi K3.1 Pro & 98.7 & 76.0 & 95.1 & 61.9 & 82.9 & 99.6 & 40.0 & 85.4 & 89.5 & 78.6 \\
\quad \quadIcon{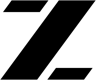}GLM-5.3 & 99.1 & 84.0 & 92.7 & 52.7 & 82.1 & \textbf{100.0} & 38.0 & 89.2 & 77.5 & 76.2 \\
\quad \quadIcon{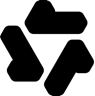}Qwen3.8 27B & 99.6 & 70.0 & 89.5 & 51.4 & 77.6 & \textbf{100.0} & 36.0 & 74.6 & 83.0 & 73.4 \\
\quad \quadIcon{deepseek}DeepSeek V4.1 Flash & 85.6 & 56.0 & 89.5 & 53.4 & 71.1 & 86.5 & 58.0 & 82.2 & 65.0 & 72.9 \\
\midrule
\multicolumn{11}{l}{\textbf{Trained detectors}} \\
\faPen\ Pangram 4.1 & \textbf{100.0} & 96.0 & 96.5 & -- & 97.5\textsuperscript{\dag} & -- & -- & -- & -- & -- \\
\faPen\ ProseLens & \textbf{100.0} & \textbf{100.0} & 93.4 & 48.2 & 85.4 & -- & -- & -- & -- & -- \\
\faLightbulb\ IdeaLens & -- & -- & -- & -- & -- & \textbf{100.0} & 68.0 & 83.8 & 93.4 & 86.3 \\
\bottomrule
\end{tabular}
\end{table*}

\subsection{Results}
\label{sec:results}
On all documents (\S\ref{app:rq1_results}), GPT-6.1 Sol, Opus~5.5, and Gemini~4 Carbon are competitive with trained detectors, achieving 94--95\% accuracy on prose and 89--90\% on ideas, compared to 87\% for IdeaLens on ideas; on prose, over the three cases Pangram~4.1 is scored on, they reach 98--99\% against its 98\%. 

\textbf{Pangram and our benchmarks define AI-edited text differently.} Our human-ideas, AI-prose case counts any human text that has passed through an LLM as AI prose, however light the edit: OpAI-Bench's lightest version changes a median of 2\% of words. In contrast, Pangram only labels text as AI-assisted after some threshold of AI editing has been reached, which is determined by soft n-gram matching \citep[\S3.5]{glickenhaus2026pangram4technicalreport}. Because the two definitions assign these documents different labels, we do not score Pangram on this case; its prose accuracy is the mean of the other three (calibration in \autoref{fig:atrey_rq1_pangram_calibration}). 

\textbf{LLMs have near-zero false-positive rates (FPR).}
In real-world settings---teachers grading essays or books reviewed for prestigious awards---falsely accusing a human writer is far worse than not labeling AI text \citep{liang2023gptdetectors}, making a near-zero FPR essential for any serious detector. Based on 9,687 fully human documents, Pangram~4.1 flags none as AI on prose and Gemini~4 Carbon flags one (0.01\%), and Carbon beats IdeaLens on ideas (0.00\% against 0.21\%) (\autoref{tab:atrey_rq1_operating}); however, the rarity of errors precludes definitive model performance rankings.
FPR does rise on newer text. Carbon flags 0.5\% of 812 post-release news articles that Pangram~4.1 labels human (\autoref{tab:atrey_rq1_contamination}). These may not be true false positives, since authorship cannot be independently verified and some articles may come from newer models with unfamiliar profiles. Dirty data such as pre-2022 web snippets with AI-written summaries appended or machine-translated web spam (\S\ref{app:rq1_data}) and all non-English data are excluded from this set on the suspicion of machine translation involvement.

\textbf{Model accuracy has improved over time.}
In the 860-document sample ({\autoref{tab:rq1_timeline}}), accuracy on both labels climbed from 27\% for GPT-4o in 2024 (55\% on prose, 54\% on ideas) and 59\% for GPT-5.2 in 2025 (75\%, 80\%) to 91\% for present-day frontier models (Claude Fable~5.2 94\%, 97\%). The best open-weight models still trail, with Kimi~K3.1 Pro reaching 63\% (83\%, 79\%). Across the 61 models with an ECI score, accuracy correlates strongly with overall capabilities ($r=0.94$; \S\ref{app:rq1_results}). Newer models are also cheaper than commercial detectors, with Sol costing \$27 per 1,000 documents at standard API prices (\autoref{tab:atrey_detectors}) versus \$376 on Pangram~4.1 at its standard API price (\$0.05 per 100 words).\footnote{The average length for a document in our corpus is 751 words.} We summarize the available experiment cost records in \autoref{tab:project_costs}.

\subsection{Robustness \& Explanations}
\label{sec:explanations}
Because these benchmarks are public, the high LLM accuracy could be due to contamination in pretraining data. We explicitly test this by measuring agreement between LLMs and Pangram on texts published after the release dates of all evaluated models. Then, we examine agreement between explanations generated from different models for the same text, and we also compare the explanation content to prior analyses of AI writing. 

\textbf{Detectors maintain high accuracy on post-release documents.}
\label{sec:robustness}
Most of our benchmarks are public, so the detectors may have seen them during training. On 200 stories published or generated after the detectors' training cutoffs, accuracy is only at most 3.4\% lower than on the full corpus's fiction (\autoref{tab:rq1_fresh_fiction}).\footnote{Both detectors score GPT-6.1 Sol's stories above Claude Opus~5.5's, showing minimal effect of recognizing their own writing.}
Additionally, on 1,000 news articles published after all model releases, frontier models maintain high agreement with Pangram~4.1 (93--98\%) and falsely label 0.5--5\% of the human-written articles (as classified by Pangram) as AI-written (\S\ref{app:ty_contamination}). Frontier models may struggle more with this data because newer models could have generated these articles.

\paragraph{Dates are referenced, but not decisive in strong detector predictions.}
In 9.5\% of its explanations of fully human (human ideas and human prose) texts, Opus~5.5 argues that the text predates LLMs because of a date in the text (``before 2022''). When it makes this argument on human-ideated, AI-written texts its prose accuracy falls from 80.3\% to 35.4\%. Replacing all years with "2026`` in 190 dated documents changes at most four of Opus~5.5's, GPT-6.1 Sol's or Gemini~4 Carbon's 190 verdicts per attribute, no more than rerunning the same prompt (\autoref{tab:rq1_year2026}). On the other hand, DeepSeek~V4.1 Flash (open-weight model) changes 29 ideas verdicts to AI and one to human.

\paragraph{Explanations are consistent across detectors and echo known traits of AI writing.}
\label{sec:expl_design}\label{sec:expl_agreement}\label{sec:expl_features}
Taking advantage of LLMs' capabilities to explain their judgments~\citep{huang2023largelanguagemodelsexplain, zheng2023judgingllmasajudge}, we ask the detectors to disclose the key observations used to make their decisions, where each observation includes quoted evidence, whether it suggests a human or an AI, and why (\autoref{fig:explanation-example}).\footnote{The exact prompt is in our \repolink{repository}.} Asking for observations before the verdict also raises accuracy. Sol gains more than 10 points of idea accuracy on human ideas in AI prose  (\autoref{tab:prompt_dev}). 
Across pairs of detectors, the explanations largely cite the same evidence and read it the same way, but agree less on which piece of evidence decided the verdict.\footnote{A Claude Opus~5.5 judge, checked against two human annotators, matches the evidence in each pair of explanations on 2,854 documents.
Agreement statistics in \S\ref{app:explanation-agreement}.} 
The cues they cite match prior work (\autoref{tab:prior}). Hedged assertions point to AI prose and explicitly stated themes and morals to AI ideas~\citep{kim2026argumentcollapsellmsflatten, russell2026storyscope}, while precise factual and technical detail points to human ideas~\citep{rajendhran2026idealensdetectingaiideas}.

\section{Which model wrote the text?}
\label{sec:model_detection}

Frontier models can tell whether a human or an AI wrote the text (\S\ref{sec:ai_vs_human}), but can they tell which model wrote it? Prior work finds that the writing of different LLMs is highly homogeneous \citep{jiang2026artificial, kim2026argumentcollapsellmsflatten}, that their stories converge on a shared narrative space \citep{russell2026storyscope}, and that frontier models from different labs share a dense style that is harder for people to read \citep{clark2026hard}. This convergence makes attributing a text to its source model much more difficult, especially given that there are numerous labs, each with several model versions.
We ask five frontier models
to name the model that wrote each text, with no list of candidates. We find that their errors are consistent with public reports of distillation
\citep{anthropic2026distillation, openai2026deepseek}. For example, the three strongest detectors attribute almost every text by Moonshot's Kimi~K3 to Claude.

\paragraph{Task and evaluation.}
\label{sec:md_task}\label{sec:method_holistic}
We evaluate five frontier models, Claude Fable~5.2, Claude Opus~5.5, GPT-6.1 Sol, Gemini~4 Argon, and DeepSeek~V4.1 Flash.\footnote{Each detector is run at medium effort, except DeepSeek~V4.1 Flash, which has no medium level and runs at high.} 
Using a minimally modified version of the human vs. AI prompt (\S\ref{sec:ai_vs_human}), each names the model it thinks wrote the text in a zero-shot setting, without tools or a candidate list. Along with its answer, each detector then gives its key observations, reasoning and decisive factors for its decision in the format of \autoref{fig:prompt-overview}. We score each answer at three levels: the lab (e.g., Anthropic), the model family (e.g., Claude Opus) and the exact model (e.g., Claude Opus~5.5). We report the share of texts attributed correctly at each level.

\paragraph{Data.}

We generate new texts for the model attribution task in the following three domains (\autoref{tab:rq2_domains}), using 34 models from 11 labs released between September 2024 and October 2026:
\begin{itemize}[leftmargin=1.7em, labelsep=0.5em, nosep]
    \item[\faBookOpen] \textbf{Stories}: We take 25 story prompts from StoryScope \citep{russell2026storyscope}, a dataset of AI-written fiction, and generate one story per prompt.\footnote{We reuse the original StoryScope stories for the five models it covers.}
    \item[\faNewspaper] \textbf{News}: We take 25 articles\footnote{Pangram~4.1 labels 23 of the 25 articles human and two, both from Global Voices, mixed.} from 17 outlets, published from 9--20 October 2026.
    Each model receives an article's topic and length and writes its own article.
    \item[\faPenNib] \textbf{Op-eds}: We take 25 English briefs from MARGO~\citep{gill2026llms}, a dataset of op-eds written in eight languages by five models. Each model receives a brief and writes an op-ed from it.\footnote{MARGO was released after all five detectors became available. We reuse its original op-eds for five writing models.}
    Briefs in the other seven languages are covered in \S\ref{app:ai_vs_ai_full}.
\end{itemize}

\begin{table}[t]
\centering\footnotesize
\caption[Model Attribution]{Percentage of texts correctly attributed to their lab, model family, or exact model. Lab accuracy is also split by the generating lab: Anthropic, Google, and OpenAI (logos), and the other eight labs. See \autoref{tab:ai_vs_ai_holistic} for per-model results.}
\label{tab:ai_vs_ai_levels}
\setlength{\tabcolsep}{2.25pt}
\begin{tabular}{@{}l>{\columncolor{tabkey}}rrr@{\hspace{6pt}}rr@{}}
\toprule
 & \multicolumn{3}{c}{\textbf{Attribution level}} & \multicolumn{2}{c}{\textbf{Lab by group}} \\
\cmidrule(lr){2-4}\cmidrule(lr){5-6}
Detector & \multicolumn{1}{r}{Lab} & Family & Model &
\multicolumn{1}{r}{%
  \labicon{anthropic}\,%
  \labicon{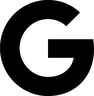}\,%
  \labicon{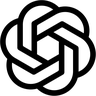}} & Other \\
\midrule
\jFable{Fable~5.2} & \textbf{59.9} & \textbf{41.0} & \textbf{10.7} & \textbf{99.8} & \textbf{35.1} \\
\jOpus{Opus~5.5} & 49.7 & 31.6 & 5.8 & 97.8 & 19.8 \\
\jArgon{Gemini~4 Argon} & 44.8 & 27.9 & 7.1 & 97.5 & 12.0 \\
\jSol{GPT-6.1 Sol} & 22.6 & 7.9 & 0.1 & 58.8 & 0.2 \\
\jDeepSeek{DeepSeek~V4.1 Flash} & 11.1 & 3.1 & 0.0 & 28.7 & 0.1 \\
\bottomrule
\end{tabular}
\end{table}

\subsection{Lab and model attribution}
\label{sec:md_shared}

\paragraph{Fable~5.2, Opus~5.5 and Argon identify the big three labs almost perfectly, while Sol and DeepSeek default to OpenAI and Anthropic.}
Identifying a lab is important to understand which provider's model wrote a text.
Fable~5.2, Opus~5.5 and Argon can identify text from OpenAI, Anthropic, and Google with 98\% accuracy or more. Opus~5.5 does so even for the two models released after it (Sol and Argon). Their overall accuracy is much lower, at 60\%, 50\% and 45\% (\autoref{tab:ai_vs_ai_levels}), because they rarely recognize models from the other eight labs, though Fable~5.2 does so far more often than Opus~5.5 (35\% vs.\ 20\%).
 Sol and DeepSeek lag far behind at 23\% and 11\%, and mostly answer OpenAI and Anthropic, so we focus on Fable~5.2, Opus~5.5 and Argon for the rest of this section.

\paragraph{Naming the exact model depends on how long it has been public.}
Even Fable~5.2, Opus~5.5 and Argon rarely name the exact model, and instead name an older model from the same lab. How often they name it depends on how long the model has been public---Opus~5.5 names the exact model of about half of the texts from the three models released before mid-2025, but names none from the 24 models released in 2026. Fable~5.2 does better on the three older models (64\%), but names the exact model for only one of 1,792 texts from the 2026 models.

\paragraph{Misattributions of new open-weight models to Anthropic and OpenAI are consistent with reported distillation.}
A model trained on another lab's outputs can inherit that lab's style, so a detector may name the lab it learned from. Anthropic has reported distillation attacks on Claude by DeepSeek, Moonshot, and MiniMax \citep{anthropic2026distillation}, and later named Zhipu and Alibaba \citep{anthropic2026misuse}. Consistent with these reports, Fable~5.2, Opus~5.5, and Argon attribute nearly all Kimi~K3 and GLM-5.3 texts to Claude (89\% or more), even though Fable~5.2 names Moonshot for 93\% and 77\% of Kimi~K2.5 and K2.6 texts, the two releases before K3.
Similarly, Elon Musk testified that xAI ``partly'' distilled OpenAI models \citep{muskvaltman2026transcript} and a US advisory says that Alibaba and MiniMax distilled GPT-5 \citep{cisa2026distillation}: detectors attribute most texts by Grok~4.7, Qwen3.8~Max, MiniMax~M3, and M3.1~Flash to GPT.\footnote{To check that these patterns are not an artifact of our detectors falling back to a known lab for models  released after knowledge cutoff, we also run a stylometric analysis, and find the same shifts, even after removing the information-dense style shared by recent models \citep{clark2026hard} (\S\ref{app:rq2_analysis}).}

\subsection{Model writing profiles}
\label{sec:md_explanations}

Fable~5.2, Opus~5.5, and Argon tell the big three labs apart almost perfectly, but what do they think each lab reads like? Each detector explains its answer for every text (\S\ref{sec:md_task}), so we can create a writing profile of each lab from these explanations.\footnote{We group explanations into cues and keep those the detector notes on a lab's texts significantly more often than on any other lab's (\S\ref{app:rq2_analysis}). We treat the other eight labs as one lab.}

\begin{table*}[!t]
\centering\footnotesize
\definecolor{lpClay}{HTML}{B4501C}\definecolor{lpClayBg}{HTML}{FBEDE4}
\newcommand{\lpmark}[1]{{\setlength{\fboxsep}{0pt}\colorbox{lpClayBg}{\textcolor{lpClay}{\strut #1}}}}
\caption[Lab writing profiles]{\jOpus{Opus~5.5} and \jArgon{Argon} agree on what sets each lab's writing apart. Each cell gives examples of the cues a detector often cites for that lab, using only texts where it names the right lab. \lpmark{Highlighted} cues are often cited by both detectors. Full model profiles can be found \href{https://model-profiles.github.io/}{here}.}
\label{tab:lab-profiles}
\vspace{6pt}
\begingroup
\newcommand{\lpchip}[1]{{\setlength{\fboxsep}{1.0pt}\setlength{\fboxrule}{0.4pt}\fbox{\rule[-0.6pt]{0pt}{5.2pt}{\fontsize{5.4}{6}\selectfont\MakeUppercase{#1}}}}}
\newcommand{\lpname}[3]{\labicon{#3}\,\textbf{#1}\par\vspace{1pt}{\scriptsize #2}}
\newcommand{\lpgenre}[1]{\textbf{#1}\enspace}
\newcommand{\lpsty}[1]{\par\vspace{3.5pt}\lpchip{Style}\enspace\textit{#1}}
\setlength{\tabcolsep}{5pt}\renewcommand{\arraystretch}{1.05}
\begin{tabularx}{\linewidth}{@{}>{\raggedright\arraybackslash}p{.125\linewidth}>{\raggedright\arraybackslash}X>{\raggedright\arraybackslash}X@{}}
\toprule
\textbf{Lab} & \textbf{\jOpus{Opus~5.5}} & \textbf{\jArgon{Gemini~4 Argon}} \\
\midrule\addlinespace[2pt]
\lpname{OpenAI}{GPT}{openai} &
GPT's stories run on \lpmark{aphoristic }\allowbreak \lpmark{lines} and \lpmark{clipped, }\allowbreak \lpmark{sparse }\allowbreak \lpmark{dialogue}.\par
Its news \lpmark{flags }\allowbreak \lpmark{gaps }\allowbreak \lpmark{in }\allowbreak \lpmark{the }\allowbreak \lpmark{evidence }\allowbreak \lpmark{and }\allowbreak \lpmark{does }\allowbreak \lpmark{not }\allowbreak \lpmark{fill }\allowbreak \lpmark{them}, \lpmark{hedges }\allowbreak \lpmark{and }\allowbreak \lpmark{calibrates }\allowbreak \lpmark{its }\allowbreak \lpmark{claims}, spells out what it is not claiming, and stays generic where other models supply specifics.\par
Its op-eds return to \lpmark{aphoristic }\allowbreak \lpmark{lines} and hedge their claims.
\lpsty{\lpmark{one-line }\allowbreak \lpmark{paragraphs} (stories); \lpmark{no }\allowbreak \lpmark{serial }\allowbreak \lpmark{comma} (news); balanced antitheses (op-eds).} &
GPT's stories run on \lpmark{clipped, }\allowbreak \lpmark{sparse }\allowbreak \lpmark{dialogue} and \lpmark{aphoristic }\allowbreak \lpmark{lines}, and keep the climax understated.\par
Its news \lpmark{flags }\allowbreak \lpmark{gaps }\allowbreak \lpmark{in }\allowbreak \lpmark{the }\allowbreak \lpmark{evidence}, \lpmark{hedges }\allowbreak \lpmark{and }\allowbreak \lpmark{calibrates }\allowbreak \lpmark{its }\allowbreak \lpmark{claims}, avoids inventing specifics and separates concepts that are easily conflated.\par
Its op-eds return to \lpmark{aphoristic }\allowbreak \lpmark{lines}, pull apart easily conflated concepts and end paragraphs on a kicker.
\lpsty{\lpmark{one-line }\allowbreak \lpmark{paragraphs} (stories); balanced antitheses (stories); \lpmark{no }\allowbreak \lpmark{serial }\allowbreak \lpmark{comma} (news).} \\
\addlinespace[4pt]\specialrule{0.3pt}{0pt}{0pt}\addlinespace[4pt]
\lpname{Anthropic}{Claude}{anthropic} &
Claude's stories return to recurring motifs, end quietly and call back to their opening.\par
Its news keeps a balanced, non-partisan stance, gathers reactions from several stakeholders, gives quoted speakers \lpmark{polished, }\allowbreak \lpmark{aphoristic }\allowbreak \lpmark{lines} and closes on a quotation.\par
Its op-eds are \lpmark{dense }\allowbreak \lpmark{with }\allowbreak \lpmark{accurate }\allowbreak \lpmark{real-world }\allowbreak \lpmark{detail}, grant an opposing point before countering it and treat the other side charitably.
\lpsty{\lpmark{clauses }\allowbreak \lpmark{strung }\allowbreak \lpmark{together }\allowbreak \lpmark{with }\allowbreak \lpmark{repeated }\allowbreak \lpmark{``and'', }\allowbreak \lpmark{or }\allowbreak \lpmark{with }\allowbreak \lpmark{no }\allowbreak \lpmark{conjunctions }\allowbreak \lpmark{at }\allowbreak \lpmark{all} (stories); \lpmark{spaced }\allowbreak \lpmark{dashes} (news).} &
Claude's stories correct themselves mid-sentence and repeat a word for emphasis.\par
Its news gives quoted speakers \lpmark{polished, }\allowbreak \lpmark{aphoristic }\allowbreak \lpmark{lines}, attributes them to sources that are unnamed or labeled by role, and includes specialist detail.\par
Its op-eds are \lpmark{dense }\allowbreak \lpmark{with }\allowbreak \lpmark{accurate }\allowbreak \lpmark{real-world }\allowbreak \lpmark{detail}, adopt the voice of a publication and end on a crafted final line.
\lpsty{\lpmark{clauses }\allowbreak \lpmark{strung }\allowbreak \lpmark{together }\allowbreak \lpmark{with }\allowbreak \lpmark{repeated }\allowbreak \lpmark{``and'', }\allowbreak \lpmark{or }\allowbreak \lpmark{with }\allowbreak \lpmark{no }\allowbreak \lpmark{conjunctions }\allowbreak \lpmark{at }\allowbreak \lpmark{all} (stories); ``the way\,\dots'' phrasing (stories); \lpmark{spaced }\allowbreak \lpmark{dashes} (news); short, punchy sentences (news, op-eds).} \\
\addlinespace[4pt]\specialrule{0.3pt}{0pt}{0pt}\addlinespace[4pt]
\lpname{Google}{Gemini}{google} &
Gemini's news gives invented people \lpmark{common }\allowbreak \lpmark{default }\allowbreak \lpmark{names}.
\lpsty{stacked adjectives (stories); clustered sensory detail (stories); italics for emphasis and thought (stories); \lpmark{mixed }\allowbreak \lpmark{curly }\allowbreak \lpmark{and }\allowbreak \lpmark{straight }\allowbreak \lpmark{quotes} (stories); \lpmark{datelines} (news); intensifiers and dramatic adjectives (news, op-eds).} &
Gemini's stories give invented people common default names.\par
Its news uses the same \lpmark{common }\allowbreak \lpmark{default }\allowbreak \lpmark{names}.
\lpsty{\lpmark{mixed }\allowbreak \lpmark{curly }\allowbreak \lpmark{and }\allowbreak \lpmark{straight }\allowbreak \lpmark{quotes} (stories); \lpmark{datelines} (news); scare quotes (news); stock idioms and clich\'es (op-eds).} \\
\bottomrule
\end{tabularx}
\endgroup
\end{table*}

\paragraph{Fable~5.2, Opus~5.5, and Argon hold consistent profiles of each lab.}
The detectors describe the big three labs in largely the same way (\autoref{tab:lab-profiles}).\footnote{We show only cues that a detector cites on at least 25\% of a lab's texts within a domain (stories, news, op-eds), and at least 15 points and 1.5 times more often than on any other lab's texts.} \textbf{GPT} hedges its claims, flags gaps instead of inventing specifics, and leans on aphorisms. \textbf{Claude} treats opposing views charitably, writes with dry wit, and ends quietly. \textbf{Gemini} stacks adjectives and piles on intensifiers and stock clich\'es, and invents named sources and quotes in its news articles. 
All three also tie the same punctuation habits to each lab, though Fable~5.2 and Argon cite them more than twice as often as Opus~5.5. GPT uses curly quotes and drops the Oxford comma, Gemini mixes curly and straight quotes, and Claude puts spaces around its dashes. For Claude, Opus~5.5 and Argon tend to identify different traits, while Fable~5.2 finds most of both.

\paragraph{Kimi~K3 and GLM-5.3 share Claude's traits, while Grok~4.7, Qwen3.8~Max and MiniMax share GPT's.}
A detector may attribute a text to a lab either because the text shares that lab's traits or because the detector does not recognize the model that wrote it and falls back to a known lab. 
For most open-model texts attributed to Claude it is the latter. The detectors cite few of the traits that set Claude apart. Kimi~K3 and GLM-5.3 are the exception. All three detectors note the same dry wit in their texts as in Claude's. Opus~5.5 also notes Claude's bookend endings and closing quotations, and Argon notes the same spaced dashes and parenthetical asides. The detectors find all of these far less often in earlier Kimi and GLM models' texts.\footnote{Claude Opus~5.5 and Gemini~4 Argon attribute most Kimi~K2 texts (88\% and 93\%) and nearly all Kimi~K3 texts to Claude, but they find far more of Claude's traits in K3's texts than in K2's.} Likewise, Fable~5.2 and Opus~5.5 also find several of GPT's traits in Grok~4.7, Qwen3.8~Max, MiniMax~M3 and M3.1~Flash texts, with little to no Claude traits.

\section{Which human wrote the text?}
\label{sec:human_attribution}
While the space of plausible outputs in the model attribution task is just a few dozen models, naming the \emph{person} behind a human-written text is an open-world task with countless possible outputs~\citep{koppel2011wild,luyckx2011aasize} and much less data available per author. To evaluate this capability while mitigating the confound of data contamination, we collect 18,677 texts from six domains, each either \emph{private} (never posted publicly) or \emph{recent} (published after each tested model's knowledge cutoff, and after most of their release dates). Even so, frontier models unmask many authors across all of these domains from just a single text, given a minimally modified prompt from the previous sections and no tools, web access, or candidate lists. The best-performing model we test is Claude Fable~5.2, which names the reviewer of 55\% of 15K anonymous ICML 2020 reviews and the username behind 31\% of recent Reddit comments. Paraphrasing a human text with AI is only a partial defense, as content-related cues persist.

\subsection{Experimental setup}
Given a single text and a minimally modified version of the prompt from previous sections, models return the human author's name (or username for Reddit), a list of supporting observations, and a confidence level. Unlike \citet{lermen2026deanonymization}, who augment LLMs with web search and candidate pools to match users across platforms, we provide no tools, web access, or candidate lists. To keep personal account information out of the models' context, we query every model through an API, not a chat app or coding CLI (\autoref{tab:rq3_access}).\footnote{There were a few exceptions run through the Claude Code CLI due to budget considerations, though only when familiarity with the CLI account holder was unlikely to impact accuracy, including the Reddit reasoning-effort comparison (\autoref{tab:rq3_reddit_effort_familiarity}), as well as Opus~5.5's name-only familiarity ratings of Reddit accounts and Nature reviewers.} In an early run, we observed the Claude Code CLI quote the account user's email when attributing a text the account owner wrote, a phenomenon studied in \citet{zhong2026userawareness}.

\paragraph{Private Data.}
We collect two sets of peer reviews whose reviewer identities were never made public and therefore could not have been memorized from model training data.
\begin{itemize}[leftmargin=1.7em, labelsep=0.5em, itemsep=0.1em]
  \item[\raisebox{-0.15em}{\includegraphics[height=1em]{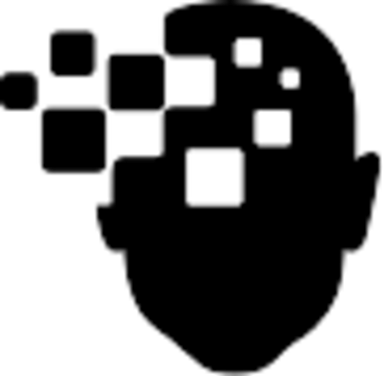}}] \textbf{ICML 2020 peer reviews}: 15,136 peer reviews written by 3,342 unique reviewers for ICML 2020 and hosted on Microsoft's Conference Management Toolkit (CMT). We can study these reviews because coauthor Hal Daum\'e~III served as program co-chair for the conference and has IRB approval to analyze them. Daum\'e ran all of the experiments on this corpus, so reviews were never shared with the other authors. We report only aggregate results and do not reveal any text from individual peer reviews or reviewer identities.
  \item[\faUsers] \textbf{Reviews from this paper's authors}: 398 peer reviews written between 2015 and 2026 and voluntarily contributed by four of this paper's coauthors (JR, JW, MK, MI), as well as two PhD students in the UMD CLIP Lab. The group includes faculty members, industry researchers, and students. While the number of unique reviewers is much lower than in the ICML 2020 set, these reviews were submitted to 26 distinct double-blind ML/NLP venues (main conferences and workshops) such as ACL, EMNLP, and NAACL. They were pulled from Softconf and OpenReview, with none of the reviewer identities and very few\footnote{The text of 19 of these reviews is publicly available on OpenReview without an associated reviewer identity.} of the review texts being public.
\end{itemize}

\begin{figure*}[!t]
  \centering
  \includegraphics{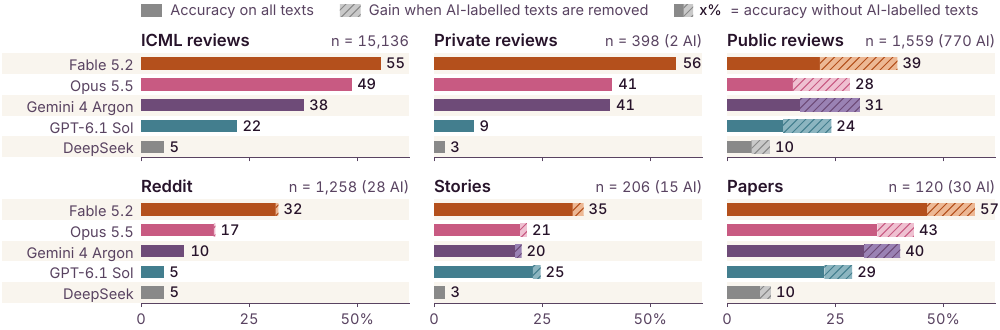}
  \caption{Author accuracy by domain on all human authorship attribution datasets. Solid bars show accuracy on all texts in the domain; striped portions indicate the gain in performance when texts Pangram~4.1 labels as AI are removed. The printed numbers are accuracy on non-AI labelled texts (includes human and mixed). Accuracy is over each model's answered texts, leaving out refusals and invalid replies.}
  \label{fig:rq3_domains}
\end{figure*}

\paragraph{Recent Data.}
We collect four datasets of publicly available writing published online after September 22, 2026, the release date of Claude Opus~5.5. Since a model's reported knowledge cutoff generally applies only to pretraining and not post-training, only data published after a model's release date is completely safe from contamination.\footnote{By design, Claude Opus~5.5 is completely free from training data contamination in our experiments. Eight percent of public data is from before GPT-6.1 Sol's release, 23\% is before Gemini 4 Argon, and 28\% before DeepSeek V4.1 Flash; pre-release data stems only from stories and public peer reviews. Claude Fable~5.2 was released about 10 days after all public data collection completed, and its knowledge cutoff is June 2026.}

\begin{itemize}[leftmargin=1.7em, labelsep=0.5em, itemsep=0.1em]
    \item[\faGlobe] \textbf{Public peer reviews}: 1,559 peer reviews from 47 open-review journals, including 90 reviews from the Nature family.
    Unlike ICML and our private reviews, these reviews are from disciplines outside AI/NLP and provide a mix of prominent expert reviewers (from Nature Portfolio journals) and lesser-known researchers.
  \item[\raisebox{-0.15em}{\includegraphics[height=1em]{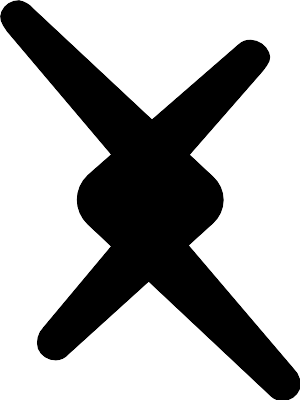}}] \textbf{NLP papers}: 120 papers uploaded to arXiv between 19--22 October 2026. Papers were sampled from the most recent papers categorized as cs.CL (Computer Science, Computation and Language) on arXiv. 
  \item[\faBookOpen] \textbf{Stories}: 206 short stories and novel excerpts from 187 authors primarily gathered from literary magazine sites, posted between 23 September--23 October 2026.
  \item[\faReddit] \textbf{Reddit}: 1,258 comments from 870 unique usernames in 32 subreddits, posted between 18--29 October 2026. Each comment is at least 200 words and excludes quoted text. Active subreddits with long posts generally lean towards technical subjects (e.g., programming, math) as shown in \autoref{fig:rq3_subreddits}.
\end{itemize}

\paragraph{Models and evaluation.}
We benchmark four frontier models from Anthropic (Fable 5.2, Opus 5.5), OpenAI (GPT-6.1 Sol), and Google (Gemini 4 Argon), as well as one open-weight model with accessible chain-of-thought (DeepSeek~V4.1 Flash).\footnote{All models run at the provider's default reasoning effort: Claude Opus~5.5, GPT-6.1 Sol and Gemini 4 Argon at medium and Claude Fable~5.2 and DeepSeek~V4.1 Flash at high. See \S\ref{sec:who_identified} for how reasoning affects detection, and \autoref{tab:rq3_full} for results with Gemini~4 Carbon.} We report exact match accuracy (after normalizing titles, accents, and punctuation) for each dataset other than ICML 2020, for which we employ a more lenient match that allows for abbreviated first names and omitted or added middle names.\footnote{Claude Fable~5.2 names the reviewer of 50.2\% of ICML reviews with exact match vs. 55.5\% with the more lenient matching, and qualitative inspection shows few false matches in the latter setting.} For NLP papers, we consider a prediction correct if the model names any co-author, while confirmed pen-names are verified and counted as correct for fiction. For Reddit texts, we use a slightly modified prompt to ask specifically for usernames. For public reviews and NLP papers, accuracy is reported on texts that Pangram~4.1 does not label AI (human and mixed labels), unless otherwise specified.

\subsection{Results}
\label{sec:ha_results}
Despite the difficulty of this task, frontier models are surprisingly effective on both private and recent datasets, and they are rapidly improving. Fable~5.2, the best performing model in all domains (\autoref{fig:rq3_domains}), exhibits considerable gains over Opus 5.5, which was released less than two months prior. 

\paragraph{Models can effectively deanonymize peer reviews.} On the ICML 2020 corpus, Fable~5.2 correctly identifies the author of 55.5\% of the reviews, followed by Opus~5.5 (48.8\%), Argon (37.7\%), Sol (22.3\%), and DeepSeek (5.4\%). On private peer reviews, Fable~5.2 also leads with 56\% accuracy, while on public reviews from the \emph{Nature} journal family it reaches 78\%. The private reviews consist of 266 standard reviews and 132 meta-reviews, the latter of which are much shorter in length, come from only two authors, and are harder to attribute. Fable~5.2 names the writer of 68\% of reviews but only 31\% of meta-reviews. Opus~5.5 and Argon also have twice the accuracy on standard reviews as on meta-reviews (\autoref{tab:rq3-private-meta}). DeepSeek, the only open model, performs very poorly in all domains despite reporting the highest average confidence levels; its chain-of-thought reveals no clear reasoning insights and consists primarily of hedging and indecision.

\paragraph{Models can name pseudonymous users.} For example, models can identify the username behind Reddit comments without any additional context. Fable~5.2 identifies the Reddit user behind 31\% of comments, nearly double Opus~5.5 which identifies 17\%. Accuracy varies greatly based on subreddit. Fable~5.2 names 79\% of comments in r/cpp and 50\% in r/ProgrammingLanguages, but only 13\% in r/AskAnthropology. This may also reflect training data sampling procedures from frontier labs, as subreddits about programming might be valuable to improve coding performance. As discussed in \S\ref{sec:who_identified}, pseudonymity is especially fragile for active users in niche communities. Fable~5.2 names 26\% of the 860 users at least once (\S\ref{app:rq3_author_level}).

\paragraph{Confident answers are almost always right.} Each model's prediction includes a confidence level. Claude models tend to understate theirs. On public data, Fable~5.2's mean reported confidence is 17\% while achieving 27\% accuracy. When Fable~5.2's confidence is 25\% or more (25\% of responses), its predictions are 87\% correct (\autoref{fig:rq3_confidence}). Furthermore, confident mistakes are rare. Of all Fable~5.2 answers, 0.3\% report both at least 80\% confidence and name the wrong person; the corresponding percents are 0.4\% for Opus~5.5, 3.3\% for Argon, and 5.1\% for DeepSeek. The threat to anonymity therefore is greater than the accuracy levels suggest; one can rely on the strongest model's confident answers and be right almost every time.

\paragraph{Contamination is unlikely to be causing the high accuracy.} Models do well on our private datasets, whose authorship was never made public. In fact, Fable~5.2, Opus~5.5, and Argon perform better on our private datasets than on the public Reddit comments and stories. All of the public data is posted after the release date of Opus 5.5, which correctly names an author of 43\% of recent NLP papers\footnote{43\% is Opus~5.5's accuracy on the 90 non-AI written papers as labeled by Pangram. Opus~5.5 names 35\% of all 120 papers.} and the usernames behind 17\% of Reddit comments. Accuracy instead correlates with training data exposure and familiarity, which we discuss in \S\ref{sec:who_identified}.

\paragraph{Accuracy drops on AI-generated texts.} Since our prompt explicitly requests a human author's name, models perform poorly on texts across the four recent datasets that Pangram~4.1 labels as AI-generated. The discrepancy is most noticeable in public peer reviews and papers that increasingly contain AI-written text~\citep{liang2024monitoring, emi2025iclr, naddaf2025conference}. Pangram flags 770 of the 1,559 public reviews as AI-generated, and models name the author of only 1--4\% of these reviews (\autoref{tab:atrey_rq3_pangram}). Omitting AI-generated texts, visualized by the striped bars in \autoref{fig:rq3_domains}, raises accuracy on public reviews by 75--87\% for every model (Opus~5.5 rises from 15\% to 28\%), and raises it on papers. The remaining domains have negligible numbers of AI-generated texts.

\begin{figure}[t]
  \centering
  \includegraphics{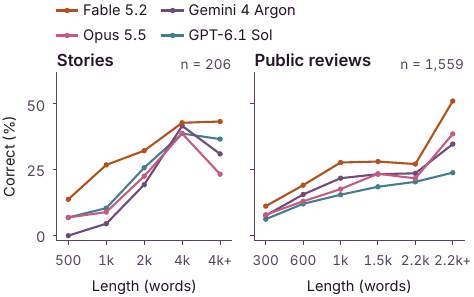}
  \caption{Author attribution accuracy increases with the number of words in a text. Tick labels are the upper limit of each length bin.}
  \label{fig:rq3_length_accuracy}
\end{figure}

\paragraph{Longer texts are easier to attribute for all models.} The correlation between length and accuracy is strongest for public reviews and stories, the latter of which has the widest range of lengths (181--20,947 words). Fable~5.2 identifies only 13.8\% of 29 stories under 500 words and 43.3\% of 30 over 4K words (\autoref{fig:rq3_length_accuracy}). These results align with findings in AI-text detection, where shorter texts are harder to detect~\citep{emi2024pangram}.

\paragraph{Attribution accuracy is stable across multiple runs.}
We test the stability of Fable~5.2 predictions on Reddit comments over five independent runs. Overall accuracy stays nearly identical, between 29.9--31.1\%, though there is some variation in predictions. Only 32.6\% of comments receive the same author prediction every time. When predictions change, it is usually from one wrong answer to another; 63.2\% of comments are predicted incorrectly in all five runs, while 25\% are predicted correctly in all five runs. Familiar authors are predicted far more reliably. Fable~5.2 is correct in all five runs for 92.7\% of the 96 comments from the 39 most familiar users (rated 3), but for only 0.4\% of 281 comments from completely unfamiliar users (rated 0). Accuracy remains stable over other domains as well; see \S\ref{app:rq3_stability}.

\subsection{What aspects of the text give authors away?}
\label{sec:ha_analysis}%
We analyze the generated explanations to see which features of the text each model relies on when naming the author. Generally, the explanations cover both the text's content as well as its writing style.\footnote{Their claims are also largely factual. We extract verifiable claims from Opus~5.5, Argon, Sol and DeepSeek explanations for 2,000 ICML 2020 reviews. Gemini~4 Argon with Google Search Grounding marks 95.9\% of the 5,414 verifiable claims it assessed as supported by external evidence (\S\ref{app:rq3_grounding}).} For instance, when identifying the author of the short story ``Lost Child'' (published in ``The New Yorker'' magazine in September 2026), Fable~5.2 cites ``this particular pairing of syntax and subject matter'' as decisive factors when correctly predicting Joyce Carol Oates (\autoref{tab:attribution-examples}).

\paragraph{Self-citations make reviewers easy to name, but few reviews contain them.}
Reviewers often cite their own papers in their reviews, and these self-citations have historically been used to identify authors under blind review~\citep{matsubara2020citations, caragea2019myth, legoues2018anonymization}. Of the 1,559 public reviews, only 106 cite the reviewer's own work (82 explicitly list reviewer's name in citation, 24 do not). On these reviews, Fable~5.2 names the reviewer 80\% of the time, compared with 17\% of reviews without a self-citation (\autoref{tab:atrey_rq3_checks}).

\begin{figure}[t]
  \centering
  \includegraphics{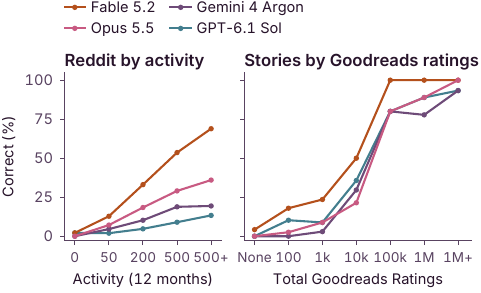}
  \caption{Accuracy is higher for more active Reddit users and authors with many Goodreads ratings. \textbf{Left:} Reddit commenters, binned by total number of posts and comments in the subreddit over the previous 12 months. \textbf{Right:} authors, binned by the total number of Goodreads ratings across their books; ``None'' indicates authors without a Goodreads profile (69/187).}
  \label{fig:rq3_fame}
\end{figure}

\begin{table}[t]
\centering\footnotesize
\setlength{\tabcolsep}{3pt}

\newcommand{\jStackT}[3]{\begin{tabular}[b]{@{}c@{}}#1\\[-1pt]{\scriptsize #2}\\[-2pt]{\scriptsize #3}\end{tabular}}
\caption{Accuracy (\%) with and without self-citations in the text; bold marks the best per column in each group. Self-citation is an obvious clue to author identity, but models still name writers without them. The first row of each group is accuracy on all its texts; the indented rows below are accuracy on subsets, with and without self-citations.}

\label{tab:atrey_rq3_checks}
\begin{tabularx}{\columnwidth}{>{\raggedright\arraybackslash\hangindent=1em}X r rrrr}
\toprule
Texts & \textit{n} & \jStackT{\iconFable}{\textcolor{inkFable}{Fable}}{\textcolor{inkFable}{5.2}} & \jStackT{\iconOpus}{\textcolor{inkOpus}{Opus}}{\textcolor{inkOpus}{5.5}} & \jStackT{\iconArgon}{\textcolor{inkArgon}{Gemini 4}}{\textcolor{inkArgon}{Argon}} & \jStackT{\iconSol}{\textcolor{inkSol}{GPT-6.1}}{\textcolor{inkSol}{Sol}} \\
\midrule
\textbf{All public reviews} & 1,559 & 21.4 & 15.2 & 16.9 & 13.0 \\
\quad Cites own work & 106 & \textbf{80.2} & \textbf{76.2} & \textbf{76.4} & \textbf{60.4} \\
\quad No self-citation & 1,453 & 17.1 & 10.7 & 12.5 & 9.5 \\
\qquad $\hookrightarrow$Not AI-labelled & 704 & 33.7 & 21.7 & 24.6 & 18.8 \\
\midrule
\textbf{All papers} & 120 & 46.2 & 34.7 & 31.7 & 22.5 \\
\quad Cites own work & 18 & \textbf{72.2} & \textbf{61.1} & \textbf{66.7} & \textbf{50.0} \\
\quad No self-citation & 102 & 41.6 & 30.0 & 25.5 & 17.6 \\
\bottomrule
\end{tabularx}
\end{table}

\begin{table*}[!t]
\centering\footnotesize
\definecolor{cueTeal}{HTML}{437E8E}\definecolor{cueTealBg}{HTML}{E3EFF2}\definecolor{cueClay}{HTML}{B4501C}\definecolor{cueClayBg}{HTML}{FBEDE4}\definecolor{cueRose}{HTML}{6E2042}\definecolor{cueRoseBg}{HTML}{FBEBF1}
\newcommand{\exmark}[2]{{\setlength{\fboxsep}{0pt}\colorbox{#1Bg}{\textcolor{#1}{\strut #2}}}}
\caption[Claude Fable 5.2 examples]{\protect\iconFable\,Claude Fable~5.2 bases each decision on concrete evidence ranging from mixed apostrophes to a reviewer's self-description. Each explanation lists quoted evidence and names decisive factors. In each row, a \exmark{cueRose}{highlighted} reason shares its color with the words in the text it describes.}
\label{tab:attribution-examples}
\vspace{6pt}
\begingroup
\definecolor{predictionCorrect}{HTML}{00AB89}
\newcommand{\predlabel}[1]{{\fontsize{5.5}{7}\selectfont\MakeUppercase{#1}}}
\newcommand{\predtag}[1]{{\setlength{\fboxsep}{1.6pt}\mbox{\colorbox{black!9}{\rule{0pt}{6.6pt}\textbf{#1}}\,\textcolor{predictionCorrect}{\ensuremath{\checkmark}}}}}
\newcommand{\predmeta}[1]{{\scriptsize #1}}
\setlength{\tabcolsep}{2pt}\renewcommand{\arraystretch}{1.05}
\begin{tabularx}{\linewidth}{@{}>{\raggedright\arraybackslash\hyphenpenalty=10000}p{.132\linewidth}>{\raggedright\arraybackslash}p{.30\linewidth}>{\raggedright\arraybackslash}p{.165\linewidth}>{\raggedright\arraybackslash}X@{}}
\toprule
\textbf{Task} & \textbf{Text (excerpt)} & \textbf{\iconFable\,Prediction} & \textbf{\iconFable\,Explanation (excerpt)} \\
\midrule
\textbf{Human or AI?}\par Story from TwiceTold: \textit{lantern and firefly} & ``\exmark{cueRose}{The }\allowbreak \exmark{cueRose}{most }\allowbreak \exmark{cueRose}{meaningful }\allowbreak \exmark{cueRose}{light }\allowbreak \exmark{cueRose}{is }\allowbreak \exmark{cueRose}{that }\allowbreak \exmark{cueRose}{which }\allowbreak \exmark{cueRose}{is }\allowbreak \exmark{cueRose}{provided }\allowbreak \exmark{cueRose}{in }\allowbreak \exmark{cueRose}{moments }\allowbreak \exmark{cueRose}{of }\allowbreak \exmark{cueRose}{need.} [\ldots] \exmark{cueTeal}{From }\allowbreak \exmark{cueTeal}{that }\allowbreak \exmark{cueTeal}{day }\allowbreak \exmark{cueTeal}{on,} travelers to the forest always saw two lights [\ldots] the lantern never again \exmark{cueClay}{mistook }\allowbreak \exmark{cueClay}{greatness }\allowbreak \exmark{cueClay}{with }\allowbreak \exmark{cueClay}{size}.'' & \predlabel{Ideas}\par\predtag{AI}\ \predmeta{90\% AI}\par\vspace{4pt}\predlabel{Prose}\par\predtag{Human}\ \predmeta{10\% AI} & \textbf{Ideas:} ``The deciding factor is the concentration of stock choices: [\ldots] and the \exmark{cueRose}{stated }\allowbreak \exmark{cueRose}{moral} with a \exmark{cueTeal}{"from }\allowbreak \exmark{cueTeal}{that }\allowbreak \exmark{cueTeal}{day }\allowbreak \exmark{cueTeal}{on" }\allowbreak \exmark{cueTeal}{coda}.''\par\smallskip \textbf{Prose:} ``The deciding factors are the \exmark{cueClay}{non-idiomatic }\allowbreak \exmark{cueClay}{and }\allowbreak \exmark{cueClay}{ungrammatical }\allowbreak \exmark{cueClay}{slips: }\allowbreak \exmark{cueClay}{"mistook }\allowbreak \exmark{cueClay}{greatness }\allowbreak \exmark{cueClay}{with }\allowbreak \exmark{cueClay}{size"}'' \\
\addlinespace[2.5pt]
\textbf{Which AI?}\par AI story from \textit{StoryScope} & ``Luke \exmark{cueTeal}{didn\textquotesingle{}t} hesitate. [\ldots] He looked at \exmark{cueTeal}{Harry\textquoteright{}s} trembling shoulders. [\ldots] \exmark{cueRose}{The }\allowbreak \exmark{cueRose}{Gulf }\allowbreak \exmark{cueRose}{was }\allowbreak \exmark{cueRose}{a }\allowbreak \exmark{cueRose}{graveyard, }\allowbreak \exmark{cueRose}{but }\allowbreak \exmark{cueRose}{for }\allowbreak \exmark{cueRose}{tonight, }\allowbreak \exmark{cueRose}{the }\allowbreak \exmark{cueRose}{brothers }\allowbreak \exmark{cueRose}{were }\allowbreak \exmark{cueRose}{still }\allowbreak \exmark{cueRose}{ghosts }\allowbreak \exmark{cueRose}{in }\allowbreak \exmark{cueRose}{the }\allowbreak \exmark{cueRose}{making, }\allowbreak \exmark{cueRose}{drifting }\allowbreak \exmark{cueRose}{on }\allowbreak \exmark{cueRose}{the }\allowbreak \exmark{cueRose}{tide }\allowbreak \exmark{cueRose}{of }\allowbreak \exmark{cueRose}{a }\allowbreak \exmark{cueRose}{dying }\allowbreak \exmark{cueRose}{world.} [\ldots] \exmark{cueRose}{Row. }\allowbreak \exmark{cueRose}{Steer. }\allowbreak \exmark{cueRose}{Live. }\allowbreak \exmark{cueRose}{And }\allowbreak \exmark{cueRose}{never, }\allowbreak \exmark{cueRose}{ever }\allowbreak \exmark{cueRose}{laugh.}'' & \predlabel{Model}\par\predtag{Gemini 3 Flash}\par\vspace{3pt}\predlabel{Developer}\par\predtag{Google}\par\vspace{3pt}\predmeta{Confidence 40\%} & ``The deciding factor is the combination of the \exmark{cueRose}{stacked, }\allowbreak \exmark{cueRose}{aphoristic }\allowbreak \exmark{cueRose}{endings }\allowbreak \exmark{cueRose}{with }\allowbreak \exmark{cueRose}{a }\allowbreak \exmark{cueRose}{staccato }\allowbreak \exmark{cueRose}{coda} and the \exmark{cueTeal}{mixed }\allowbreak \exmark{cueTeal}{curly }\allowbreak \exmark{cueTeal}{and }\allowbreak \exmark{cueTeal}{straight }\allowbreak \exmark{cueTeal}{apostrophes} with trailing spaces. The stylistic habits alone could fit several models, but the formatting artifacts are much more specific to Gemini [\ldots] The specific version remains a best guess.'' \\
\addlinespace[2.5pt]
\textbf{Which human?}\par Review from \textit{Nature Sustainability} & ``\exmark{cueRose}{I }\allowbreak \exmark{cueRose}{am }\allowbreak \exmark{cueRose}{primarily }\allowbreak \exmark{cueRose}{a }\allowbreak \exmark{cueRose}{wellbeing }\allowbreak \exmark{cueRose}{researcher} [\ldots] \exmark{cueClay}{emphasises} [\ldots] \exmark{cueClay}{what }\allowbreak \exmark{cueClay}{matters }\allowbreak \exmark{cueClay}{for }\allowbreak \exmark{cueClay}{policy }\allowbreak \exmark{cueClay}{seems }\allowbreak \exmark{cueClay}{rather }\allowbreak \exmark{cueClay}{to }\allowbreak \exmark{cueClay}{be }\allowbreak \exmark{cueClay}{any }\allowbreak \exmark{cueClay}{substantive }\allowbreak \exmark{cueClay}{magnitudes }\allowbreak \exmark{cueClay}{of }\allowbreak \exmark{cueClay}{any }\allowbreak \exmark{cueClay}{relationships.} [\ldots] \exmark{cueTeal}{between }\allowbreak \exmark{cueTeal}{long-run }\allowbreak \exmark{cueTeal}{changes }\allowbreak \exmark{cueTeal}{in }\allowbreak \exmark{cueTeal}{GDP }\allowbreak \exmark{cueTeal}{per }\allowbreak \exmark{cueTeal}{capita }\allowbreak \exmark{cueTeal}{and }\allowbreak \exmark{cueTeal}{long-run }\allowbreak \exmark{cueTeal}{changes }\allowbreak \exmark{cueTeal}{in }\allowbreak \exmark{cueTeal}{wellbeing}'' & \predlabel{Reviewer}\par\predtag{Caspar Kaiser}\par\vspace{3pt}\predmeta{Confidence 27\%} & ``The deciding factor is the combination of three things: the \exmark{cueRose}{self-description }\allowbreak \exmark{cueRose}{as }\allowbreak \exmark{cueRose}{"primarily }\allowbreak \exmark{cueRose}{a }\allowbreak \exmark{cueRose}{wellbeing }\allowbreak \exmark{cueRose}{researcher"}, the \exmark{cueTeal}{specialist }\allowbreak \exmark{cueTeal}{long-run }\allowbreak \exmark{cueTeal}{reading }\allowbreak \exmark{cueTeal}{of }\allowbreak \exmark{cueTeal}{the }\allowbreak \exmark{cueTeal}{Easterlin }\allowbreak \exmark{cueTeal}{Paradox}, and \exmark{cueClay}{UK-style }\allowbreak \exmark{cueClay}{orthography }\allowbreak \exmark{cueClay}{with }\allowbreak \exmark{cueClay}{faintly }\allowbreak \exmark{cueClay}{non-native }\allowbreak \exmark{cueClay}{syntax}. The Easterlin point narrows the field to a few specialists, and the spelling and phrasing eliminate most of the others.'' \\
\addlinespace[2.5pt]
\textbf{Which human?}\par Human story: \textit{Lost Child} & ``\exmark{cueTeal}{For}, with Kit\textquoteright{}s father, something was always either going to happen or had already happened, [\ldots] \exmark{cueClay}{Hurry, }\allowbreak \exmark{cueClay}{sweetie! }\allowbreak \exmark{cueClay}{God }\allowbreak \exmark{cueClay}{damn.} \exmark{cueRose}{Not }\allowbreak \exmark{cueRose}{used }\allowbreak \exmark{cueRose}{to }\allowbreak \exmark{cueRose}{dressing }\allowbreak \exmark{cueRose}{herself.} [\ldots] but maybe Daddy wouldn\textquoteright{}t notice.'' & \predlabel{Author}\par\predtag{\shortstack[l]{Joyce Carol\\Oates}}\par\vspace{3pt}\predmeta{Confidence 93\%} & ``The deciding factor is the combination of style and theme: \exmark{cueRose}{fragmentary, }\allowbreak \exmark{cueRose}{subject-dropping }\allowbreak \exmark{cueRose}{sentences}, \exmark{cueTeal}{sentence-initial }\allowbreak \exmark{cueTeal}{"For"}, and \exmark{cueClay}{unquoted }\allowbreak \exmark{cueClay}{menacing }\allowbreak \exmark{cueClay}{speech}, [\ldots] this particular pairing of syntax and subject matter is characteristic of Oates and outweighs the generic literary-fiction features.'' \\
\bottomrule
\end{tabularx}
\endgroup
\end{table*}

\begin{figure}[!t]
  \centering
  \includegraphics{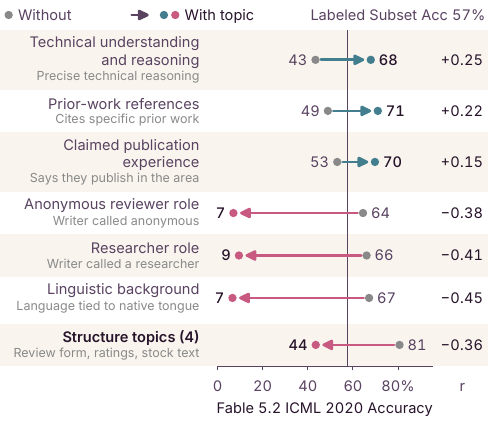}
  \caption{The presence of topics related to structure has a negative correlation with accuracy, while references to prior works or technical reasoning have positive correlation. Claude Fable~5.2's accuracy when naming ICML 2020 reviewers is shown with its strongest positive and negatively correlated topics (top of arrow) and without them (gray dot). Each topic's Pearson correlation is listed on the right side. Labeled Subset Acc is Fable~5.2's accuracy on the 5,000 subset of reviews we labeled.}
  \label{fig:rq3-topics}
\end{figure}

\paragraph{The full text, not just its topic, is necessary to name a paper's authors.}\label{sec:ha_topic_only}
A model could name the author of a paper by knowing who works on its topic, without recognizing anyone's writing. To measure how far topic knowledge alone contributes, we take the 34 cs.CL papers posted to arXiv after Opus~5.5's training cutoff that Pangram~4.1 labels human. Opus~5.5 receives either five keywords describing each paper's topic or its title and abstract, and is asked to name two researchers (early career and senior) who work on the problem. It is explicitly told that it is not being asked who wrote the paper. A given name is correct if it is any of that paper's co-authors (\autoref{tab:rq3_topic_only}). We compare the minimal topic inputs with our default method of giving the full main text. Topic keywords identify an author for only 12\% of papers, against 53\% from the title and abstract and 76\% from the full text.

\paragraph{Using a list of candidate reviewers raises accuracy on ICML 2020 reviews.}\label{sec:ha_candidates}
Our main results are in the open-world setting, meaning we don't give a list of candidate authors. However, someone trying to unmask say, an ICML reviewer, could easily use the candidate list of authors publicly posted on the ICML website. We run Fable~5.2 on 2,000 ICML reviews, reusing our prompt with the addition of the 3,672 names on the public ICML 2020 reviewer list\footnote{See \href{https://icml.cc/Conferences/2020/Reviewers}{https://icml.cc/Conferences/2020/Reviewers}.}, telling the model that its predicted author must come from this list. Since authors could opt out from public mentions, the maximum accuracy that Fable~5.2 could achieve using the candidate list on the 2,000 reviews is 84.1\%. Fable~5.2 achieves 76.1\% ($\pm$2.1) accuracy, compared with 55.41\% ($\pm$2.5) on the same 2,000 reviews without the list ($\pm$ indicates 95\% confidence interval).

\paragraph{Topics extracted from explanations correlate with accuracy and signal model uncertainty.}
We use GPT-6.1 Sol to independently generate and assign topics to the explanations generated for five models on 5,000 ICML 2020 reviews and 1,258 Reddit comments~\citep{pham2024topicgptpromptbasedtopicmodeling}. We calculate the correlation of each topic with attribution accuracy, finding weak to moderate, though statistically significant, correlations for more than half of the topics.\footnote{See \autoref{tab:deniz_agent_rq3_topic_desc} for a list and description of each topic.} For Fable~5.2, the presence of a structure-related topic (form, layout) is negatively correlated with accuracy (r=$-0.36$), while some topics like prior-work references are positively associated with accuracy (\autoref{fig:rq3-topics}). Topics like ``researcher role'' or ``anonymous reviewer role'' (model describing the reviewer as a ``researcher'' or ``anonymous'') have lower than average confidence levels and among the strongest negative correlations. These results suggest that correct predictions cite content-oriented evidence like technical reasoning or references to prior work in their explanations, while incorrect ones contain comments on generic review structure or the author's anonymity.

\section{Who can be identified?}
\label{sec:who_identified}

In this section, we look at what makes certain authors (AI or human) recognizable. We observe that the common factor is exposure in the training data. Detectors are most accurate on the writers they have plausibly seen most during training. We first show how this holds across our three research questions and then introduce a measure of familiarity that works without access to pretraining data. We use this measure to explore how reasoning effort and paraphrasing change who can be identified. Finally, we discuss the broader implications of our experiments for anonymity.

\subsection{Exposure and familiarity}
Exposure in pretraining data is a plausible common contributor to performance across all author attribution settings.
For AI text detection (\S\ref{sec:ai_vs_human}), models now train on a web in which a large share of text is machine-written, which likely improves their ability to detect it. For instance, 31.1\% of quality-filtered web tokens from August 2026 are labeled as AI-generated~\citep{russell2026aitoken}. For attribution among models (\S\ref{sec:model_detection}), detectors name OpenAI, Anthropic, and Google far more reliably than the other eight labs. This is consistent with the model market share reported by Pangram~\citep{masrour2026marketshare}.\footnote{We note that this gap is also consistent with distillation from larger labs, and it is likely that both exposure and distillation contribute to weaker performance on these labs' models.} The rest of this section focuses on what makes \emph{human authors} identifiable: we test several proxies for exposure (e.g., mentions of their names in open pretraining corpora, comment activity on Reddit, visibility and seniority) and find that directly prompting the model for its familiarity with a person correlates most strongly with how identifiable their texts are. 

\paragraph{Higher author visibility correlates with increased human attribution accuracy.}
In \S\ref{sec:human_attribution}, we define domain-specific measures for human author visibility (\autoref{fig:rq3_fame}). On Reddit, we measure visibility by a user's post and comment count in a given subreddit over the past 12 months. Using this metric, Claude Fable~5.2's attribution accuracy rises from 2\% for inactive users to 13\% for those with 1--50 posts, reaching 69\% for users with over 500 posts. Over the same span, Claude Opus~5.5's accuracy increases from 0\% to 36\%. For stories, we use the total number of Goodreads ratings across an author's books. Here, Fable~5.2 identifies 4\% of stories by authors without a Goodreads profile, 50\% by those with 1k--10k ratings, and 100\% by authors with over 10k ratings. Similarly, Opus~5.5, Gemini~4 Argon, and GPT-6.1 Sol fail to identify any authors without a profile, but successfully attribute 96\% of those with over 1M ratings. A similar trend holds for private reviews (\autoref{tab:rq3-private-meta}). Opus~5.5 and Argon correctly identify the two most senior reviewers on 66\% and 71\% of their 174 standard (non meta-reviews) reviews, respectively, compared to 23\% and 14\% across the 92 reviews written by the other four contributors.

\paragraph{A model's self-reported familiarity predicts attribution.}
Language models can retrieve details about entities like writers and public figures from their parametric memory~\citep{kadavath2022language, yin2023know, ferrando2025entity}. Because the pretraining data of frontier models is unavailable to inspect directly, we elicit this knowledge by prompting the model to rate its familiarity with a name or username on a 0--3 scale (see our \repolink{repository} for prompt; \S\ref{app:rq3-familiarity}). We make two independent calls and average results to compute \emph{self-reported familiarity}, which strongly corresponds with attribution accuracy. Opus~5.5 and Argon identify 1.3--4.2\% of texts by writers they rate 0, compared to 57.7--60.8\% of texts by writers they rate 3. About 90\% of Opus~5.5's successful attributions and 79\% of Argon's come from writers rated $\geq 1.5$, who account for 45\% and 30\% of texts, respectively (\autoref{tab:rq3_familiarity_ratings}). The rating also transfers across models, with one model's familiarity score predicting another model's accuracy (\autoref{tab:rq3_familiarity_cross}). These results mirror factual recall, where obscure entities are recalled less often~\citep{mallen2023trust, kandpal2023large}.

\paragraph{Self-reported familiarity outperforms other exposure proxies.}
Where we can compare them directly, a model's familiarity rating predicts attribution better than other domain-specific visibility measures. On Reddit, Opus~5.5's rating separates the Reddit accounts it does and does not identify with an AUC of 0.89, compared to 0.78 for posting activity over the prior 12 months. Even among accounts with similar posting activity, Opus~5.5 identifies 17--56\% of those it rates familiar ($\geq$ 1.5 average rating), compared to at most 14\% of the rest. On ICML 2020 reviews, Fable~5.2 names the reviewer of 0\% and 76\% of reviews by reviewers it rates 0 and 3, respectively, a monotonic increase with familiarity. The strongest training exposure proxy for ICML 2020 reviews, how often a reviewer's name appears near an ML venue name in Dolma~v1.7\footnote{We count occurrences of the reviewer's name within 1,000 tokens (before or after, in the same document) of ICML, NeurIPS, ICLR, or ``Advances in Neural Information Processing Systems'', using the infini-gram index of Dolma~v1.7 (2.6T Llama-2 tokens; \citealp{liu2024infinigram}).}~\citep{soldaini2024dolma}, has $\rho = 0.31$ correlation with Fable~5.2's per-reviewer recall,\footnote{Per-reviewer recall is the fraction of each reviewer's reviews that Fable~5.2 attributes correctly.} against $0.13$ for paper count and $0.11$ for citations. %

\begin{figure}[t]
  \centering
  \includegraphics{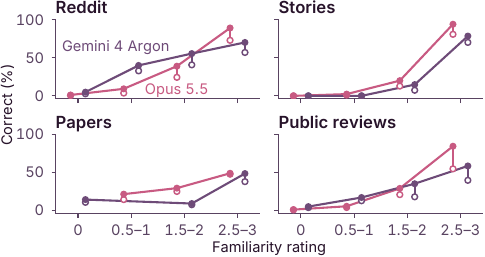}
  \caption{Models are more likely to name authors they are familiar with. Familiarity is the model's self-reported rating from 0--3. Solid dots on lines are original texts; hollow dots show average accuracy of two paraphrased versions.} 
  \label{fig:atrey_rq3_familiarity}
  \label{fig:rq3-rewrite-familiarity}
\end{figure}

\paragraph{Increasing reasoning effort helps most on accounts the models know only a little.}
We collect 442 comments from 216 Reddit accounts, sampled across all familiarity levels, and group accounts by each model's own rating. Fable~5.2, Opus~5.5, and Argon all show their largest accuracy gains from reasoning on accounts rated \emph{a little familiar}, +35.8, +21.4, and +12.4 points, respectively (\autoref{tab:rq3_reddit_effort_familiarity}). For Fable~5.2, accuracy on these 19 accounts increases from 8.4\% at low effort to 44.2\% at max, 
with most of the improvement coming at the highest setting. Fable~5.2 names about 90\% of \emph{very familiar} accounts at every effort level, and no \emph{unfamiliar} accounts until max effort, where it reaches 8\%. Higher reasoning can recover weakly memorized data, similar to factual question answering, where thinking helps models recall rare facts more than popular ones~\citep{calderon2026emptyshelves, gekhman2026thinkingtorecall}.

\begin{table*}[t]
\centering

\caption{Reasoning effort helps most when the model is only a little familiar with the account, and cannot make up for an account it does not know. Reddit author attribution accuracy (\%) by reasoning effort and by how familiar the model says it is with the account. \tabDelta: paired gain from the lowest to the highest effort; it is calculated using the subset of accounts with valid answers at both efforts, so it can differ from the difference between the two cells. We sort each account by the mean of its own familiarity ratings of the account (0--3 scale); the range of the mean familiarity ratings is in parentheses.}
\label{tab:rq3_reddit_effort_familiarity}
\begin{minipage}{\textwidth}
\footnotesize
\setlength{\tabcolsep}{2.7pt}
\renewcommand{\arraystretch}{1.05}
\begin{tabular*}{\linewidth}{@{\extracolsep{\fill}}c@{\hspace{3pt}}l
  rrrr>{\columncolor{tabkey}}r
  rrrr>{\columncolor{tabkey}}r
  rrrr>{\columncolor{tabkey}[\tabcolsep][0pt]}r@{}}
\toprule
 & & \multicolumn{5}{c}{\textbf{\jFable{Fable~5.2}}}
 & \multicolumn{5}{c}{\textbf{\jOpus{Opus~5.5}}}
 & \multicolumn{5}{c}{\textbf{\jArgon{Gemini~4 Argon}}} \\
\cmidrule(lr){3-7}\cmidrule(lr){8-12}\cmidrule(lr){13-17}
 & Familiarity
 & Low & Med. & High & Max & \multicolumn{1}{r}{\tabDelta}
 & Low & Med. & High & Max & \multicolumn{1}{r}{\tabDelta}
 & None & Low & Med. & High & \multicolumn{1}{r}{\tabDelta} \\
\midrule
 & Any familiarity {\scriptsize(0.75--3)} & 67.5 & 68.9 & 69.8 & 80.2 & +11.2 & 51.9 & 55.5 & 56.0 & 68.7 & +15.5 & 54.0 & 53.0 & 58.6 & 63.2 & +9.1 \\
\multirow{4}{*}{\tikz[baseline=0pt]{\draw[black,line width=.7pt,-{Stealth[length=3pt,width=3.5pt]}] (0,13pt) -- (0,-13pt);}} & \quad Unfamiliar {\scriptsize{(0--0.5)}} &0.0 & 0.0 & 0.0 & 7.7 & +8.0 & 0.0 & 0.0 & 0.5 & 0.5 & +0.5 & 4.8 & 5.4 & 6.6 & 8.4 & +3.5 \\
 & \quad A little familiar {\scriptsize{(0.75--1.5)}} &8.0 & 6.7 & 18.0 & 38.2 & \textbf{+35.8} & 11.2 & 12.9 & 14.7 & 33.6 & \textbf{+21.4} & 27.6 & 27.0 & 35.2 & 40.0 & \textbf{+12.4} \\
 & \quad Moderately familiar {\scriptsize{(1.75--2.5)}} &69.7 & 71.9 & 71.1 & 84.9 & +11.9 & 61.0 & 65.4 & 65.3 & 76.0 & +14.5 & 57.4 & 55.6 & 59.4 & 64.2 & +6.8 \\
 & \quad Very familiar {\scriptsize{(2.75--3)}} &88.1 & 89.5 & 90.1 & 89.0 & \tabminus1.4 & 76.0 & 77.5 & 78.2 & 87.8 & +11.8 & 70.5 & 70.7 & 77.6 & 81.5 & +11.0 \\
\midrule
 & 95\% CI for pooled gain & \multicolumn{5}{c}{(6.2--16.4)} & \multicolumn{5}{c}{(10.3--20.8)} & \multicolumn{5}{c}{(3.4--15.1)} \\
\bottomrule
\end{tabular*}
\end{minipage}
\end{table*}

\paragraph{Models can imitate the users they recognize.}
The knowledge that lets a model recognize a writer also lets it write like one. We ask Opus~5.5 to generate two Reddit comments in the style of 565 accounts, given only the username and subreddit, and pass each comment to Opus~5.5. It names the target from at least one imitation for 74 of 166 accounts (45\%) that Opus~5.5 both rates as familiar and identifies from real comments, vs. 9 of 219 (4\%) familiar accounts it fails to identify, 5 of 31 (16\%) accounts it identifies but does not rate familiar, and 0 of 149 unfamiliar, unidentified accounts. Although this test only measures whether an imitation is attributed to the target by a model rather than a human reader, it suggests that memorized author profiles can be used for impersonation.

\subsection{Implications for anonymity}
\paragraph{Once a model is familiar with a writer, it is hard to stay anonymous.}
Even when a text is brand new, a model can identify its writer from what it learned about them elsewhere; once that knowledge is in the weights, the writer cannot do anything about it. This leaves people with the largest public footprints, such as active social media users or established authors, permanently exposed. Even paraphrasing the text does not completely remove the risk of deanonymization. Moreover, the set of vulnerable writers expands as models improve. On Reddit posts, Fable~5.2 has 14.3\% higher accuracy than Opus~5.5 and additional reasoning extends identification to writers a model knows only weakly.

\paragraph{Refusals are rare.}
Fable~5.2 and Opus~5.5 refuse most often, on under 2\% of all documents (counting provider blocks and written refusals that decline to do the task at all) and mostly on public peer reviews.\footnote{Refusals are far rarer through the API than through Claude Code (\S\ref{app:rq3_refusal_types}). Claude Fable~5.2 gives no answer on 323 of 1,258 Reddit comments through Claude Code but on only 14 (1.1\%, 3 blocked by the provider, 11 written refusals) through OpenRouter batches pinned to Anthropic, which our results use. Models also sometimes make anonymous predictions, which give an explanation and conclude that the author is ``anonymous''.} The Anthropic provider blocks Fable~5.2 responses on 20 public peer reviews, 3 Reddit comments, and 1 paper, and Opus~5.5 responses on 28 public peer reviews and 2 papers. We additionally identify 151 anonymous predictions from Fable~5.2 and 130 from Opus~5.5, mostly citing insufficient evidence; only 50 of Fable~5.2's 162 written refusals and anonymous predictions mention privacy or concerns about incorrect attributions. Summing provider blocks, written refusals and anonymous predictions, Fable~5.2 names no author on 6.9\% of documents and Opus~5.5 on 6.0\%.

\paragraph{Paraphrasing lowers accuracy, though more familiar authors remain easier to identify.}
We test paraphrasing as a defense against author identification, using Argon at medium effort to revise texts from \S\ref{sec:human_attribution}. The prompt asks it to make the author's style less identifiable while preserving content and meaning, a form of authorship obfuscation \citep{brennan2012adversarial, mahmood2019girl, fisher-etal-2024-jamdec}. We rewrite each text twice, run authorship attribution on both rewrites with Opus~5.5 and Argon, and average the two accuracies. Accuracy drops in nearly all domains. On Reddit, Opus~5.5 drops from 16.8\% to 10.4\% and Argon from 10.0\% to 6.9\%; on public reviews, the two models drop from 20.8--23.5\% to 14.8--15.2\%. Private reviews show larger drops, from 40.7--41.2\% to 14.9--15.6\%. Stories and papers are the least affected by paraphrasing. Argon drops from 18.6\% to 15.8\% on stories and Opus~5.5 from 34.7\% to 31.4\% on papers (\autoref{tab:rq3-rewrite-accuracy}). Familiar authors remain easier to identify after paraphrasing. Of the 34 stories whose authors Opus~5.5 rates very familiar (2.5--3 on the 0--3 scale), it names 32 authors before rewriting and 27.5 after (average over the two rewrites) (\autoref{fig:rq3-rewrite-familiarity}).

\section{Related work}
\label{sec:related}

\paragraph{Detecting AI-generated text and model attribution.}
Traditional AI-text detectors rely on statistical approaches \citep{gehrmann-etal-2019-gltr, mitchell2023detectgpt, bao2024fastdetectgpt, hans2024spotting} or fine-tuned classifiers \citep{hu2023radar, glickenhaus2026pangram4technicalreport, adam2026gptzerorobustdetectionllmgenerated}. Earlier prompt-based detection concluded prompting was unreliable at flagging AI-generated text \citep{bhattacharjee2024fighting, russell-etal-2025-people}. \citet{Bao2025DecouplingCA} and \citet{rajendhran2026idealensdetectingaiideas} detect AI-generated prose and AI-generated ideas separately.
\citet{uchendu-etal-2020-authorship} introduce the task of classifying which model generated a text, given the options of several generators. \citet{uchendu-etal-2021-turingbench-benchmark} and \citet{sun2025idiosyncrasies} find that classifiers trained on a fixed set of generators can tell them apart. In zero-shot settings, LLMs can pick up on AI-generated stylistic cues \citep{barkhordar2026style, davidson-etal-2024-self} but identification fails once those cues are controlled. \citet{panickssery2024llm} find that models identify and favor their own generations; \citet{choi-etal-2025-agent} show that this capability extends to conversational partners and that LLMs reliably recognize peers from their own family and detect prominent out-of-family models (such as GPT and Claude) across stylistic, reasoning, and alignment signatures.

\paragraph{Identifying human authors.}
Early approaches used stylometry or embeddings to identify authorship (\citealp{mosteller1964inference, narayanan2012feasibility, rivera-soto-etal-2021-learning, stamatatos2009survey, koppel2011wild, wegmann2022same}). For peer review, \citet{hill2003myth} identify authors 40--45\% of the time from their citations alone, and \citet{bauersfeld2023cracking, caragea2019myth} identify the authors of 73\% of papers from their text and references. Many prior studies point to the threat of deanonymization, but use limited candidate pools, retrieval setups, or agentic tooling to extract information about an author (\citealp{lermen2026deanonymization, zhang-zhang-2026-de, gloire2026threaten, staab2024beyond,tan2025open, huang-etal-2024-large,wang2026echo}). \citet{adewumi-etal-2025-limitations} and \citet{hicke2025inner} find LLMs can identify famous authors from text alone but struggle with less popular writers, and in open-world quote attribution, accuracy likewise rises with the speaker's fame \citep{berman2026attribution}. \citet{wang2025catch} find that pre-2026 models like GPT-4o struggle to imitate everyday authors' styles from a few samples, while our results suggest that Opus~5.5 can imitate Reddit users it already recognizes without any writing samples (\S\ref{sec:who_identified}). LLMs' familiarity with a text relates to how often that specific text appears in pretraining data \citep{chang-etal-2023-speak,kandpal2023large}.

\section{Conclusion}
\label{sec:conclusion}
We demonstrate that closed-source frontier models can match trained AI-text detectors, name the specific model behind AI-generated texts, and perform human authorship attribution given only a single text and a zero-shot prompt. They can name the reviewer of about half of 15K anonymous ICML reviews that were never made public and therefore could not have been memorized during training. Attribution is easier when the model is familiar with the human author; we can predict which authors a model can unmask by asking it how familiar it is with a given name. Models are also rapidly improving at this task. Claude Fable~5.2 (released on November 9, 2026) is the strongest model we tested at naming both labs and people. Defenses are limited, since models rarely refuse through the API, writing cannot be deleted from existing weights, and AI paraphrasing only partly helps. Anyone who has published enough writing online can no longer assume that an unsigned text keeps them anonymous.

\section*{Acknowledgements}
We thank the University of Maryland Computational Linguistics and Information Processing (CLIP) Lab and the SFU AI lab for their feedback and support.
We thank Google for HG's Gemini Academic Program award and Pangram Labs for a credit award, both of which enabled this research. MK is supported by NSERC Discovery Grant RGPIN-2026-04984 and Rajan Faculty Recruitment Award. We also thank Bin Hu for contributing Anthropic API credits for this project.

\subsection*{AI use statement}
We used AI tools to assist in data collection, code implementation, prompt engineering, editing our draft, and to generate tables and figures from our results. Claude Opus~5.5 was used to judge agreement between model explanations, and was validated by human annotators. AI tools were also queried to provide feedback on our draft and identify gaps in our experiments; Claude Code suggested the inclusion of author-level accuracies and confidence levels in \S\ref{sec:human_attribution}. We have manually reviewed all AI-assisted content and take full responsibility for the final content of this work.

\subsection*{Ethical considerations}

The people most affected by this work are those who rely on anonymity. Reviewers in double-blind review could face retaliation from authors whose papers they rejected, and pseudonymous forum users could be harassed or doxxed once their accounts are linked to them. This is especially dangerous for those whose safety depends on staying anonymous, such as whistleblowers, activists, and survivors of abuse. Models are also able to imitate users they recognize (\S\ref{sec:who_identified}), so the same knowledge could be used to impersonate them. We recognize that publishing this work may lead some readers to try these attacks themselves; however, we believe it is more important to raise awareness of these capabilities to spur future research on defenses, especially given how simple and accessible they are.

\textbf{Existing capability.} All of our results come from simply prompting publicly available models. Each model names the author of a single text in a zero-shot setting, without tools, web access, a candidate list, or writing samples, nor do we train any models. This requires no programming or data collection, as anyone is able to paste a text and our prompt into a chat interface. Models also rarely refuse this task when queried through an API. Even Fable 5.2 and Opus 5.5, the models that refused most often, did so on fewer than 2\% of documents through the API. The risk is therefore already present in widely accessible models, and exists regardless of whether this paper was published.

\textbf{The risk will grow without intervention.} We expect this risk to increase as models improve. Claude Fable~5.2 identifies the author of nearly twice as many Reddit comments as Opus~5.5, which was only released two months earlier. Furthermore, a higher reasoning effort also allows models to identify authors they are barely familiar with. Nothing a writer does now can remove what a model has already learned about them from its weights. Delaying this work would not make these writers any safer, and would only keep them from knowing they are at risk.

\textbf{Benefits of publication.} We believe the benefits of publishing this work outweigh the risks, since most people who write anonymously are likely unaware that a model can identify them from a single text. Model developers are also able to reduce this risk, for example by having models refuse these requests more often (we informed Anthropic and OpenAI about our findings prior to publication). This is already possible, with Fable~5.2 refusing 323 of 1,258 Reddit comments through Claude Code compared to only 14 through the API. Our results also show that paraphrasing a text is only a partial defense, and better defenses against both identification and impersonation still need to be developed. 

\textbf{Data protection.} The ICML 2020 reviews were analyzed under IRB approval (\#4656-1) by the conference's program co-chair, a coauthor of this paper. Only he accessed the reviews, and we report only aggregate results, without publishing any part of the reviews or revealing author identities. Model providers received only the text of each review, with no identifying metadata. Reviewer names, emails and IDs, and the key linking reviews to reviewers were never sent to any model provider or other third party. All scoring was done locally. DeepSeek~V4.1 Flash was restricted to OpenRouter hosts with zero data retention. All other models were accessed through paid APIs (via OpenRouter or Google's Gemini API) whose terms exclude API inputs and outputs from model training. The private reviews were contributed voluntarily by their own authors. All other data is publicly available, and we report results on it in aggregate. The only people we mention by name are already well-known authors or researchers and their associated texts are publicly signed (not anonymous); we name them only as illustrative examples. We do not release any author predictions or private data.

\FloatBarrier
\bibliography{references}
\bibliographystyle{icml2026}

\appendix
\onecolumn
\let\appendixsection\section\renewcommand{\section}{\FloatBarrier\appendixsection}

\renewcommand{\topfraction}{0.95}
\renewcommand{\bottomfraction}{0.9}
\renewcommand{\textfraction}{0.05}
\renewcommand{\floatpagefraction}{0.85}
\setcounter{topnumber}{4}
\setcounter{bottomnumber}{3}
\setcounter{totalnumber}{6}

\let\originalsection\section\let\originalsubsection\subsection\let\originalsubsubsection\subsubsection
\newcommand{\appendixdemoteon}{\let\section\originalsubsection\let\subsection\originalsubsubsection\let\subsubsection\paragraph}
\newcommand{\appendixdemoteoff}{\let\section\originalsection\let\subsection\originalsubsection\let\subsubsection\originalsubsubsection}

\section{Detectors and reproducibility}
\label{app:shared}

In this section, we list every model we evaluate in \S\ref{sec:ai_vs_human}, with its lab, weights, release date (the lab's announcement, or the day OpenRouter first listed it when the lab gives no day), reasoning effort, access route and cost per 1,000 documents (\autoref{tab:atrey_detectors}). Our prompts are available in our \repolink{repository}.

{\footnotesize\setlength{\tabcolsep}{3.5pt}
\begin{longtable}{l l l l l l r}
\caption{Complete information of all models used in the timeline (\autoref{sec:ai_vs_human}).}\label{tab:atrey_detectors}\\
\toprule
Detector & Lab & Weights & Released & Effort & Access & \$/1k docs \\
\midrule
\endfirsthead
\toprule
Detector & Lab & Weights & Released & Effort & Access & \$/1k docs \\
\midrule
\endhead
Gemini 4 Neon & Google & closed & 2026-11-18 & high & Vertex Batch & 97.22 \\
Gemini 4 Hydrogen & Google & closed & 2026-11-12 & high & Vertex Batch & 29.96 \\
Gemini 4 Carbon & Google & closed & 2026-11-10 & medium & Gemini API & 92.76 \\
Gemini 4 Argon & Google & closed & 2026-10-13 & medium & Gemini Batch & 11.21 \\
Gemini 3.8 Flash & Google & closed & 2026-09-02 & medium & Gemini Batch & 7.30 \\
Gemini 3.6 Flash & Google & closed & 2026-07-21 & medium & Gemini Batch & 8.11 \\
Gemini 3.5 Flash-Lite & Google & closed & 2026-07-21 & minimal & Gemini Batch & 1.80 \\
Gemini 3.1 Flash-Lite & Google & closed & 2026-05-07 & minimal & Gemini Batch & 0.95 \\
Gemini 3.1 Pro & Google & closed & 2026-02-19 & high & Gemini Batch & 31.70 \\
Gemini 3 Flash & Google & closed & 2025-12-17 & high & Gemini Batch & 8.90 \\
Gemini 2.5 Pro & Google & closed & 2025-06-17 & dynamic & OpenRouter\textsuperscript{b} & 50.62 \\
\addlinespace[3pt]
GPT-6.2 Sol & OpenAI & closed & 2026-11-17 & medium & OpenAI Batch & 12.58 \\
GPT-6.2 Astra & OpenAI & closed & 2026-11-17 & medium & OpenAI Batch & 51.16 \\
GPT-6.1 Terra & OpenAI & closed & 2026-10-26 & medium & OpenAI Batch & 5.24 \\
GPT-6.1 Sol & OpenAI & closed & 2026-09-29 & medium & OpenAI Batch & 12.52 \\
GPT-6 Luna & OpenAI & closed & 2026-09-22 & medium & OpenAI Batch & 0.75 \\
GPT-6 Astra & OpenAI & closed & 2026-09-03 & medium & OpenAI Batch & 74.02 \\
GPT-5.6 Terra & OpenAI & closed & 2026-07-09 & medium & OpenAI Batch & 14.20 \\
GPT-5.6 Luna & OpenAI & closed & 2026-07-09 & medium & OpenAI Batch & 1.94 \\
GPT-5.6 Sol & OpenAI & closed & 2026-07-09 & medium & OpenAI Batch & 38.25 \\
GPT-5.5 & OpenAI & closed & 2026-04-24 & medium & OpenAI Batch & 59.89 \\
GPT-5.4 & OpenAI & closed & 2026-03-05 & medium & OpenAI API & 74.26 \\
GPT-5.2 & OpenAI & closed & 2025-12-11 & medium & OpenAI Batch & 18.12 \\
GPT-5.1 & OpenAI & closed & 2025-11-13 & medium & OpenAI Batch & 23.18 \\
GPT-5 & OpenAI & closed & 2025-08-07 & medium & OpenAI Batch & 26.94 \\
o3 & OpenAI & closed & 2025-04-16 & medium & OpenAI Batch & 12.10 \\
GPT-4o (2024-08-06) & OpenAI & closed & 2024-08-06 & none & OpenAI Batch & 6.70 \\
\addlinespace[3pt]
Claude Fable 5.2 & Anthropic & closed & 2026-11-09 & high & Claude Code & 195.31 \\
Claude Haiku 5.5 & Anthropic & closed & 2026-10-07 & medium & Claude Code & 2.57 \\
Claude Sonnet 5.5 & Anthropic & closed & 2026-09-28 & high & Anthropic Batch & 13.40 \\
Claude Opus 5.5 & Anthropic & closed & 2026-09-22 & medium & Claude Code & 54.51 \\
Claude Fable 5.1 & Anthropic & closed & 2026-09-01 & high & Claude Code & SUB\textsuperscript{c} \\
Claude Opus 5 & Anthropic & closed & 2026-07-24 & high & Claude Code & SUB\textsuperscript{c} \\
Claude Sonnet 5 & Anthropic & closed & 2026-06-30 & high & Anthropic Batch & 23.90 \\
Claude Fable 5 & Anthropic & closed & 2026-06-09 & high & Claude Code & 222.19 \\
Claude Opus 4.8 & Anthropic & closed & 2026-05-28 & high & Claude Code & SUB\textsuperscript{c} \\
Claude Opus 4.7 & Anthropic & closed & 2026-04-16 & high & Anthropic Batch & 32.40 \\
Claude Sonnet 4.6 & Anthropic & closed & 2026-02-17 & high & Anthropic Batch & 33.60 \\
Claude Opus 4.6 & Anthropic & closed & 2026-02-05 & high & Anthropic Batch & 42.20 \\
Claude Opus 4.5 & Anthropic & closed & 2025-11-24 & high & Anthropic Batch & 37.10 \\
Claude Haiku 4.5 & Anthropic & closed & 2025-10-15 & budget 8,192 & Anthropic Batch & 13.71 \\
Claude Sonnet 4.5 & Anthropic & closed & 2025-09-29 & budget 8,192 & Anthropic Batch & 32.34 \\
\addlinespace[3pt]
Kimi K3.1 Pro & Moonshot & open & 2026-10-23 & max & OpenRouter\textsuperscript{b} & 83.30 \\
Kimi K2 (0711) & Moonshot & open & 2025-07-11 & none & OpenRouter\textsuperscript{b} & 8.75 \\
\addlinespace[3pt]
DeepSeek V4.1 Flash 1017 & DeepSeek & open & 2026-10-17 & high & OpenRouter\textsuperscript{b} & 2.39 \\
DeepSeek V4.1 Flash (Sept build) & DeepSeek & open & 2026-09-10 & high & OpenRouter\textsuperscript{b} & 2.00 \\
DeepSeek V4 Pro 0813 & DeepSeek & open & 2026-08-13 & high & OpenRouter\textsuperscript{b} & 41.30 \\
DeepSeek V4 Flash 0731 & DeepSeek & open & 2026-07-31 & high & OpenRouter\textsuperscript{b} & 4.24 \\
DeepSeek R1-0528 & DeepSeek & open & 2025-05-28 & thinking on & OpenRouter\textsuperscript{b} & 6.22 \\
DeepSeek V3 & DeepSeek & open & 2024-12-26 & none & OpenRouter\textsuperscript{b} & 1.27 \\
\addlinespace[3pt]
Beam 1.0 (501B-A23B) & Reflection & open & 2026-10-16 & medium & OpenRouter\textsuperscript{b} & 5.59 \\
\addlinespace[3pt]
Nemotron 3.5 & NVIDIA & open & 2026-10-15 & thinking on & OpenRouter\textsuperscript{b} & 4.19 \\
\addlinespace[3pt]
Meta Muse Spark 1.3 & Meta & closed & 2026-09-02 & high & OpenRouter\textsuperscript{b} & 36.00 \\
Meta Muse Spark 1.2 Contributor & Meta & closed & 2026-08-05 & high & OpenRouter\textsuperscript{b} & 1.39 \\
\addlinespace[3pt]
GLM-5.3 & Z.ai & open & 2026-08-18 & max & OpenRouter\textsuperscript{b} & 33.90 \\
GLM-5.1 & Z.ai & open & 2026-04-07 & thinking on & OpenRouter\textsuperscript{b} & 28.20 \\
GLM-5 & Z.ai & open & 2026-02-12 & thinking on & OpenRouter\textsuperscript{b} & 21.20 \\
GLM-4.7 & Z.ai & open & 2025-12-22 & thinking on & OpenRouter\textsuperscript{b} & 13.10 \\
GLM-4.6 & Z.ai & open & 2025-09-30 & thinking on & OpenRouter\textsuperscript{b} & 13.80 \\
\addlinespace[3pt]
Qwen3.8 27B xhigh & Qwen & open & 2026-08-14 & xhigh & Self-hosted (vLLM) & 10.92 \\
Qwen3.8 2.4T-A95B & Qwen & open & 2026-08-12 & xhigh & OpenRouter\textsuperscript{b} & 89.00 \\
Qwen3.6 27B & Qwen & open & 2026-04-22 & thinking on & Self-hosted (vLLM) & 5.02 \\
Qwen3.5 27B & Qwen & open & 2026-02-24 & thinking on & Self-hosted (vLLM) & 9.15 \\
Qwen3.5 397B-A17B & Qwen & open & 2026-02-16 & thinking on & OpenRouter\textsuperscript{b} & 17.70 \\
Qwen3-235B-A22B Thinking 2507 & Qwen & open & 2025-07-25 & thinking on & OpenRouter\textsuperscript{b} & 16.20 \\
Qwen3 32B & Qwen & open & 2025-04-29 & thinking on & Self-hosted (vLLM) & 1.50 \\
QwQ-32B-Preview & Qwen & open & 2024-11-28 & thinking on & Self-hosted (vLLM) & -- \\
Qwen2.5-72B Instruct & Qwen & open & 2024-09-19 & none & OpenRouter\textsuperscript{b} & 1.33 \\
\addlinespace[3pt]
MiniMax M3 & MiniMax & open & 2026-06-01 & thinking on & OpenRouter\textsuperscript{b} & 19.60 \\
\addlinespace[3pt]
Hermes 3 405B & Nous Research & open & 2024-08-15 & none & OpenRouter\textsuperscript{b} & 3.20 \\
\addlinespace[3pt]
Mistral Large 2 (2407) & Mistral & open & 2024-07-24 & none & Self-hosted (vLLM) & -- \\
\addlinespace[3pt]
Llama 3.1 70B Instruct & Meta & open & 2024-07-23 & none & OpenRouter\textsuperscript{b} & 1.18 \\
\midrule
Pangram 4.1\textsuperscript{a} & Pangram Labs & closed & -- & -- & Pangram bulk API & -- \\
ProseLens & UMD & open & -- & -- & Self-hosted & -- \\
IdeaLens & UMD & open & -- & -- & Self-hosted & -- \\
\bottomrule
\end{longtable}}

\textsuperscript{a}\,Every Pangram result uses Pangram~4.1, through its bulk API.

\textsuperscript{b}\,We use the following OpenRouter providers: DeepInfra (Llama 3.1 70B Instruct, Hermes 3 405B, Qwen2.5-72B Instruct, DeepSeek V3, GLM-5.3); Novita (Kimi K2 (0711), GLM-4.6, GLM-4.7, DeepSeek V4 Pro 0813); Z.ai (GLM-5, GLM-5.1); Alibaba (Qwen3.5 397B-A17B); Meta (Meta Muse Spark 1.2 Contributor, Meta Muse Spark 1.3); SiliconFlow (DeepSeek R1-0528); Google AI Studio (Gemini 2.5 Pro); Venice (Qwen3-235B-A22B Thinking 2507); MiniMax (MiniMax M3); Moonshot AI (Kimi K3.1 Pro); Nebius (Beam 1.0 (501B-A23B)); Crusoe (Nemotron 3.5); Morph (DeepSeek V4.1 Flash 1017); Sail Research (DeepSeek V4.1 Flash (Sept build)); Novita, Alibaba and Together (Qwen3.8 2.4T-A95B); Parasail, Baidu, BaseTen, SiliconFlow, Mancer and DeepInfra (DeepSeek V4 Flash 0731).

\textsuperscript{c}\,We evaluate some Claude models through a subscription plan (\S\ref{sec:ai_vs_human} data). Every Claude Code call runs in print mode with all tools disabled (so no web search or fetch), no MCP servers or user settings, and our prompt in place of Claude Code's own system prompt.
\par

\section{Human or AI: who wrote the prose, who had the ideas?}
\label{app:rq1}

In this section, we discuss data, additional results, and analysis related to \S\ref{sec:ai_vs_human}.

\subsection{Human vs AI data}
\label{app:rq1_data}
The RQ1 corpus pools 37,575 documents from 24 public datasets. We leave out 694 documents whose source labels do not place them in a single case, and 454 that at least one of the ten full-corpus models did not answer (refusals, errors, or replies that could not be parsed). The remaining 36,427 documents are 11,689 fully human, 16,149 fully AI, 8,539 with human ideas in AI prose, and 50 with AI ideas in human prose. \autoref{tab:rq1_datasets} lists the 24 datasets, and \autoref{tab:atrey_detectors} lists the models.

For false-positive experiments, we exclude 2,002 fully human documents on suspicion of machine translation confounding some non-English documents. This creates a 9,687-document English human set. We apply the same principles to the timeline experiment, removing 140 documents from our initial 1,000-document sample, creating a final set of 860 documents: 229 fully human, 287 fully AI, 294 with human ideas in AI prose, and 50 with AI ideas in human prose.

\begin{table}[H]
\centering\footnotesize
\caption{The 24 datasets of the RQ1 corpus: documents in each case after the drops, and before them.}
\label{tab:rq1_datasets}
\setlength{\tabcolsep}{3.5pt}
\begin{tabularx}{\textwidth}{@{}>{\raggedright\arraybackslash\hsize=1.5\hsize}X>{\raggedright\arraybackslash\hsize=.5\hsize}X rrrr rr@{}}
\toprule
 & & \multicolumn{4}{c}{\textbf{Documents by case}} & & \\
\cmidrule(lr){3-6}
Dataset & Domain & \begin{tabular}[b]{@{}r@{}}Fully\\human\end{tabular} & \begin{tabular}[b]{@{}r@{}}AI ideas in\\human prose\end{tabular} & \begin{tabular}[b]{@{}r@{}}Fully\\AI\end{tabular} & \begin{tabular}[b]{@{}r@{}}Human ideas\\in AI prose\end{tabular} & Total & \begin{tabular}[b]{@{}r@{}}Before\\drops\end{tabular} \\
\midrule
Total & & 11,689 & 50 & 16,149 & 8,539 & 36,427 & 37,575 \\
\midrule
AcademicIntegrity \mbox{\citep{Karr2026WhyAD}} & paper abstracts & 300 & 0 & 0 & 1,307 & 1,607 & 1,624 \\
AI in Peer Review \mbox{\citep{saha2026policies}} & peer reviews & 659 & 0 & 1,519 & 1,322 & 3,500 & 3,612 \\
AI Peer Review Detection \mbox{\citep{ICLR2026_64592277}} & peer reviews & 662 & 0 & 661 & 0 & 1,323 & 1,326 \\
AI-Researcher (papers) \mbox{\citep{si2025can,si2026the}} & research papers & 19 & 0 & 0 & 0 & 19 & 43 \\
AI-Researcher (proposals and reviews) \mbox{\citep{si2025can,si2026the}} & research proposals, expert reviews & 467 & 0 & 79 & 49 & 595 & 600 \\
ARB \mbox{\citep{perrone2026arbmatchedauthorshiprewritingbenchmark}} & web text, news, short stories & 657 & 0 & 1,323 & 662 & 2,642 & 2,649 \\
C4 \mbox{\citep{c4dataset}} & web pages & 652 & 0 & 0 & 0 & 652 & 663 \\
CoCoNUTS \mbox{\citep{chen-etal-2026-coconuts}} & peer reviews & 654 & 0 & 662 & 660 & 1,976 & 2,134 \\
DetectionAI \mbox{\citep{jabarian2025artificial}} & news, fiction, blogs, r\'esum\'es, reviews & 663 & 0 & 1,321 & 0 & 1,984 & 1,989 \\
ELLIPSE \mbox{\citep{crossley2023ellipse}} & essays by English learners & 657 & 0 & 0 & 0 & 657 & 663 \\
Epoch author imitation \mbox{\citep{lee2026ai}} & blogs, fiction, science writing & 492 & 0 & 297 & 0 & 789 & 942 \\
GEDE \mbox{\citep{Gehring2025LimitsOL}} & student essays & 659 & 0 & 1,985 & 659 & 3,303 & 3,313 \\
GEN \mbox{\citep{shan2026stylometric}} & Wikipedia, WikiHow, arXiv, stories & 660 & 0 & 604 & 571 & 1,835 & 2,139 \\
HART \mbox{\citep{Bao2025DecouplingCA}} & news (several languages), essays, creative writing, arXiv & 1,322 & 0 & 2,637 & 1,320 & 5,279 & 5,302 \\
IdeaShift \mbox{\citep{rajendhran2026idealensdetectingaiideas}} & web pages & 40 & 0 & 80 & 38 & 158 & 280 \\
IdeaShift-X \mbox{\citep{rajendhran2026idealensdetectingaiideas}} & web pages in 24 languages & 642 & 0 & 658 & 650 & 1,950 & 1,989 \\
LAMP \mbox{\citep{LAMP}} & fiction, creative nonfiction & 0 & 0 & 662 & 0 & 662 & 663 \\
MELD \mbox{\citep{li2026meldmultitaskequilibratedlearning}} & news, books, Wikipedia, Reddit, poetry, recipes, reviews, abstracts & 659 & 0 & 1,227 & 0 & 1,886 & 1,986 \\
OpAI-Bench \mbox{\citep{Bsharat2026OperationGuidedPH}} & reports, essays, news, abstracts & 523 & 0 & 0 & 643 & 1,166 & 1,173 \\
\citet{perkins2024simple} & academic writing & 0 & 0 & 89 & 0 & 89 & 89 \\
Sem-Detect \mbox{\citep{duarte2026semdetect}} & peer reviews & 657 & 0 & 661 & 658 & 1,976 & 1,985 \\
StoryScope \mbox{\citep{russell2026storyscope}} & short fiction & 645 & 0 & 655 & 0 & 1,300 & 1,326 \\
StoryScope (newer generators) \mbox{\citep{russell2026storyscope}} & short fiction & 0 & 0 & 980 & 0 & 980 & 985 \\
TwiceTold \mbox{\citep{rajendhran2026idealensdetectingaiideas}} & short fiction & 0 & 50 & 49 & 0 & 99 & 100 \\
\bottomrule
\end{tabularx}
\end{table}

\subsection{Additional results}
\label{app:rq1_results}

\subsubsection{Full corpus}
\label{app:atrey_rq1_full}
\begin{figure}[H]
\centering
\begin{minipage}[t]{.48\textwidth}\centering\includegraphics[width=\textwidth]{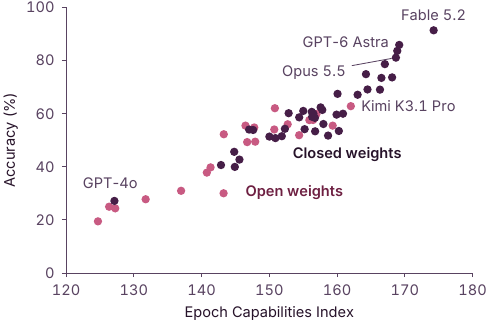}
\caption{Detection accuracy (prose and ideas both correct) is highly correlated with general capabilities measured in the Epoch Capabilities Index ($r=0.94$).}
\label{fig:atrey_rq1_capability}\end{minipage}\hfill
\begin{minipage}[t]{.48\textwidth}\centering\includegraphics{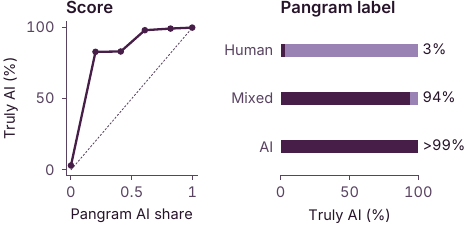}
\caption{Pangram~4.1 is rarely wrong when it reports AI: share of documents with AI prose by its AI score (left) and by its label (right), leaving out human ideas in AI prose}.
\label{fig:atrey_rq1_pangram_calibration}\end{minipage}
\end{figure}

For the ten models run over the full corpus, \autoref{tab:atrey_rq1_full} shows the accuracy splits and \autoref{tab:atrey_rq1_operating} shows false-positive rates and recall at matched false-positive rates. Additionally, we show general capability trends in \autoref{fig:atrey_rq1_capability} and Pangram~4.1's calibration in \autoref{fig:atrey_rq1_pangram_calibration}.

\begin{table}[H]
\centering\footnotesize
\caption{RQ1 accuracy (\%) on the full corpus. \emph{Avg.} is an average of the four cases; \emph{All} is accuracy over all documents. \textsuperscript{\dag}Pangram~4.1 is not scored on human ideas in AI prose; its average is over the other three cases.}
\label{tab:atrey_rq1_full}
\setlength{\tabcolsep}{3.4pt}
\providecommand{\quadIcon}[1]{\labicon{#1}\,}\providecommand{\quadNoIcon}{\hphantom{\labicon{anthropic}\,}}
\begin{tabular}{l rrrrr rrrrr rr}
\toprule
 & \multicolumn{5}{c}{\textbf{\faPen\ Prose correct}} & \multicolumn{5}{c}{\textbf{\faLightbulb\ Ideas correct}} & \multicolumn{2}{c}{\textbf{Both}} \\
\cmidrule(lr){2-6}\cmidrule(lr){7-11}\cmidrule(lr){12-13}
\faPen\ \textit{Prose by} & Human & Human & AI & AI & & Human & Human & AI & AI & & & \\
\faLightbulb\ \textit{Ideas by} & Human & AI & AI & Human & Avg. & Human & AI & AI & Human & Avg. & All & Avg. \\
\midrule
\multicolumn{13}{l}{\textbf{Prompted LLMs}} \\
\quad \iconSol\,\textcolor{inkSol}{GPT-6.1 Sol} & 99.0 & 96.0 & 98.9 & 83.3 & 94.3 & 99.8 & 72.0 & 92.5 & 96.6 & 90.2 & 91.5 & 84.8 \\
\quad \iconOpus\,\textcolor{inkOpus}{Claude Opus 5.5} & 99.3 & 100.0 & 98.6 & 82.4 & 95.0 & 99.8 & 64.0 & 93.2 & 97.7 & 88.7 & 91.8 & 84.0 \\
\quad \iconCarbon\,\textcolor{inkCarbon}{Gemini 4 Carbon} & 99.7 & 100.0 & 97.3 & 81.3 & 94.6 & 99.9 & 62.0 & 94.5 & 98.0 & 88.6 & 92.1 & 83.6 \\
\quad \quadIcon{anthropic}Claude Sonnet 5.5 & 99.4 & 100.0 & 95.4 & 69.2 & 91.0 & 99.8 & 36.0 & 91.5 & 93.0 & 80.1 & 86.6 & 72.0 \\
\quad \quadIcon{openai}GPT-6 Luna & 97.1 & 52.0 & 95.6 & 61.1 & 76.5 & 98.3 & 72.0 & 93.0 & 70.6 & 83.5 & 79.6 & 61.4 \\
\quad \quadIcon{deepseek}DeepSeek V4.1 Flash & 96.8 & 70.0 & 93.4 & 62.5 & 80.7 & 98.5 & 64.0 & 86.7 & 76.3 & 81.3 & 78.2 & 63.9 \\
\quad \quadIcon{google}Gemini 3.8 Flash & 99.2 & 90.0 & 90.2 & 38.5 & 79.5 & 99.8 & 10.0 & 86.5 & 88.0 & 71.1 & 76.3 & 53.0 \\
\quad \quadIcon{anthropic}Claude Haiku 5.5 & 96.9 & 72.0 & 89.5 & 52.3 & 77.7 & 97.2 & 34.0 & 76.0 & 81.3 & 72.1 & 71.7 & 55.1 \\
\quad \quadIcon{google}Gemini 3.5 Flash-Lite & 81.8 & 2.0 & 91.1 & 68.3 & 60.8 & 85.5 & 98.0 & 86.4 & 48.9 & 79.7 & 68.2 & 46.2 \\
\quad \quadNoIcon Nemotron 3.5 & 95.8 & 58.0 & 56.0 & 25.8 & 58.9 & 99.0 & 12.0 & 24.8 & 95.1 & 57.7 & 46.2 & 35.6 \\
\midrule
\multicolumn{13}{l}{\textbf{Trained detectors}} \\
\quad \quadNoIcon Pangram 4.1 & 99.9 & 96.0 & 96.7 & -- & 97.6\textsuperscript{\dag} & -- & -- & -- & -- & -- & -- & -- \\
\quad \quadNoIcon ProseLens & 99.9 & 100.0 & 93.9 & 45.7 & 84.9 & -- & -- & -- & -- & -- & -- & -- \\
\quad \quadNoIcon IdeaLens & -- & -- & -- & -- & -- & 99.7 & 68.0 & 85.0 & 95.5 & 87.1 & -- & -- \\
\bottomrule
\end{tabular}
\end{table}

\begin{table}[H]
\centering\footnotesize
\caption{Frontier models have few false positives, where a fully human document is misclassified as AI. Recall is defined as the share of AI text flagged as AI (in \%). For prompted models, we classify a document as AI if its reported AI share is at least 0.5. For trained detectors, we use their pre-existing cutoffs. @1\% is the recall such that the FPR is at or below 1\%. \textsuperscript{\dag}Pangram~4.1's recall is on fully AI documents only, since it is not scored on human ideas in AI prose. \textsuperscript{\ddag}Gemini~4 Neon is run on the fully human documents only.}
\label{tab:atrey_rq1_operating}
\setlength{\tabcolsep}{3.5pt}
\providecommand{\quadIcon}[1]{\labicon{#1}\,}\providecommand{\quadNoIcon}{\hphantom{\labicon{anthropic}\,}}
\begin{tabular}{l rrr rrr}
\toprule
 & \multicolumn{3}{c}{\textbf{\faPen\ Prose}} & \multicolumn{3}{c}{\textbf{\faLightbulb\ Ideas}} \\
\cmidrule(lr){2-4}\cmidrule(lr){5-7}
System & FPR & Recall & @1\% & FPR & Recall & @1\% \\
\midrule
\multicolumn{7}{l}{\textbf{Prompted LLMs}} \\
\quad \iconOpus\,\textcolor{inkOpus}{Claude Opus 5.5} & 0.10 & 93.0 & \textbf{96.1} & 0.07 & 93.1 & \textbf{99.4} \\
\quad \iconSol\,\textcolor{inkSol}{GPT-6.1 Sol} & 0.10 & \textbf{93.5} & 95.3 & 0.05 & 92.4 & 97.8 \\
\quad \iconCarbon\,\textcolor{inkCarbon}{Gemini 4 Carbon} & 0.01 & 91.8 & 94.6 & \textbf{0.00} & \textbf{94.4} & 97.4 \\
\quad \quadIcon{anthropic}Claude Sonnet 5.5 & 0.09 & 86.4 & 92.5 & 0.07 & 91.4 & 97.0 \\
\quad \quadIcon{deepseek}DeepSeek V4.1 Flash & 2.15 & 82.7 & 62.0 & 1.49 & 86.6 & 80.9 \\
\quad \quadIcon{google}Gemini 4 Neon\textsuperscript{\ddag} & 0.03 & -- & -- & \textbf{0.00} & -- & -- \\
\quad \quadIcon{openai}GPT-6 Luna & 1.81 & 83.7 & 76.2 & 1.55 & 93.0 & 88.7 \\
\midrule
\multicolumn{7}{l}{\textbf{Trained detectors}} \\
\quad \quadNoIcon Pangram 4.1 & \textbf{0.00} & 96.7\textsuperscript{\dag} & -- & -- & -- & -- \\
\quad \quadNoIcon ProseLens & 0.03 & 77.2 & -- & -- & -- & -- \\
\quad \quadNoIcon IdeaLens & -- & -- & -- & 0.21 & 85.0 & -- \\
\bottomrule
\end{tabular}
\end{table}

\subsubsection{Timeline of detectors}
\label{app:atrey_rq1_timeline}

\begin{figure*}[!htbp]
  \centering
  \begin{minipage}[t]{.49\textwidth}
    \vspace{0pt}\centering
    \includegraphics{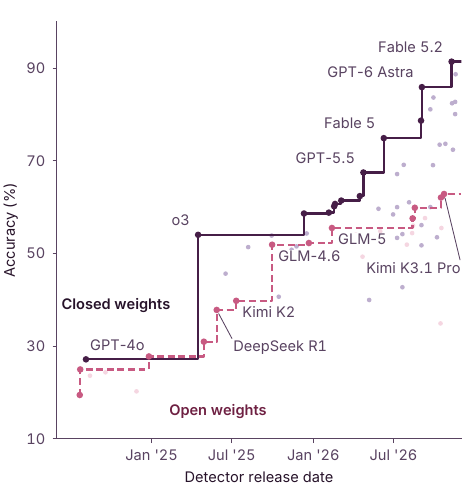}
  \end{minipage}\hfill
  \begin{minipage}[t]{.49\textwidth}
    \vspace{0pt}\centering
    \includegraphics{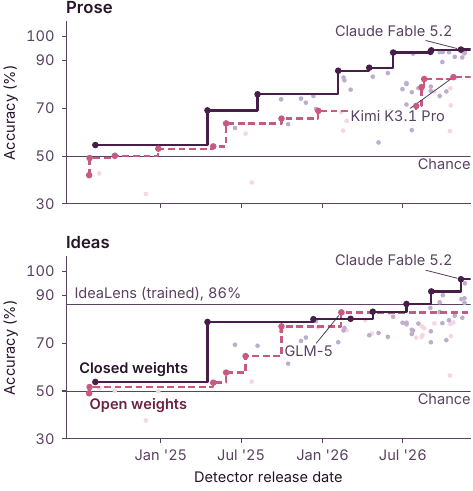}
  \end{minipage}
  \caption{Full-history versions of the \S\ref{sec:ai_vs_human} model timeline, including all 72 models and all four provenance cases. Left: both prose and ideas must be correct. Right: prose and ideas scored separately, with IdeaLens's decision rule as a reference line; Pangram~4.1, which is not scored on human ideas in AI prose, has none here. Accuracy weights the four cases equally, on the 860 retained documents from the 1,000-document sample%
  Lines mark the best accuracy attained by release date within closed-weight (solid purple) and open-weight (dashed pink) models.}
  \label{fig:rq1_frontier_full_views}
\end{figure*}

{\footnotesize\setlength{\tabcolsep}{4pt}\setlength{\LTcapwidth}{\textwidth}
\begin{longtable}{l r r r r r r}
\caption{All 72 detectors on the 860-document sample, in release order (release dates and settings are in \autoref{tab:atrey_detectors}). \emph{Both} requires the prose and ideas labels to be right; \emph{avg.}\ averages it over the four cases, while Prose and Ideas are calculated over all documents. Fully human documents leave out those excluded from false-positive rates (\S\ref{app:rq1_data}). Errors and unparsable verdicts count as wrong. \textsuperscript{\tabdagger}\,partial run; \textsuperscript{\UseTextSymbol{TS1}{\textdaggerdbl}}\,two refusals excluded.}\label{tab:rq1_timeline}\\
\toprule
Detector & \textit{n} & Prose & Ideas & Both & Both (avg.) & \shortstack[r]{Both, human\\ideas in AI prose} \\
\midrule
\endfirsthead
\toprule
Detector & \textit{n} & Prose & Ideas & Both & Both (avg.) & \shortstack[r]{Both, human\\ideas in AI prose} \\
\midrule
\endhead
Llama 3.1 70B Instruct & 860 & 30.5 & 58.3 & 21.1 & 19.5 & 9.2 \\
Mistral Large 2 (2407)\textsuperscript{\tabdagger} & 794 & 34.5 & 60.5 & 26.3 & 25.0 & 2.1 \\
GPT-4o (2024-08-06) & 860 & 55.6 & 53.4 & 31.7 & 27.2 & 14.3 \\
Hermes 3 405B & 860 & 32.0 & 60.2 & 25.8 & 23.6 & 7.1 \\
Qwen2.5-72B Instruct & 860 & 33.7 & 60.5 & 26.4 & 24.4 & 5.8 \\
QwQ-32B-Preview & 860 & 24.8 & 40.5 & 18.4 & 20.3 & 4.1 \\
DeepSeek V3 & 860 & 52.4 & 59.1 & 32.6 & 27.8 & 29.2 \\
o3 & 860 & 76.4 & 77.2 & 59.5 & 54.0 & 32.0 \\
Qwen3 32B & 849 & 63.6 & 59.4 & 39.6 & 31.0 & 58.4 \\
DeepSeek R1-0528 & 860 & 64.3 & 65.0 & 44.3 & 37.8 & 53.1 \\
Gemini 2.5 Pro & 860 & 72.4 & 72.7 & 55.6 & 45.7 & 41.5 \\
Kimi K2 (0711) & 860 & 60.9 & 71.6 & 46.5 & 39.8 & 21.1 \\
Qwen3-235B-A22B Thinking 2507\textsuperscript{\tabdagger} & 858 & 40.0 & 47.9 & 30.8 & 30.1 & 14.3 \\
GPT-5 & 860 & 71.2 & 76.5 & 55.6 & 51.4 & 23.8 \\
Claude Sonnet 4.5 & 860 & 79.4 & 78.7 & 62.9 & 53.9 & 51.0 \\
GLM-4.6 & 860 & 73.0 & 77.7 & 59.1 & 51.9 & 30.6 \\
Claude Haiku 4.5 & 860 & 64.4 & 69.2 & 46.9 & 40.7 & 25.2 \\
GPT-5.1 & 860 & 68.6 & 77.0 & 53.0 & 50.8 & 16.3 \\
Claude Opus 4.5 & 860 & 74.9 & 78.8 & 60.2 & 51.5 & 24.8 \\
GPT-5.2 & 860 & 81.7 & 80.3 & 65.8 & 58.6 & 33.3 \\
Gemini 3 Flash & 860 & 82.2 & 79.8 & 65.2 & 54.3 & 27.6 \\
GLM-4.7 & 860 & 75.1 & 78.7 & 61.2 & 52.3 & 33.0 \\
Claude Opus 4.6 & 860 & 84.4 & 80.6 & 68.8 & 58.8 & 36.0 \\
GLM-5 & 860 & 75.7 & 83.6 & 64.5 & 55.5 & 35.7 \\
Qwen3.5 397B-A17B\textsuperscript{\tabdagger} & 860 & 68.1 & 80.5 & 55.4 & 49.5 & 15.0 \\
Claude Sonnet 4.6 & 860 & 84.5 & 84.0 & 70.8 & 60.2 & 50.7 \\
Gemini 3.1 Pro & 860 & 81.7 & 85.9 & 71.4 & 60.7 & 48.0 \\
Qwen3.5 27B & 860 & 70.9 & 77.1 & 56.4 & 48.5 & 25.2 \\
GPT-5.4 & 860 & 81.0 & 84.4 & 69.2 & 61.4 & 45.6 \\
GLM-5.1\textsuperscript{\tabdagger} & 859 & 75.2 & 81.5 & 62.8 & 54.1 & 35.8 \\
Claude Opus 4.7 & 860 & 84.4 & 84.7 & 72.1 & 62.4 & 59.5 \\
Qwen3.6 27B & 860 & 67.9 & 78.5 & 54.0 & 49.3 & 20.8 \\
GPT-5.5 & 860 & 88.4 & 86.0 & 76.2 & 67.5 & 54.1 \\
Gemini 3.1 Flash-Lite & 860 & 67.0 & 71.3 & 48.7 & 40.0 & 22.4 \\
Claude Opus 4.8\textsuperscript{\tabdagger} & 858 & 80.8 & 85.9 & 69.9 & 59.7 & 44.9 \\
MiniMax M3 & 860 & 81.6 & 82.1 & 66.7 & 54.8 & 37.4 \\
Claude Fable 5\textsuperscript{\tabdagger} & 852 & 90.7 & 91.7 & 83.3 & 74.9 & 72.0 \\
Claude Sonnet 5 & 860 & 81.9 & 85.5 & 69.5 & 58.4 & 46.6 \\
GPT-5.6 Luna & 860 & 80.7 & 81.9 & 65.1 & 53.4 & 31.6 \\
GPT-5.6 Sol & 860 & 86.4 & 90.5 & 77.9 & 67.2 & 57.5 \\
GPT-5.6 Terra & 860 & 84.8 & 85.5 & 71.6 & 60.0 & 42.5 \\
Gemini 3.5 Flash-Lite & 860 & 72.3 & 71.3 & 51.9 & 42.7 & 15.0 \\
Gemini 3.6 Flash & 860 & 79.5 & 82.0 & 64.2 & 54.2 & 23.1 \\
Claude Opus 5\textsuperscript{\tabdagger} & 857 & 90.1 & 90.5 & 82.3 & 69.1 & 71.3 \\
DeepSeek V4 Flash 0731\textsuperscript{\tabdagger} & 859 & 75.3 & 74.7 & 59.1 & 51.9 & 20.8 \\
Meta Muse Spark 1.2 Contributor & 860 & 80.3 & 77.1 & 63.0 & 61.1 & 24.8 \\
Qwen3.8 2.4T-A95B\textsuperscript{\tabdagger} & 854 & 75.1 & 82.5 & 61.7 & 54.4 & 28.7 \\
DeepSeek V4 Pro 0813\textsuperscript{\tabdagger} & 859 & 78.8 & 80.7 & 62.8 & 57.6 & 21.4 \\
Qwen3.8 27B xhigh & 860 & 78.0 & 82.0 & 64.0 & 56.0 & 35.0 \\
GLM-5.3 & 860 & 80.2 & 85.1 & 67.4 & 59.9 & 30.9 \\
Claude Fable 5.1\textsuperscript{\tabdagger} & 853 & 91.1 & 92.7 & 84.5 & 78.7 & 73.4 \\
Gemini 3.8 Flash & 860 & 74.2 & 85.1 & 62.6 & 51.8 & 21.1 \\
Meta Muse Spark 1.3 & 860 & 75.1 & 81.9 & 62.4 & 56.1 & 28.9 \\
\textcolor{inkAstra}{GPT-6 Astra} & 860 & 92.6 & 94.2 & 87.1 & 85.9 & 76.5 \\
DeepSeek V4.1 Flash (Sept build) & 860 & 80.7 & 80.2 & 64.1 & 57.6 & 22.4 \\
\textcolor{inkOpus}{Claude Opus 5.5} & 860 & 90.9 & 92.6 & 84.5 & 81.1 & 73.5 \\
GPT-6 Luna & 860 & 79.8 & 84.9 & 67.2 & 60.0 & 28.2 \\
Claude Sonnet 5.5 & 860 & 85.9 & 90.2 & 78.3 & 69.1 & 59.9 \\
\textcolor{inkSol}{GPT-6.1 Sol} & 860 & 92.2 & 94.3 & 86.9 & 83.7 & 76.2 \\
Claude Haiku 5.5 & 860 & 76.9 & 79.8 & 60.9 & 53.5 & 30.3 \\
\textcolor{inkArgon}{Gemini 4 Argon} & 860 & 90.9 & 92.0 & 83.7 & 73.5 & 74.8 \\
Nemotron 3.5 & 860 & 58.4 & 66.2 & 39.8 & 34.9 & 21.8 \\
Beam 1.0 (501B-A23B) & 860 & 77.0 & 76.7 & 60.4 & 62.1 & 24.1 \\
DeepSeek V4.1 Flash 1017 & 860 & 74.2 & 76.0 & 61.4 & 55.5 & 28.9 \\
Kimi K3.1 Pro\textsuperscript{\tabdagger} & 860 & 83.6 & 87.9 & 73.3 & 62.9 & 51.4 \\
GPT-6.1 Terra & 860 & 89.4 & 88.8 & 79.4 & 73.7 & 55.8 \\
\textcolor{inkFable}{Claude Fable 5.2}\textsuperscript{\UseTextSymbol{TS1}{\textdaggerdbl}} & 858 & 92.4 & 96.0 & 88.9 & 91.4 & 81.2 \\
\textcolor{inkCarbon}{Gemini 4 Carbon} & 860 & 91.4 & 94.9 & 86.9 & 82.4 & 76.5 \\
Gemini 4 Hydrogen & 860 & 88.5 & 91.6 & 81.0 & 72.4 & 63.9 \\
GPT-6.2 Astra & 860 & 91.0 & 93.7 & 85.6 & 82.7 & 74.8 \\
GPT-6.2 Sol & 860 & 90.9 & 92.8 & 84.3 & 80.1 & 72.1 \\
Gemini 4 Neon & 860 & 91.2 & 96.5 & 88.0 & 88.7 & 76.2 \\
\bottomrule
\end{longtable}}

\subsection{Analysis}
\label{app:rq1_analysis}

\subsubsection{Prompt development}
\label{app:rq1_prompt_dev}

\autoref{tab:prompt_dev} shows GPT-6.1 Sol's accuracy under four prompts of increasing structure, from a bare question (P1) to the final joint prompt (P4); the final prompt is in our repository, linked in the abstract.

Our final prompt asks about prose and ideas in a single call, and asks the detector to reason about how it weighs competing evidence and which factors decided its verdict. From P1 to P4, Sol's idea accuracy rises from 85.0\% to 97.0\%, mostly because its false-positive rate falls from 22.9\% to 2.2\%, and its idea accuracy on human ideas in AI prose rises from 64.2\% to 96.7\%. Its prose accuracy changes little (88.3\% to 89.8\%).
\begin{table}[!htbp]
\centering\footnotesize%
\caption{Accuracy of GPT-6.1 Sol on $\sim$5.5k documents from 13 datasets scored with prompts of increasing complexity; a document counts as AI when its stated AI share is at least 0.5.}
\label{tab:prompt_dev}
\begin{tabular}{@{}lrrrr@{}}
\toprule
 & Overall (\%) & TPR (\%) & FPR (\%) & \shortstack[r]{Human ideas\\in AI prose (\%)} \\
\midrule
\multicolumn{5}{@{}l}{\textbf{Ideas}} \\
P1: bare question & 85.0 & 98.9 & 22.9 & 64.2 \\
P2: definitions added & 92.0 & 97.9 & 11.3 & 84.2 \\
P3: structured reasoning & 96.6 & 96.8 & 3.4 & 94.8 \\
P4: joint prose and ideas & 97.0 & 95.6 & 2.2 & 96.7 \\
\midrule
\multicolumn{5}{@{}l}{\textbf{Prose}} \\
P1: bare question & 88.3 & 85.7 & 3.4 & 73.1 \\
P2: definitions added & 88.8 & 86.5 & 3.8 & 74.9 \\
P3: structured reasoning & 88.5 & 85.9 & 3.2 & 73.7 \\
P4: joint prose and ideas & 89.8 & 87.7 & 3.3 & 77.3 \\
\bottomrule
\end{tabular}
\end{table}

\subsubsection{Contamination check}
\label{app:ty_contamination}

\paragraph{News published after every detector.}
We test five detectors on 1,000 US news articles and op-eds published on 10--11 November 2026: 188 that Pangram~4.1 labels AI and 812 it labels human (\autoref{tab:atrey_rq1_contamination}).

\begin{table}[H]
\centering\footnotesize
\caption{Detectors against Pangram~4.1 on 1,000 news articles and op-eds published after every detector\textquotesingle s release (\%). Recall: Pangram-AI articles flagged as AI; FPR: Pangram-human articles flagged as AI.}
\label{tab:atrey_rq1_contamination}
\providecommand{\quadIcon}[1]{\labicon{#1}\,}\providecommand{\quadNoIcon}{\hphantom{\labicon{anthropic}\,}}
\begin{tabular}{l rrr}
\toprule
Detector & Recall & FPR & Agreement \\
\midrule
\iconFable\,\textcolor{inkFable}{Claude Fable 5.2} (high) & 95.2 & 1.8 & 97.6 \\
\iconAstra\,\textcolor{inkAstra}{GPT-6 Astra} (medium) & 100.0 & 5.0 & 95.9 \\
\iconCarbon\,\textcolor{inkCarbon}{Gemini 4 Carbon} (medium) & 78.2 & 0.5 & 95.5 \\
\iconOpus\,\textcolor{inkOpus}{Claude Opus 5.5} (medium) & 77.1 & 3.4 & 92.9 \\
\quadIcon{deepseek}DeepSeek V4.1 Flash (max) & 75.0 & 20.6 & 78.6 \\
\bottomrule
\end{tabular}
\end{table}

\paragraph{Stories written after both detectors' training cutoffs.}
We also test Claude Opus~5.5 and GPT-6.1 Sol on 200 stories published or generated after both models' training cutoffs (\autoref{tab:rq1_fresh_fiction}).

\begin{table}[H]
\centering\footnotesize
\caption{Detectors predict authors of stories published or generated after their training cutoffs about as well as the full corpus's fiction. Accuracy (\%) on 200 stories (mean of five runs): 50 human stories published 1--16 October 2026 in 20 venues, 50 TwiceTold rewrites with AI ideas in human prose, and 25 stories by Claude Opus~5.5 and GPT-6.1 Sol in each of the two AI-prose cases, written from human stories.}
\label{tab:rq1_fresh_fiction}
\begin{tabular}{l rrr rrr}
\toprule
 & \multicolumn{3}{c}{Prose} & \multicolumn{3}{c}{Ideas} \\
\cmidrule(lr){2-4}\cmidrule(lr){5-7}
Detector & Post-cutoff & Full corpus & Gap & Post-cutoff & Full corpus & Gap \\
\midrule
\jOpus{Claude Opus~5.5} & 98.7 & 99.2 & \tabminus0.5 & 89.7 & 89.2 & +0.6 \\
\jSol{GPT-6.1 Sol} & 96.5 & 98.1 & \tabminus1.6 & 87.1 & 90.5 & \tabminus3.4 \\
\bottomrule
\end{tabular}
\end{table}

\paragraph{Effect of dates in texts.}
To test whether detectors rely on a document's dates, we replace every year in the 190 dated documents of the 860-document sample with 2026, leaving years in URLs and arXiv identifiers unchanged, and run each detector on both versions with the same prompt (\autoref{tab:rq1_year2026}).

\begin{table}[H]
\centering\footnotesize
\caption{Replacing every year in a document with 2026 does not change frontier models' predictions. Accuracy (\%) on the 190 documents of the 860-document sample that mention a year (76 fully human, 40 fully AI and 74 with human ideas in AI prose), with the original text and with every year set to 2026. \emph{Changed}: verdicts that switch to AI\,/\,to human when the years change. \emph{Rerun}: the same count between two runs of the unchanged text.}
\label{tab:rq1_year2026}
\begin{tabular}{l rrrr rrrr}
\toprule
 & \multicolumn{4}{c}{Ideas} & \multicolumn{4}{c}{Prose} \\
\cmidrule(lr){2-5}\cmidrule(lr){6-9}
Detector & Original & 2026 & Changed & Rerun & Original & 2026 & Changed & Rerun \\
\midrule
\jOpus{Claude Opus~5.5} & 97.9 & 97.9 & 2\,/\,0 & 2\,/\,1 & 90.5 & 89.5 & 1\,/\,3 & 1\,/\,3 \\
\jSol{GPT-6.1 Sol} & 98.9 & 97.9 & 1\,/\,1 & 2\,/\,0 & 90.0 & 89.5 & 0\,/\,1 & 0\,/\,2 \\
\jCarbon{Gemini~4 Carbon} & 98.4 & 98.4 & 1\,/\,0 & 0\,/\,1 & 88.9 & 90.0 & 1\,/\,0 & 1\,/\,1 \\
\jDeepSeek{DeepSeek~V4.1 Flash} & 84.7 & 69.5 & 29\,/\,1 & 8\,/\,3 & 68.9 & 71.6 & 18\,/\,4 & 9\,/\,6 \\
\bottomrule
\end{tabular}
\end{table}

\subsubsection{Explanation agreement}
\label{app:explanation-agreement}

\paragraph{Grouping.}
An LLM judge (Claude Opus~5.5) aligns two explanations of the same document, $E_A$ and $E_B$, separately for prose and ideas. Each explanation lists evidence items (a statement, quotes, a lean and its strength, and a justification), names a decisive factor and gives an AI share $s \in [0,1]$ (\autoref{fig:explanation-example}). 
In a first call, which hides leans, justifications, and verdicts, the judge groups items from the two explanations that make the same point, even when they quote different passages; every group holds at least one item from each explanation, and items that fit into no group stay unmatched. It also maps each point in the decisive factors to a matched group or an unmatched item. 
In a second call, the judge is provided each group's leans and justifications and decides whether both explanations rest on the same property of evidence in the group.

\paragraph{Metrics.}
Let $\mathcal{G}$ be the set of groups, $U_A$ and $U_B$ the set of unmatched items each from documents $A$ and $B$, and $D_A$ and $D_B$ the set of decisive factors mapped to a group or an unmatched item.
\begin{itemize}[leftmargin=*,itemsep=1pt,topsep=2pt]
  \item \textbf{Verdict} is 1 if both AI shares fall on the same side of 0.5.
  \item \textbf{Evidence F1} $= 2|\mathcal{G}|/(2|\mathcal{G}|+|U_A|+|U_B|)$, the harmonic mean of the share of each explanation's points that the other also makes (each group counts as one shared point).
  \item \textbf{Interpretation} is the share of groups whose items all have the same lean and the judge determines the justification of those items to be identical.
  \item \textbf{Decisive F1} $= 2|D_A \cap D_B|/(|D_A|+|D_B|)$, and 0 when exactly one set is empty.
\end{itemize}
Undefined values (both sets empty, or no groups for interpretation) are dropped for that metric only. We score one cross-model pair per document and average documents with equal weight; 95\% intervals use a percentile bootstrap over documents (2,000 resamples). Opus~5.5, Sol, and Gemini~4 Argon explanations are compared in pairs on 2,854 documents (\autoref{tab:explanation-agreement}).

\begin{figure}[!htbp]
\centering
\definecolor{imgframe}{HTML}{C0BDB7}%
\newcommand{\uihalf}[1]{%
  \sbox0{\includegraphics[width=\dimexpr(\textwidth-10pt)/2\relax,trim=#1,clip]{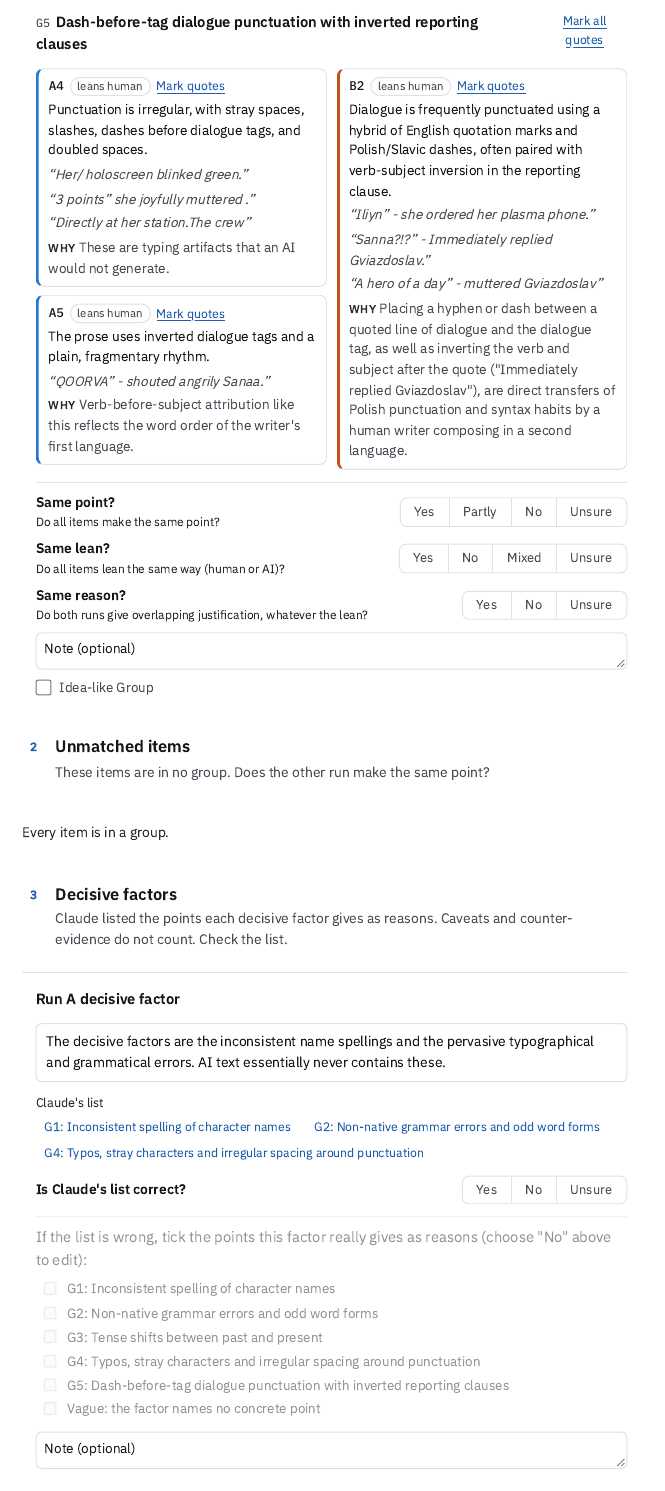}}%
  \begin{tikzpicture}[baseline={(0,\ht0)}]
  \begin{scope}\clip[rounded corners=3pt] (0,0) rectangle (\wd0,\ht0);\node[inner sep=0pt,anchor=south west] at (0,0) {\usebox0};\end{scope}
  \draw[imgframe,line width=0.5pt,rounded corners=3pt] (0,0) rectangle (\wd0,\ht0);
  \end{tikzpicture}}%
\uihalf{6 370.5 14 3}\hfill\uihalf{6 7 14 345}
\caption{Human annotation interface used to manually verify explanation agreement. For each group of evidence items from two detector explanations, annotators assess whether the items make the same point, lean the same way (human or AI), and state the same reason. Annotators further inspect whether unmatched items do not belong to any formed group, and check the correctness of the list of points made in the decisive-factor, which was formed by the judge.}
\label{fig:explanation-annotation-interface}
\end{figure}
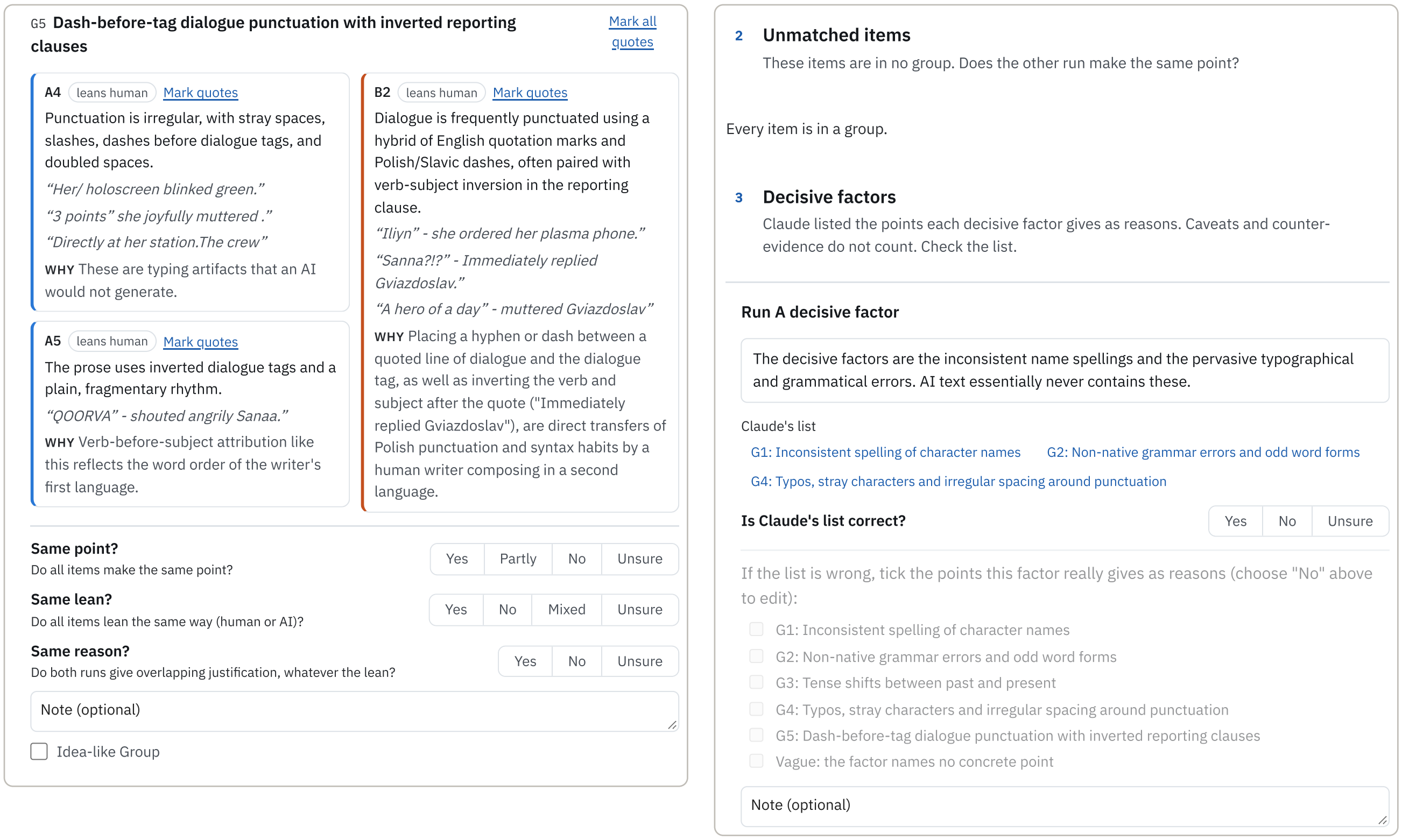

\paragraph{Judge validation.}
Two annotators audit the judge on 32 documents sampled evenly from the four cases. They annotate on one pair of explanations at a time, in the interface of \autoref{fig:explanation-annotation-interface}. They check whether each group makes the same point, whether its items share a justification, whether unmatched items should belong in a group, and whether the decisive factors are mapped correctly. Annotator~1 completed 32 documents and Annotator~2 completed 24. \autoref{tab:manual-agreement} reports Gwet's AC1~\citep{gwet2014handbook} and F1, giving each document equal weight; ``partly'' responses made by annotators count as no, and ``unsure'' responses are dropped. The annotators saw the groups and decisive sets formed by the judge, so humans validate the proposed judgments rather than suggesting new groupings from scratch.

\begin{table}[!htbp]
\centering\footnotesize
\caption{Every pair of detectors agrees on verdicts more than on evidence, interpretation, or decisive factors. A Claude Opus~5.5 judge compares an explanation from each detector of the pair (\S\ref{app:explanation-agreement}). Opus~5.5--GPT-6.1 Sol covers 2,854 documents (951 fully human, 950 fully AI, and 953 with human ideas in AI prose, from 24 datasets); the pairs involving Gemini 4 Argon cover 2,849 of them, as five have no usable Gemini 4 Argon explanation. Decisive F1 drops pairs in which neither explanation names a decisive point.}
\label{tab:explanation-agreement}\label{tab:opus-argon-agreement}\label{tab:argon-sol-agreement}
\setlength{\tabcolsep}{4pt}
\begin{tabular}{@{}llrrrrr@{}}
\toprule
Detectors & Case & Docs & Verdict & Evidence F1 & Interpretation & Decisive F1 \\
\midrule
\multicolumn{7}{@{}l}{\textbf{Ideas}} \\
\jOpus{Opus~5.5}--\jSol{GPT-6.1 Sol} & All & 2{,}854 & 0.98 & 0.85 & 0.68 & 0.56 \\
 & \quad Fully human & 951 & 0.99 & 0.83 & 0.67 & 0.57 \\
 & \quad Fully AI & 950 & 0.96 & 0.85 & 0.70 & 0.52 \\
 & \quad Human ideas, AI prose & 953 & 0.97 & 0.88 & 0.68 & 0.59 \\
\addlinespace
\jOpus{Opus~5.5}--\jArgon{Argon} & All & 2{,}849 & 0.98 & 0.92 & 0.76 & 0.64 \\
 & \quad Fully human & 947 & 1.00 & 0.92 & 0.77 & 0.67 \\
 & \quad Fully AI & 950 & 0.96 & 0.89 & 0.75 & 0.58 \\
 & \quad Human ideas, AI prose & 952 & 0.98 & 0.93 & 0.77 & 0.65 \\
\addlinespace
\jArgon{Argon}--\jSol{GPT-6.1 Sol} & All & 2{,}849 & 0.97 & 0.88 & 0.63 & 0.54 \\
 & \quad Fully human & 947 & 0.99 & 0.87 & 0.62 & 0.57 \\
 & \quad Fully AI & 950 & 0.94 & 0.86 & 0.63 & 0.46 \\
 & \quad Human ideas, AI prose & 952 & 0.97 & 0.92 & 0.64 & 0.60 \\
\midrule
\multicolumn{7}{@{}l}{\textbf{Prose}} \\
\jOpus{Opus~5.5}--\jSol{GPT-6.1 Sol} & All & 2{,}854 & 0.97 & 0.76 & 0.84 & 0.63 \\
 & \quad Fully human & 951 & 0.99 & 0.76 & 0.85 & 0.66 \\
 & \quad Fully AI & 950 & 0.99 & 0.76 & 0.86 & 0.62 \\
 & \quad Human ideas, AI prose & 953 & 0.93 & 0.77 & 0.81 & 0.61 \\
\addlinespace
\jOpus{Opus~5.5}--\jArgon{Argon} & All & 2{,}849 & 0.97 & 0.81 & 0.88 & 0.68 \\
 & \quad Fully human & 947 & 0.99 & 0.83 & 0.90 & 0.72 \\
 & \quad Fully AI & 950 & 0.98 & 0.79 & 0.87 & 0.65 \\
 & \quad Human ideas, AI prose & 952 & 0.94 & 0.80 & 0.86 & 0.66 \\
\addlinespace
\jArgon{Argon}--\jSol{GPT-6.1 Sol} & All & 2{,}849 & 0.96 & 0.74 & 0.84 & 0.64 \\
 & \quad Fully human & 947 & 0.99 & 0.74 & 0.87 & 0.65 \\
 & \quad Fully AI & 950 & 0.98 & 0.73 & 0.85 & 0.63 \\
 & \quad Human ideas, AI prose & 952 & 0.92 & 0.75 & 0.81 & 0.62 \\
\bottomrule
\end{tabular}
\end{table}

\begin{table}[!htbp]
\centering\footnotesize
\caption{Human--human and Claude Opus~5.5--human agreement on prose and ideas judgments. Annotator 1 completed 32 documents and Annotator 2 completed 24. Each score gives eligible documents equal weight. The decisive row uses candidate-membership AC1 and selected-set F1; other F1 rows compare positive labels.}
\label{tab:manual-agreement}
\label{tab:manual-agreement-ideas}
\label{tab:manual-agreement-prose}
\setlength{\tabcolsep}{4pt}
\renewcommand{\arraystretch}{1.12}
\begin{tabular}{@{}llrrrrrr@{}}
\toprule
 & & \multicolumn{2}{c}{\textbf{Annotator 1--Annotator 2}} & \multicolumn{2}{c}{\textbf{\jOpus{Opus~5.5}--Annotator 1}} & \multicolumn{2}{c}{\textbf{\jOpus{Opus~5.5}--Annotator 2}} \\
\cmidrule(lr){3-4}\cmidrule(lr){5-6}\cmidrule(lr){7-8}
Dimension & Judgment & AC1 & F1 & AC1 & F1 & AC1 & F1 \\
\midrule
\multirow{5}{*}{Ideas} & Groups: same point & 0.92 & 0.96 & 0.96 & 0.98 & 0.94 & 0.97 \\ %
 & Unmatched: no match & 0.69 & 0.77 & 0.93 & 0.94 & 0.82 & 0.86 \\ %
 & Same reason & 0.83 & 0.91 & 0.77 & 0.88 & 0.84 & 0.91 \\ %
 & Interpretation & 0.87 & 0.93 & 0.82 & 0.90 & 0.87 & 0.93 \\ %
 & Decisive factors & 0.87 & 0.87 & 0.90 & 0.89 & 0.96 & 0.95 \\ %
\midrule
\multirow{5}{*}{Prose} & Groups: same point & 0.84 & 0.92 & 0.91 & 0.95 & 0.88 & 0.93 \\ %
 & Unmatched: no match & 0.77 & 0.82 & 0.91 & 0.93 & 0.85 & 0.91 \\ %
 & Same reason & 0.80 & 0.91 & 0.79 & 0.88 & 0.87 & 0.93 \\ %
 & Interpretation & 0.83 & 0.92 & 0.80 & 0.89 & 0.88 & 0.94 \\ %
 & Decisive factors & 0.89 & 0.94 & 0.92 & 0.95 & 0.95 & 0.97 \\ %
\bottomrule
\end{tabular}
\end{table}

\begin{figure}[!htbp]
\centering\footnotesize
\begin{tabularx}{\linewidth}{@{}>{\raggedright\arraybackslash}p{2.35cm}>{\raggedright\arraybackslash}X@{}}
\toprule
\textbf{Evidence item 1} & 1. \textbf{IDEA:} A beautiful butterfly is isolated and made lonely by its beauty, inverting the usual transformation story. \textbf{LEANS:} human \textbf{STRENGTH:} medium \textbf{QUOTE:} ``it was lonely for its beauty isolated it,'' \textbf{WHY:} The premise reverses the metamorphosis motif in a naive, personal way. It reads as an emotionally driven choice rather than an engineered twist. \\\addlinespace[3pt]
\textbf{Evidence item 2} & 2. \textbf{IDEA:} The butterfly wants to return to being a caterpillar, anonymous and free of vanity. \textbf{LEANS:} human \textbf{STRENGTH:} medium \textbf{QUOTE:} ``and craved to return to its original self,'' \textbf{QUOTE:} ``a catterpillar, anomonoyous and free'' \textbf{WHY:} Regression as the wish is an unusual, intuitive angle. It sounds like the writer's own feeling about being noticed. \\\addlinespace[3pt]
 & \textit{[items 3--5 omitted; the reply then gives its reasoning, which the judge does not see]} \\\addlinespace[3pt]
\textbf{Decisive factor} & The idiosyncratic inverted-metamorphosis premise and the unexplained, quick resolution decide it. Both point to a human, likely young, writer working from feeling rather than design. \\\addlinespace[3pt]
\textbf{Verdict} & ideas\_ai\_share: 0.05 \quad ideas\_prediction: human \\
\bottomrule
\end{tabularx}
\caption{Example explanation (ideas part). A Claude Opus 5.5 reply to the joint prompt for a human-written poem (MELD; fully human). Each evidence item gives a statement (\textbf{IDEA}), a lean (\textbf{LEANS}), a strength (\textbf{STRENGTH}), supporting quotes (\textbf{QUOTE}), and a justification (\textbf{WHY}). The reply ends with a decisive factor and a verdict. When forming groups, the judge sees only the statements, quotes, and decisive factor. When determining if items in each group rely on the same reason, it sees the statements, leans, and justifications. The prose part has the same form.}
\label{fig:explanation-example}
\end{figure}

\FloatBarrier

\subsubsection{Cues cited in explanations}
\label{app:feat-analysis}\label{app:feat-discovery}\label{app:feat-assignment}\label{app:feat-verification}
We analyze Opus~5.5 and Sol explanations of 2,432 documents from 24 datasets, with separate prose and idea judgments from each detector (9,728 judgments in total). Sol at high effort discovers features from their evidence and reasons, and merges them across datasets and detectors into catalogs of 132 idea and 119 prose features; each itemized evidence is then assigned one feature or none.
Through these catalogs of features, we find that the cues frequently cited by LLM detectors align with findings from prior work.
\autoref{tab:prior} presents two matches with our feature catalog per study that share both the characteristic and the attribution direction; our features describe cues cited by frontier models rather than their reliability as gold provenance markers.

\begin{table}[!htbp]
\centering\footnotesize
\caption{Selected examples of detector-cited features that align in meaning and attribution direction with prior findings.}
\label{tab:prior}
\begingroup\setlength{\tabcolsep}{4pt}\renewcommand{\arraystretch}{1.12}
\begin{tabularx}{\linewidth}{@{}l l >{\raggedright\arraybackslash}p{0.34\linewidth} >{\raggedright\arraybackslash}X@{}}
\toprule
Target & Leans & Feature in detector explanations & Corresponding finding in prior work \\
\midrule
\multicolumn{4}{@{}l}{\textbf{StoryScope}~\citep{russell2026storyscope}} \\
Ideas & AI & Explicit thematic and moral interpretation & AI stories state themes and morals more overtly. \\
Ideas & Human & Reframing revelations and timed disclosures & Human revelations more strongly change the interpretation of earlier scenes. \\
\midrule
\multicolumn{4}{@{}l}{\textbf{IdeaLens}~\citep{rajendhran2026idealensdetectingaiideas}} \\
Ideas & Human & Precise factual and technical anchors & Human ideas use interconnected particulars that anchor an assertion to a specific instance. \\
Ideas & AI & Untailored criticism and recommendations & AI ideas offer generic remedies that remain applicable when the stated problem is replaced. \\
\midrule
\multicolumn{4}{@{}l}{\textbf{Argument Collapse}~\citep{kim2026argumentcollapsellmsflatten}} \\
Ideas & Human & Tailored remedies and localized repairs & Humans propose concrete interventions. \\
Prose & AI & Hedging and diplomatic softening & AI responses hedge and balance stakeholder positions. \\
\bottomrule
\end{tabularx}
\endgroup
\par\smallskip{\footnotesize Two illustrative examples are selected per paper from the saved semantic comparisons; they do not estimate overall agreement. Leans gives the shared human/AI direction. The source studies differ in domain and measurement level.}
\end{table}

\FloatBarrier

\section{Which model wrote the text?}
\label{app:rq2}

In this section, we discuss data, additional results, and analysis related to \autoref{sec:model_detection}.

\subsection{Model attribution data}
\label{app:rq2_data}

Thirty-four models from 11 labs write the texts, released between 19~September~2024 (Qwen2.5 72B) and 19~October~2026 (MiniMax~M3.1 Flash). \autoref{tab:ai_vs_ai_writers} lists each with its release date, defined as in \autoref{tab:atrey_detectors}. Each runs at its default settings. 
Every writer writes from the same 25 prompts in each of three domains (\autoref{tab:rq2_domains}). 
\begin{enumerate}
    \item Stories answer StoryScope's story prompts. For the five models StoryScope covers we use its released stories, and GPT-6 Astra's and Claude Fable~5.1's come from IdeaLens.
    \item News articles start from 25 human-written articles from 17 outlets, published 9--20 October 2026: each writer gets an article's topic and length and writes its own article and title.
    \item Op-eds start from 25 English briefs from MARGO, each giving the topic, the question at issue, the writing date, the target language and the length. The briefs in MARGO's seven other languages are in \S\ref{app:mk_multilingual}. In all there are 2,542 texts; four writers have 71--74 rather than 75 (Claude Opus~5.5, Gemini~3.8, Mistral Large~3 and MiniMax M2.5) due to some refusals.
\end{enumerate}

\begin{table}[!htbp]
\centering\footnotesize
\caption{The three RQ2 domains. Word and token lengths describe the texts scored in Table~\ref{tab:ai_vs_ai_levels}. Token counts use \texttt{o200k\_base} (\texttt{tiktoken} 0.14.0) on the text alone. SD is the sample standard deviation.}
\label{tab:rq2_domains}
\begin{tabular}{@{}llrrrrr@{}}
\toprule
 & & & & \multicolumn{3}{c}{\textbf{Tokens per text}} \\
\cmidrule(lr){5-7}
Domain & Prompts & Texts & Words: median (IQR) & Mean & Median & SD \\
\midrule
All & & 2,542 & & 1,881 & 1,376 & 1,806 \\
\quad Stories & 25 StoryScope story prompts & 844 & 2,008 (1,178--2,696) & 3,169 & 2,566 & 2,554 \\
\quad News & 25 articles from 17 outlets, 9--20 Oct 2026 & 848 & 874 (631--1,211) & 1,195 & 1,084 & 524 \\
\quad Op-eds & 25 English MARGO briefs & 850 & 1,008 (790--1,259) & 1,287 & 1,194 & 735 \\
\bottomrule
\end{tabular}
\end{table}

\subsection{Additional results}
\label{app:ai_vs_ai_full}
\autoref{tab:ai_vs_ai_holistic} lists the 34 writing models with their release dates and, for each, how often each detector names the right lab. Its first three rows match the lab columns of \autoref{tab:ai_vs_ai_levels}. No detector does better on its own lab's texts. Fable~5.2 names Anthropic for 99.3\% of Claude texts against 100\% of OpenAI and Google texts; Opus~5.5 for 95.7\% against 97.7\% and 99.7\%; and Argon names Google for 94.7\% of Google texts against 98.7\% and 100\%. DeepSeek~V4.1 Flash names DeepSeek for none of its own lab's texts. Only Sol is most accurate on its own lab (82.7\% for OpenAI texts, against 68.2\% and 32.1\%), but it answers OpenAI or Anthropic for nearly every text (\S\ref{app:md_profiles}).
\begin{table}[!htbp]
\centering\footnotesize
\caption{\textbf{Detectors name the lab behind nearly every big-three text, but few texts from the other eight labs.} Share of each writer's 75 texts for which each detector names the right lab (\%). The Overall row and each lab's row pool the writers beneath them, and bold numbers mark the best detector in those rows. The other eight labs run from best to worst identified, and writers run newest first. Newer models from those labs are rarely traced to their own lab: Claude Opus~5.5 names Moonshot for 51\% of Kimi~K2.5 texts and for none of Kimi~K3's. Claude Fable 5.2 is the best detector, or tied for best, in every pooled row, so the other eight labs now run from best to worst identified by Fable~5.2. Fable~5.2 names Moonshot for 93\% of Kimi~K2.5 texts and for none of Kimi~K3's. Brackets give 95\% Wilson intervals.}
\label{tab:ai_vs_ai_holistic}\label{tab:ai_vs_ai_writers}
\setlength{\tabcolsep}{3.5pt}\newcommand*{\tabdag}{\rlap{\smash{\textsuperscript{\tabdagger}}}}%
\begin{tabular}{@{}llrrrrr@{\hspace{4pt}}}
\toprule
 & & \multicolumn{5}{c}{\textbf{Right lab (\%)}} \\
\cmidrule(lr){3-7}
Writer & Released & \jFable{Fable~5.2} & \jOpus{Opus~5.5} & \jArgon{Argon} & \jSol{Sol} & \jDeepSeek{DeepSeek} \\
\midrule
Overall (34 writers) & & \textbf{59.9} {\scriptsize(58--62)} & 49.7 {\scriptsize(48--52)} & 44.8 {\scriptsize(43--47)} & 22.6 {\scriptsize(21--24)} & 11.1 {\scriptsize(10--12)} \\
\quad Big three labs (13) & & \textbf{99.8} {\scriptsize(99--100)} & 97.8 {\scriptsize(97--99)} & 97.5 {\scriptsize(96--98)} & 58.8 {\scriptsize(56--62)} & 28.7 {\scriptsize(26--32)} \\
\quad Other eight labs (21) & & \textbf{35.1} {\scriptsize(33--37)} & 19.8 {\scriptsize(18--22)} & 12.0 {\scriptsize(11--14)} & 0.2 {\scriptsize(0--1)} & 0.1 {\scriptsize(0.0--0.5)} \\
\midrule
\labicon{anthropic}\,\textbf{Anthropic} (4) & & \textbf{99.3} {\scriptsize(98--100)} & 95.7 {\scriptsize(93--97)} & 98.7 {\scriptsize(97--99)} & 68.2 {\scriptsize(63--73)} & 62.5 {\scriptsize(57--68)} \\
\quad Claude Opus 5.5 & 22~Sep~2026 & 97.3 {\scriptsize(91--99)}\tabdag & 83.8 {\scriptsize(74--90)}\tabdag & 95.9 {\scriptsize(89--99)}\tabdag & 48.6 {\scriptsize(38--60)}\tabdag & 55.4 {\scriptsize(44--66)}\tabdag \\
\quad Claude Fable 5.1 & 1~Sep~2026 & 100.0 {\scriptsize(95--100)} & 98.7 {\scriptsize(93--100)} & 100.0 {\scriptsize(95--100)} & 68.0 {\scriptsize(57--77)} & 58.7 {\scriptsize(47--69)} \\
\quad Claude Opus 5 & 24~Jul~2026 & 100.0 {\scriptsize(95--100)} & 100.0 {\scriptsize(95--100)} & 98.7 {\scriptsize(93--100)} & 80.0 {\scriptsize(70--87)} & 78.7 {\scriptsize(68--86)} \\
\quad Claude Sonnet 4.6 & 17~Feb~2026 & 100.0 {\scriptsize(95--100)} & 100.0 {\scriptsize(95--100)} & 100.0 {\scriptsize(95--100)} & 76.0 {\scriptsize(65--84)} & 57.3 {\scriptsize(46--68)} \\
\midrule
\labicon{openai}\,\textbf{OpenAI} (4) & & \textbf{100.0} {\scriptsize(99--100)} & 97.7 {\scriptsize(95--99)} & \textbf{100.0} {\scriptsize(99--100)} & 82.7 {\scriptsize(78--87)} & 30.8 {\scriptsize(26--36)} \\
\quad GPT-6.1 Sol & 29~Sep~2026 & 100.0 {\scriptsize(95--100)} & 94.7 {\scriptsize(87--98)} & 100.0 {\scriptsize(95--100)} & 70.7 {\scriptsize(60--80)} & 22.7 {\scriptsize(15--33)} \\
\quad GPT-6 Astra & 3~Sep~2026 & 100.0 {\scriptsize(95--100)} & 96.0 {\scriptsize(89--99)} & 100.0 {\scriptsize(95--100)} & 68.0 {\scriptsize(57--77)} & 17.3 {\scriptsize(10--27)} \\
\quad GPT-5.6 Sol & 9~Jul~2026 & 100.0 {\scriptsize(95--100)} & 100.0 {\scriptsize(95--100)} & 100.0 {\scriptsize(95--100)} & 96.0 {\scriptsize(89--99)} & 38.7 {\scriptsize(28--50)} \\
\quad GPT-5.4 & 5~Mar~2026 & 100.0 {\scriptsize(95--100)} & 100.0 {\scriptsize(95--100)} & 100.0 {\scriptsize(95--100)} & 96.0 {\scriptsize(89--99)} & 44.6 {\scriptsize(34--56)}\tabdag \\
\midrule
\labicon{google}\,\textbf{Google} (5) & & \textbf{100.0} {\scriptsize(99--100)} & 99.7 {\scriptsize(99--100)} & 94.7 {\scriptsize(92--97)} & 32.1 {\scriptsize(28--37)} & 0.0 {\scriptsize(0--1)} \\
\quad Gemini 4 Argon & 13~Oct~2026 & 100.0 {\scriptsize(95--100)} & 98.7 {\scriptsize(93--100)} & 86.7 {\scriptsize(77--93)} & 32.0 {\scriptsize(23--43)} & 0.0 {\scriptsize(0--5)} \\
\quad Gemini 3.8\smash{\textsuperscript{*}} & 2~Sep~2026 & 100.0 {\scriptsize(95--100)}\tabdag & 100.0 {\scriptsize(95--100)}\tabdag & 93.2 {\scriptsize(85--97)}\tabdag & 25.7 {\scriptsize(17--37)}\tabdag & 0.0 {\scriptsize(0--5)}\tabdag \\
\quad Gemma 4 31B\smash{\textsuperscript{*}} & 2~Apr~2026 & 100.0 {\scriptsize(95--100)} & 100.0 {\scriptsize(95--100)} & 98.7 {\scriptsize(93--100)} & 20.0 {\scriptsize(13--30)} & 0.0 {\scriptsize(0--5)} \\
\quad Gemini 3 Flash & 17~Dec~2025 & 100.0 {\scriptsize(95--100)} & 100.0 {\scriptsize(95--100)} & 94.7 {\scriptsize(87--98)} & 14.7 {\scriptsize(8--24)} & 0.0 {\scriptsize(0--5)} \\
\quad Gemma 3 27B & 12~Mar~2025 & 100.0 {\scriptsize(95--100)} & 100.0 {\scriptsize(95--100)} & 100.0 {\scriptsize(95--100)} & 68.0 {\scriptsize(57--77)} & 0.0 {\scriptsize(0--5)} \\
\midrule
\labicon{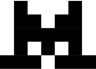}\,\textbf{Mistral}: Mistral Large 3 & 2~Dec~2025 & 91.8 {\scriptsize(83--96)}\tabdag & 64.4 {\scriptsize(53--74)}\tabdag & 72.6 {\scriptsize(61--82)}\tabdag & 0.0 {\scriptsize(0--5)}\tabdag & 0.0 {\scriptsize(0--5)}\tabdag \\
\midrule
\labicon{moonshot}\,\textbf{Moonshot} (4) & & \textbf{57.0} {\scriptsize(51--62)} & 20.7 {\scriptsize(16--26)} & 7.0 {\scriptsize(5--10)} & 0.0 {\scriptsize(0--1)} & 0.0 {\scriptsize(0--1)} \\
\quad Kimi K3 & 16~Jul~2026 & 0.0 {\scriptsize(0--5)} & 0.0 {\scriptsize(0--5)} & 0.0 {\scriptsize(0--5)} & 0.0 {\scriptsize(0--5)} & 0.0 {\scriptsize(0--5)} \\
\quad Kimi K2.6 & 20~Apr~2026 & 77.3 {\scriptsize(67--85)} & 29.3 {\scriptsize(20--40)} & 8.0 {\scriptsize(4--16)} & 0.0 {\scriptsize(0--5)} & 0.0 {\scriptsize(0--5)} \\
\quad Kimi K2.5 & 27~Jan~2026 & 93.3 {\scriptsize(85--97)} & 50.7 {\scriptsize(40--62)} & 17.3 {\scriptsize(10--27)} & 0.0 {\scriptsize(0--5)} & 0.0 {\scriptsize(0--5)} \\
\quad Kimi K2 (0905) & 4~Sep~2025 & 57.3 {\scriptsize(46--68)} & 2.7 {\scriptsize(1--9)} & 2.7 {\scriptsize(1--9)} & 0.0 {\scriptsize(0--5)} & 0.0 {\scriptsize(0--5)} \\
\midrule
\labicon{qwen}\,\textbf{Alibaba} (5) & & \textbf{51.5} {\scriptsize(46--56)} & 37.1 {\scriptsize(32--42)} & 23.5 {\scriptsize(19--28)} & 0.8 {\scriptsize(0--2)} & 0.5 {\scriptsize(0--2)} \\
\quad Qwen3.8 Max & 2~Sep~2026 & 0.0 {\scriptsize(0--5)} & 0.0 {\scriptsize(0--5)} & 0.0 {\scriptsize(0--5)} & 0.0 {\scriptsize(0--5)} & 0.0 {\scriptsize(0--5)} \\
\quad Qwen3.8 27B & 14~Aug~2026 & 1.3 {\scriptsize(0--7)} & 2.7 {\scriptsize(1--9)} & 1.3 {\scriptsize(0--7)} & 0.0 {\scriptsize(0--5)} & 0.0 {\scriptsize(0--5)} \\
\quad Qwen3.6 27B & 22~Apr~2026 & 89.3 {\scriptsize(80--94)} & 68.0 {\scriptsize(57--77)} & 29.3 {\scriptsize(20--40)} & 0.0 {\scriptsize(0--5)} & 1.3 {\scriptsize(0--7)} \\
\quad Qwen3 235B (2507) & 21~Jul~2025 & 80.0 {\scriptsize(70--87)} & 53.3 {\scriptsize(42--64)} & 37.3 {\scriptsize(27--49)} & 0.0 {\scriptsize(0--5)} & 0.0 {\scriptsize(0--5)} \\
\quad Qwen2.5 72B & 19~Sep~2024 & 86.7 {\scriptsize(77--93)} & 61.3 {\scriptsize(50--72)} & 49.3 {\scriptsize(38--60)} & 4.0 {\scriptsize(1--11)} & 1.3 {\scriptsize(0--7)} \\
\midrule
\labicon{deepseek}\,\textbf{DeepSeek} (2) & & \textbf{40.7} {\scriptsize(33--49)} & 34.0 {\scriptsize(27--42)} & 15.3 {\scriptsize(10--22)} & 0.0 {\scriptsize(0--2)} & 0.0 {\scriptsize(0--2)} \\
\quad DeepSeek V4.1 Flash & 10~Sep~2026 & 14.7 {\scriptsize(8--24)} & 9.3 {\scriptsize(5--18)} & 0.0 {\scriptsize(0--5)} & 0.0 {\scriptsize(0--5)} & 0.0 {\scriptsize(0--5)} \\
\quad DeepSeek V3.2 & 1~Dec~2025 & 66.7 {\scriptsize(55--76)} & 58.7 {\scriptsize(47--69)} & 30.7 {\scriptsize(21--42)} & 0.0 {\scriptsize(0--5)} & 0.0 {\scriptsize(0--5)} \\
\midrule
\labicon{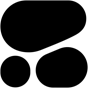}\,\textbf{Cohere}: Command A & 13~Mar~2025 & 21.3 {\scriptsize(14--32)} & 13.3 {\scriptsize(7--23)} & 4.0 {\scriptsize(1--11)} & 0.0 {\scriptsize(0--5)} & 0.0 {\scriptsize(0--5)} \\
\midrule
\labicon{zai}\,\textbf{Zhipu} (4) & & \textbf{10.5} {\scriptsize(8--15)} & 0.0 {\scriptsize(0--1)} & 0.0 {\scriptsize(0--1)} & 0.0 {\scriptsize(0--1)} & 0.0 {\scriptsize(0--1)} \\
\quad GLM-5.3 & 18~Aug~2026 & 0.0 {\scriptsize(0--5)} & 0.0 {\scriptsize(0--5)} & 0.0 {\scriptsize(0--5)} & 0.0 {\scriptsize(0--5)} & 0.0 {\scriptsize(0--5)} \\
\quad GLM-5.1 & 7~Apr~2026 & 11.0 {\scriptsize(6--20)}\tabdag & 0.0 {\scriptsize(0--5)} & 0.0 {\scriptsize(0--5)} & 0.0 {\scriptsize(0--5)} & 0.0 {\scriptsize(0--5)} \\
\quad GLM-4.7 & 22~Dec~2025 & 26.4 {\scriptsize(18--38)}\tabdag & 0.0 {\scriptsize(0--5)}\tabdag & 0.0 {\scriptsize(0--5)} & 0.0 {\scriptsize(0--5)} & 0.0 {\scriptsize(0--5)} \\
\quad GLM-4.5 & 25~Jul~2025 & 5.4 {\scriptsize(2--13)}\tabdag & 0.0 {\scriptsize(0--5)} & 0.0 {\scriptsize(0--5)} & 0.0 {\scriptsize(0--5)} & 0.0 {\scriptsize(0--5)} \\
\midrule
\labicon{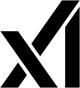}\,\textbf{xAI}: Grok 4.7 & 21~Sep~2026 & 5.3 {\scriptsize(2--13)} & 1.3 {\scriptsize(0--7)} & 0.0 {\scriptsize(0--5)} & 0.0 {\scriptsize(0--5)} & 0.0 {\scriptsize(0--5)} \\
\midrule
\labicon{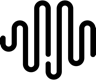}\,\textbf{MiniMax} (3) & & \textbf{2.3} {\scriptsize(1--5)} & 0.0 {\scriptsize(0--2)} & 0.5 {\scriptsize(0--3)} & 0.0 {\scriptsize(0--2)} & 0.0 {\scriptsize(0--2)} \\
\quad MiniMax M3.1 Flash & 19~Oct~2026 & 0.0 {\scriptsize(0--5)} & 0.0 {\scriptsize(0--5)} & 0.0 {\scriptsize(0--5)} & 0.0 {\scriptsize(0--5)} & 0.0 {\scriptsize(0--5)} \\
\quad MiniMax M3 & 1~Jun~2026 & 0.0 {\scriptsize(0--5)} & 0.0 {\scriptsize(0--5)} & 0.0 {\scriptsize(0--5)} & 0.0 {\scriptsize(0--5)} & 0.0 {\scriptsize(0--5)} \\
\quad MiniMax M2.5 & 12~Feb~2026 & 7.0 {\scriptsize(3--15)}\tabdag & 0.0 {\scriptsize(0--5)}\tabdag & 1.4 {\scriptsize(0--8)}\tabdag & 0.0 {\scriptsize(0--5)}\tabdag & 0.0 {\scriptsize(0--5)}\tabdag \\
\bottomrule
\end{tabular}
\par\smallskip{\scriptsize Detectors: Claude~Fable~5.2, Claude~Opus~5.5, Gemini~4 Argon, GPT-6.1 Sol and DeepSeek~V4.1 Flash. $^{*}$\,Version assumed from the op-ed alias (Gemini~3.8) or from texts written on the Gemini API rather than OpenRouter (Gemma~4 31B). $^{\dagger}$\,Out of 71--74 texts rather than 75: the writer produced fewer texts or the detector gave no answer.\par}
\end{table}

\subsubsection{Length and confidence}
\label{app:rq2_length_confidence}
\autoref{tab:rq2_length_confidence} reports lab accuracy by input length and confidence on the common English subset of 299 Anthropic, 300 OpenAI and 374 Google texts (325 news articles, 325 op-eds and 323 stories). All three detectors have a completed, scorable answer for every text except one Argon response without a parsed model answer.

Opus' accuracy stays near 98\% across all length bins, while Argon's accuracy increases from 94.7\% at 500--999 words to about 99\% above 1,000 words. Since texts within each length interval also differ in terms of genre and model author, it is possible that the difference in performance is not solely due to length; Fable makes only two errors, on texts of 521 and 525 words, so we cannot draw conclusions on length's effect.

Higher confidence accompanies higher lab accuracy for Opus and Argon, but confidence concerns the full model identification and is not directly a probability of naming the right lab. Small bins and different confidence scales limit comparisons between detectors.
\begin{table*}[t]
\centering\footnotesize
\caption[Lab accuracy by length and confidence]{{Lab attribution accuracy (\%) on 973 English texts from Anthropic, OpenAI and Google (including Gemma). Parentheses give correct/scorable counts. One Argon response without a parsed model answer is excluded. Length is measured in words as per whitespace split.}}
\label{tab:rq2_length_confidence}
{
\setlength{\tabcolsep}{7pt}
\renewcommand{\arraystretch}{1.08}
\begin{tabular*}{\textwidth}{@{\extracolsep{\fill}}lrrrr@{}}
\toprule
\multicolumn{5}{l}{\textbf{(a) Lab accuracy (\%) by text length}} \\
Words & Texts & \jFable{Fable~5.2} & \jOpus{Opus~5.5} & \jArgon{Gemini~4 Argon} \\
\midrule
All & 973 & 99.8 (971/973) & 97.8 (952/973) & 97.6 (949/972) \\
\midrule
\quad <500 & 29 & 100.0 (29/29) & 100.0 (29/29) & 96.6 (28/29) \\
\quad 500--999 & 301 & 99.3 (299/301) & 97.7 (294/301) & 94.7 (285/301) \\
\quad 1,000--1,499 & 355 & 100.0 (355/355) & 97.7 (347/355) & 99.2 (351/354) \\
\quad 1,500+ & 288 & 100.0 (288/288) & 97.9 (282/288) & 99.0 (285/288) \\
\bottomrule
\end{tabular*}
\vspace{7pt}
\begin{tabular*}{\textwidth}{@{\extracolsep{\fill}}lrlrlr@{}}
\multicolumn{6}{l}{\textbf{(b) Lab accuracy (\%) by reported confidence}} \\
\multicolumn{2}{c}{\jFable{Fable~5.2}} & \multicolumn{2}{c}{\jOpus{Opus~5.5}} & \multicolumn{2}{c}{\jArgon{Gemini~4 Argon}} \\
\cmidrule(lr){1-2}\cmidrule(lr){3-4}\cmidrule(lr){5-6}
Confidence & Accuracy (\%) & Confidence & Accuracy (\%) & Confidence & Accuracy (\%) \\
\midrule
<0.30 & 100.0 (6/6) & <0.30 & 83.3 (5/6) & 0.40--<0.50 & 100.0 (1/1) \\
0.30--<0.40 & 99.0 (190/192) & 0.30--<0.40 & 95.2 (279/293) & 0.50--<0.75 & 100.0 (10/10) \\
0.40--<0.50 & 100.0 (573/573) & 0.40--<0.50 & 99.0 (499/504) & 0.75--<0.85 & 89.5 (77/86) \\
0.50--<0.75 & 100.0 (202/202) & 0.50--<0.75 & 99.4 (169/170) & 0.85--<0.90 & 96.0 (216/225) \\
 &  &  &  & 0.90--<0.95 & 99.1 (527/532) \\
 &  &  &  & 0.95--1.00 & 100.0 (118/118) \\
\bottomrule
\end{tabular*}
}
\end{table*}

\subsubsection{Op-eds in seven other languages}
\label{app:mk_multilingual}

The 34 writers also wrote op-eds from five briefs in each of seven other languages: Bulgarian, French, Hindi, Japanese, Polish, Russian and Spanish, 1,190 texts in all. \autoref{tab:ai_vs_ai_multilingual} reports how often Opus~5.5 names the right lab, model family and exact model for them, next to the same writers' English op-eds.
\begin{table}[!htbp]
\centering\small
\caption{Claude Opus~5.5 attributes the non-English op-eds about as well as the English ones. Share of op-eds (\%) for which Opus~5.5 names the right lab, model family or exact model with the holistic prompt, scored as in \autoref{tab:ai_vs_ai_levels}. The 34 writers each wrote from five briefs per language; Opus~5.5 refused 2 of the 1,190 non-English op-eds (both by MiniMax~M3.1 Flash), and we leave them out. The English row is Opus~5.5 on the same writers' English op-eds.}
\label{tab:ai_vs_ai_multilingual}
\begin{tabular}{@{}lrrrrrr@{}}
\toprule
 & & \multicolumn{3}{c}{\textbf{Lab (\%)}} & & \\
\cmidrule(lr){3-5}
Language & Texts & All & Big three & Other eight & Family (\%) & Model (\%) \\
\midrule
All non-English & 1,188 & 47.4 & 98.2 & 15.8 & 27.9 & 2.0 \\
\quad Bulgarian & 170 & 51.2 & 96.9 & 22.9 & 31.8 & 2.9 \\
\quad French & 170 & 48.2 & 98.5 & 17.1 & 30.0 & 2.4 \\
\quad Hindi & 170 & 42.9 & 100.0 & 7.6 & 21.8 & 0.0 \\
\quad Japanese & 169 & 45.0 & 96.9 & 12.5 & 23.7 & 0.6 \\
\quad Polish & 170 & 50.6 & 98.5 & 21.0 & 31.2 & 2.4 \\
\quad Russian & 170 & 45.9 & 100.0 & 12.4 & 26.5 & 2.9 \\
\quad Spanish & 169 & 47.9 & 96.9 & 17.3 & 30.8 & 3.0 \\
\midrule
English & 850 & 49.1 & 99.1 & 18.1 & 32.0 & 5.2 \\
\bottomrule
\end{tabular}
\end{table}

\subsubsection{Lab confusion}
\label{app:rq2_confusion}
\begin{figure}[!htbp]
\centering
\includegraphics{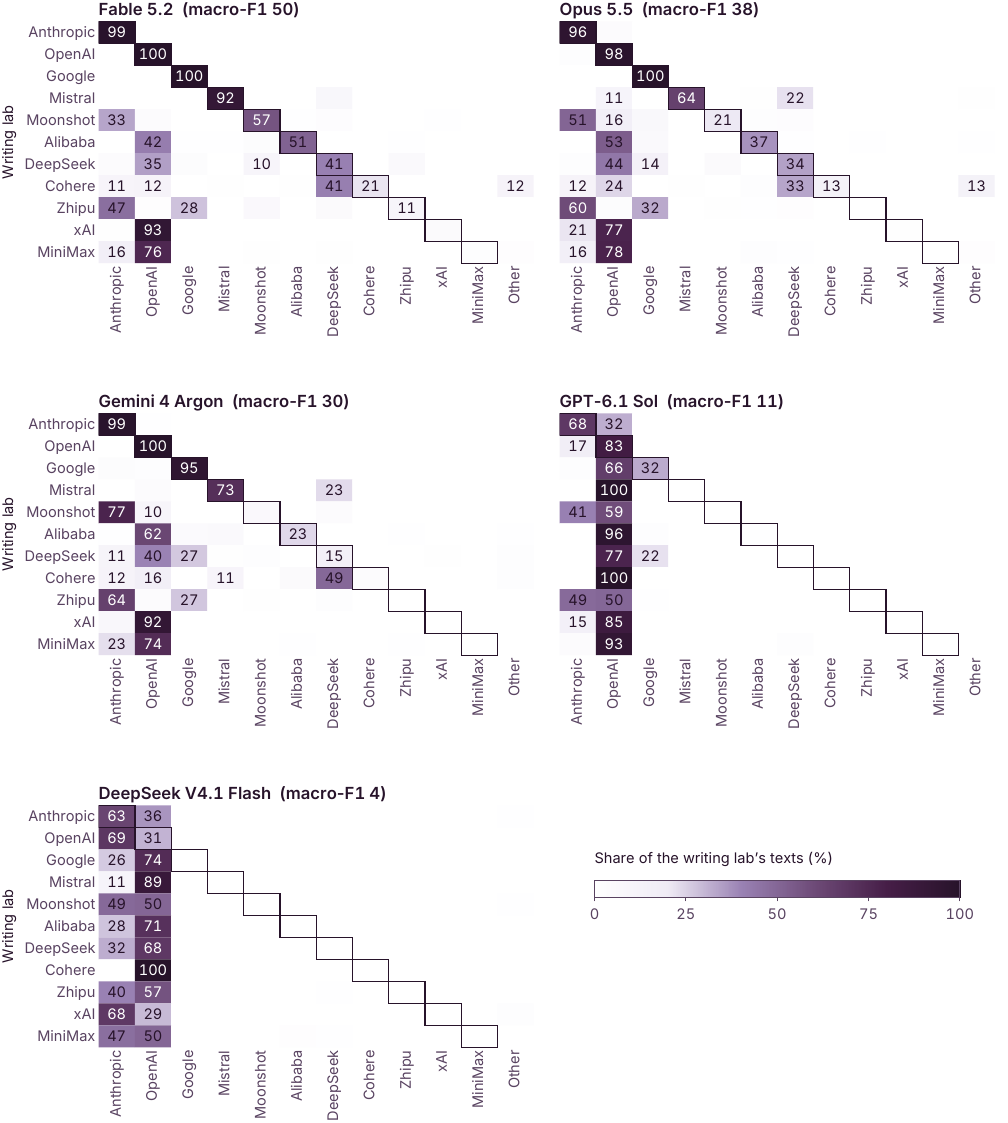}
\caption{Which lab each detector names for each writing lab's texts (34-writer panel, English texts; \% of the writing lab's texts). Outlined cells are the right lab; Other pools Meta, Microsoft, any other lab and replies that name no model. Fable~5.2, Opus~5.5 and Argon recognize the big three labs, but give most xAI and MiniMax texts to OpenAI and many Moonshot and Zhipu texts to Anthropic; Sol and DeepSeek~V4.1 Flash name Anthropic or OpenAI for nearly every text.}
\label{fig:rq2_confusion}
\end{figure}

Accuracy on the big three labs also reflects how often detectors name them. \autoref{fig:rq2_confusion} shows, for each detector, which lab it names for each writing lab's texts, and \autoref{tab:rq2_precision} gives the precision of its answers per named lab and its macro-F1 over the 11 writing labs. When Fable~5.2 names Anthropic or OpenAI it is right only 51\% and 39\% of the time, because it also gives these labs most texts from xAI and MiniMax, about half of Zhipu's and over a third of Moonshot's; its answers naming any of the other eight labs are right 79\% of the time. Opus~5.5 and Argon name Anthropic and OpenAI even more often. Sol and DeepSeek~V4.1 Flash name Anthropic or OpenAI for nearly every text, and DeepSeek never names Google, so their macro-F1 is 11 and 4. We hypothesize some of this over-predicting may be effects of distillation.

\begin{table}[!htbp]
\centering\footnotesize
\caption{Precision of each detector's lab answers (\%; in brackets, the number of texts it attributes to that lab) and macro-F1 over the 11 writing labs. Other eight pools the answers naming any lab outside Anthropic, OpenAI and Google. DeepSeek~V4.1 Flash never names Google.}
\label{tab:rq2_precision}
\setlength{\tabcolsep}{4pt}
\begin{tabular}{lrrrrr}
\toprule
 & & \multicolumn{4}{c}{Precision when it names (\%)} \\
\cmidrule(lr){3-6}
Detector & Macro-F1 & Anthropic & OpenAI & Google & Other eight \\
\midrule
\jFable{Fable~5.2} & 49.9 & 50.6 {\scriptsize(587)} & 38.7 {\scriptsize(776)} & 80.4 {\scriptsize(465)} & 79.0 {\scriptsize(694)} \\
\jOpus{Opus~5.5} & 37.8 & 41.5 {\scriptsize(689)} & 33.1 {\scriptsize(886)} & 69.5 {\scriptsize(537)} & 75.8 {\scriptsize(409)} \\
\jArgon{Gemini~4 Argon} & 29.8 & 36.2 {\scriptsize(816)} & 33.6 {\scriptsize(894)} & 71.7 {\scriptsize(494)} & 58.0 {\scriptsize(326)} \\
\jSol{GPT-6.1 Sol} & 10.7 & 36.8 {\scriptsize(555)} & 13.7 {\scriptsize(1,808)} & 73.2 {\scriptsize(164)} & 21.4 {\scriptsize(14)} \\
\jDeepSeek{DeepSeek~V4.1 Flash} & 3.5 & 17.4 {\scriptsize(1,074)} & 6.4 {\scriptsize(1,441)} & -- & 14.3 {\scriptsize(14)} \\
\bottomrule
\end{tabular}
\end{table}

\FloatBarrier%

\subsection{Analysis}
\label{app:rq2_analysis}

\subsubsection{Model writing profiles}
\label{app:md_profiles}
\paragraph{Building the profiles.} Detectors will often describe the same writing traits in different words. So to count how often the detectors cite each trait, we adapt the methods of \citet{brassington2026singleturn} and \citet{pham2024topicgptpromptbasedtopicmodeling} to assemble our model profiles:
\begin{enumerate}
\item We split each explanation into its key observations, each of which points to a lab, e.g.\ ``numerical claims come with hedges about scope and method'' (GPT). Opus~5.5, Argon, Sol and DeepSeek in total generated 288,430 observations.
\item Opus~5.5 groups each detector's observations about each lab into cues, then merges them into three lists for observations on the text overall (489 cues), its ideas (492 cues), and its prose (337 cues), so ``numerical claims come with hedges about scope and method'' and ``repeatedly hedges against overclaiming'' are counted as the same trait.
\item Opus~5.5 labels every observation with these lists, without seeing the detector or the lab. Since Fable~5.2 was released after feature discovery for the writing traits, we only run this labelling step on its observations.
\end{enumerate}
We define a lab's traits as cues at least twice as frequent in observations pointing to that lab as in those pointing to other labs (e.g.\ dry wit for Claude), and say a writer shares a trait if its rate on the writer's texts is at least the midpoint between the lab's rate and the highest other big lab's.

We invite readers to explore each detector's profiles of GPT, Claude, Gemini and the open-weight models, along with the quoted evidence behind every cue, on our interactive \href{https://model-profiles.github.io/}{model profiles page}.

\paragraph{GPT-6.1 Sol and DeepSeek~V4.1 Flash have flawed profiles of GPT and Claude.}
Sol and DeepSeek name OpenAI or Anthropic for nearly every text (93\% and 99\%). The traits they assign GPT and Claude are also given to other labs' texts. These traits are largely nonspecific and few actually set either lab apart. Other labs are assigned GPT traits more often than GPT models themselves.

\subsubsection{Misattributions and reported distillation}
\label{app:md_distill}

This section expands on the distillation results in \S\ref{sec:md_shared}.

\begin{figure}[!t]
  \centering
  \includegraphics{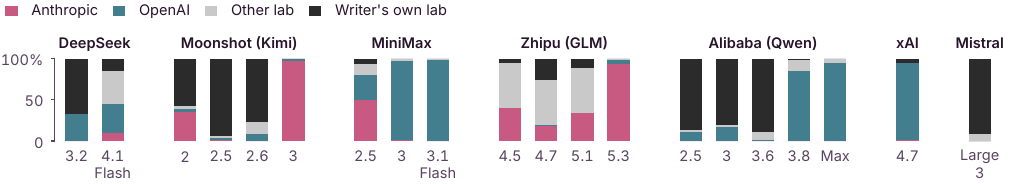}%
  \caption{Claude Fable~5.2 recognizes most texts from earlier Kimi and Qwen releases, but attributes nearly every Kimi~K3 and GLM-5.3 text to Claude and most Qwen3.8~27B and Grok~4.7 texts to GPT. Each column is one writer, oldest to newest within each lab.}
  \label{fig:md_distill_cols}
\end{figure}

\paragraph{Claude Fable~5.2 recognizes early Kimi and Qwen releases far better than Opus~5.5 and Gemini~4 Argon, but all three misattribute the newest ones.}
If a detector is able to recognize a lab's earlier models, we would expect it to also recognize its newer ones, unless their style has changed. Fable~5.2 names the right lab for most texts from Kimi~K2.5 and K2.6 and from the three Qwen releases preceding Qwen3.8, where Opus~5.5 and Argon often answer Claude or GPT instead (\autoref{tab:ai_vs_ai_holistic}). But even Fable~5.2 attributes nearly every Kimi~K3 and GLM-5.3 text to Claude, and most Qwen3.8~27B and Grok~4.7 texts to GPT (\autoref{fig:md_distill_cols}). Because Fable~5.2 knows these labs' earlier models, its errors suggest that their newest releases no longer read like those models, but like Claude or GPT. Fable~5.2 recognizes few texts from any GLM or MiniMax release, so we cannot make the same comparison for these two labs.

\paragraph{Later reports name Kimi K3 and GLM-5.3, and date campaigns against Claude shortly before their release.} The first reports named labs but not models, and did not say when the campaigns ran. Later reports do both. A joint advisory by US agencies reports that Moonshot trained Kimi~K3 on Claude Fable~5 data \citep{cisa2026distillation}, and Anthropic reports a Moonshot campaign against Claude between May and July 2026 \citep{anthropic2026misuse}. K3 came out in July. Anthropic also reports a 17-day Zhipu campaign against Claude Opus~4.8 in June and July, and that Zhipu distilled the cyber capabilities of multiple frontier models during the development of GLM-5.3, which came out in August \citep{anthropic2026misuse, anthropic2026defenses}. These are the two releases that all three detectors attribute almost entirely to Claude.

\paragraph{The detectors attribute the newest Qwen and MiniMax releases to GPT.}
Reports link Alibaba and MiniMax to both Claude and GPT. Anthropic reported more exchanges from Alibaba than from any other lab, used to distill Claude into Qwen3.5 through 3.7 \citep{anthropic2026misuse}, and the advisory reports that both labs distilled Claude models and GPT-5 \citep{cisa2026distillation}. Fable~5.2 names Alibaba for 80\% or more of the texts from each of the three earlier Qwen releases, but all three detectors attribute most Qwen3.8~27B and Qwen3.8~Max texts to GPT. Anthropic reported distillation of Claude by MiniMax in February \citep{anthropic2026distillation}, the month MiniMax~M2.5 came out, and the detectors attribute M2.5 mostly to Claude. The additional link to GPT-5 came in the September advisory, and the detectors attribute nearly every text by the newer M3 and M3.1 Flash to GPT.

\subsubsection{Exploring alternative explanations for the misattributions}
\label{app:md_stylometry}

The misattributions in \autoref{fig:md_distill_cols} are consistent with reported distillation, but there are other explanations that could produce them. Opus~5.5 and Argon often answer Claude or GPT for the earlier Kimi and Qwen releases that Fable~5.2 is able to properly attribute, so detectors may be falling back to known labs when they don't recognize the model who wrote the text, for instance because it came out after the detectors' training cutoff. Another possibility is the newest open models may resemble Claude or GPT only because \citet{clark2026hard} found that recent models from several labs share an information-dense writing style. We investigate both using stylometric analysis by checking whether the detectors' answers match stylometry, and then controlling for information density features from \citet{clark2026hard} to see if the newest releases are still closest to Claude or GPT.

\begin{figure}[!htbp]
  \centering
  \includegraphics{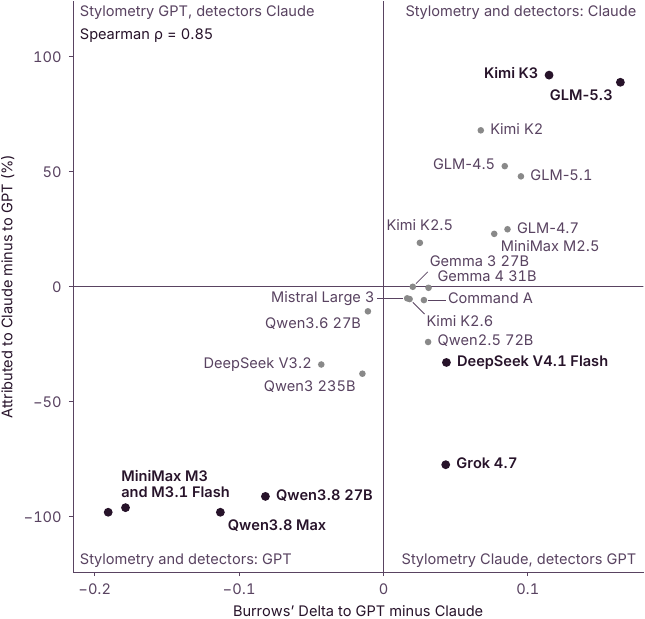}
  \caption{The closer stylometry places a writer to Claude or to GPT, the more often the detectors attribute its texts to that lab. Each point is one writer outside of the Claude, GPT, and Gemini model families (newest releases in bold), with detectors and domains pooled. Burrows' Delta is measured to each lab's closest model.}
  \label{fig:md_stylometry_agree}
\end{figure}

\paragraph{Stylometric analysis.}
We compare writers with Burrows' Delta \citep{burrows2002delta} over the relative frequencies of the 150 most common words and 16 punctuation, sentence and formatting features.\footnote{Em dashes, en dashes, semicolons, colons, parentheses, question marks, exclamation marks, ellipses, quotation marks, ``not X but Y'', words per sentence, spread of sentence length, one-sentence paragraphs, words per paragraph, markdown lines and word length.} Within each domain of \S\ref{app:rq2_data}, we z-score each feature over all texts, and the distance between two writers is the mean absolute difference of their average z-scores over the prompts both wrote (0 means identical). For each pair of consecutive releases from a lab outside of the big three, we take the change in distance to each Claude, GPT, and Gemini model and test it with a paired bootstrap over prompts (2,000 resamples) and Holm correction across those models. 
We estimate the distance between the writer and itself as about 0.2 by splitting each writer's texts into two random halves and measuring the distance between them. Stylometry is able to separate the big three, with each Claude, GPT and Gemini model nearest another model from its own lab in every domain. This is consistent with \citet{sun2025idiosyncrasies}, who find that classifiers are able to tell apart the outputs of different LLM families, and \citet{shan2026stylometric}, who find that AI writers have a consistent stylometric footprint.

\paragraph{Most misattributions reflect similarities in writing style rather than detector error.}
We first look at whether the detectors' answers follow each writer's style, or whether they fall back to a known lab regardless of how the text is written. We compare the percent of each writer's texts the detectors attribute to Claude with how much closer stylometry places it to Claude than to GPT (\autoref{fig:md_stylometry_agree}), and find that the two are strongly correlated (Spearman $\rho = 0.85$), with about the same correlation for each detector on its own. The stylometry, like our detectors, split the newest Kimi, GLM, Qwen, and MiniMax releases between Claude and GPT (all but Qwen3.8~27B are within 0.21--0.24 of their nearest model, against at least 0.27 for these labs' earlier releases). Within the MiniMax family, stylometry places M2.5 nearest Claude Sonnet~4.6 and M3 nearest GPT-5.6 Sol on news and op-eds, and the detectors go from mostly answering Claude for M2.5 to nearly always answering GPT for M3.

\paragraph{Information density explains only part of the shift toward Claude and GPT.}
\citet{clark2026hard} find that recent models from several labs write in a dense style that saves tokens, but also makes it harder for people to read. To test whether the misattributions can be explained by a convergence in AI writing style, we re-implement six of their measures: object relative clauses with and without a relativizer, stacked nouns (compounds of three or more nouns), dependency length, unattested bigrams, and trigram perplexity.\footnote{We parse with spaCy
(\texttt{en\_core\_web\_trf}) and build the reference sample from 500,000 words of English Wikipedia.} We then regress each stylometric feature on these six measures over all texts in a domain, and recompute the distances on the residuals. The six measures predict little of most features, with an average $R^2$ of 0.06--0.08, and predict word length and sentence length best. On the residuals, Kimi~K3 and GLM-5.3 are still nearest a Claude model and the newest Qwen and MiniMax releases nearest a GPT model in every domain, with 7 of the 12 shifts toward Claude and 16 of the 18 toward GPT still significant.\footnote{The shifts toward Claude keep a median of 60\% of their size (Kimi~K3 48\%, GLM-5.3 71\%), and the shifts toward GPT keep 101\% (Qwen 105\%, MiniMax 101\%).}

\section{Which human wrote the text?}
\label{app:rq3}

In this section, we discuss data, additional results, and analysis related to human author attribution (\S\ref{sec:human_attribution}).

\subsection{Data preparation}
\label{app:rq3_data_prep}
Before sending each text to the model for attribution, we remove all direct identity clues (i.e. paper bylines), while keeping some indirect clues like self-citations.

\textbf{Papers.} We extract the main text from arXiv's HTML. We remove the byline (author names, affiliations and emails), the bibliography, the appendices and any section about acknowledgements, funding, author contributions, competing interests or disclosures. 
We replace emails, URLs (including GitHub and Hugging Face links) and arXiv identifiers with placeholders, as well as any affiliation from the byline that appears verbatim. We keep in-body citations. For each paper, we count self-citations and split accuracy by these counts (\autoref{tab:atrey_rq3_checks}). Since the full main text is the condition most exposed to leakage, the topic-only baselines in \S\ref{app:rq3_topic_only} give the model only keywords or the title and abstract.

\textbf{Public reviews.} We remove every occurrence of the reviewer's name, form checklists, expertise statements and boilerplate. Reviewer names and ORCID IDs are kept only in the answer key. We report accuracy with and without self citations (\autoref{tab:atrey_rq3_checks}).

\textbf{Private reviews.} We remove submission numbers, paper titles, author lists, scores and reviewer names. Self-citations are kept.

\textbf{Stories.} We remove author bios, credits, promotions, editors' introductions, authors' notes and content warnings.

\textbf{Reddit.} We remove quoted lines and keep only comments with at least 200 words of the author's own text. We leave out moderator comments, bot-like accounts, comments that contain any form of the author's username, self-promotion, links to personal pages and sensitive disclosures.

\subsection{Experiment \& data details}
\label{app:rq3_data}
As we observe first-hand and is studied in \citet{zhong2026userawareness}, personal account information passed through harnesses like Claude Code can affect model responses. Therefore, we run all of our human author attribution experiments through APIs (\autoref{tab:rq3_access}).

\begin{table}[!htbp]
\centering\footnotesize
\caption{We list the access methods and configured reasoning effort for all models in \S\ref{sec:human_attribution}. Models receive the system prompt and the bare text, with no tools, web search, or candidate names.}
\label{tab:rq3_access}
\begin{tabularx}{\linewidth}{llX}
\toprule
Model & Effort & Access \\
\midrule
\jFable{Fable~5.2} & high & OpenRouter batches, pinned to Anthropic \\
\jOpus{Opus~5.5} & medium & Anthropic Message Batches API; OpenRouter batches for stories and ICML, and as a fallback \\
\jCarbon{Gemini~4 Carbon} & medium & Gemini Batch API; standard Gemini API for output-cap reruns \\
\jArgon{Gemini~4 Argon} & medium & Standard and Batch Gemini APIs \\
\jSol{GPT-6.1 Sol} & medium & OpenAI Responses Batch API; OpenRouter batches for ICML, pinned to OpenAI \\
\jDeepSeek{DeepSeek~V4.1 Flash} & high & OpenRouter, pinned providers \\
\bottomrule
\end{tabularx}
\end{table}

\begin{figure}[!htbp]
\centering
\includegraphics[width=\textwidth]{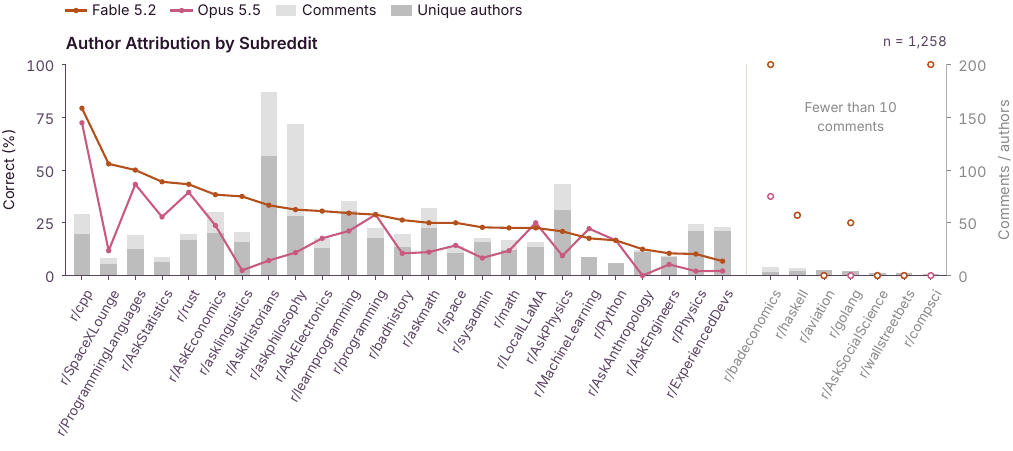}
\caption{Accuracy varies widely across subreddits. Bars (right axis): the number of Reddit comments from each of the 32 subreddits (1,258 in total). Lines (left axis): the percent of comments in each subreddit for which Fable~5.2 and Opus~5.5 name the right username.}
\label{fig:rq3_subreddits}
\end{figure}

\paragraph{Reddit communities and comment selection.}
When collecting the data, we queried Claude to choose subreddits likely to contain enough recent, long-form posts ($>200$ words). \autoref{fig:rq3_subreddits} lists the 32 subreddits and plots the number of comments from each and Fable~5.2's and Opus~5.5's accuracy within each.

\subsection{Additional results}
\label{app:rq3_results}
We present full results from our human attribution experiments that are mentioned briefly in the main paper (i.e. confidence levels, accuracy by domain etc.)

\subsubsection{Accuracy by domain}
\label{app:rq3-domains}
\autoref{tab:rq3_full} gives each judge's accuracy in every domain of \autoref{fig:rq3_domains}, including Gemini~4 Carbon, which was run on every domain but the ICML 2020 reviews. \autoref{tab:atrey_rq3_pangram} splits accuracy on public reviews and papers by the Pangram~4.1 label of the judged text.

\begin{table}[!htbp]
\centering\footnotesize
\providecommand{\jStackT}[3]{\begin{tabular}[b]{@{}c@{}}#1\\[-1pt]{\scriptsize #2}\\[-2pt]{\scriptsize #3}\end{tabular}}
\setlength{\tabcolsep}{4pt}
\caption{Human author attribution (\S\ref{sec:human_attribution} (\%)) by domain and model on the texts of \autoref{fig:rq3_domains}. \emph{Not AI}: texts Pangram~4.1 does not label AI. Accuracy is over answered texts. Gemini~4 Carbon was not run on the ICML 2020 reviews.}
\label{tab:rq3_full}
\begin{tabular}{l l rrrrrr}
\toprule
Domain & Texts & \jStackT{\iconFable}{\textcolor{inkFable}{Fable}}{\textcolor{inkFable}{5.2}} & \jStackT{\iconOpus}{\textcolor{inkOpus}{Opus}}{\textcolor{inkOpus}{5.5}} & \jStackT{\iconCarbon}{\textcolor{inkCarbon}{Gemini 4}}{\textcolor{inkCarbon}{Carbon}} & \jStackT{\iconArgon}{\textcolor{inkArgon}{Gemini 4}}{\textcolor{inkArgon}{Argon}} & \jStackT{\iconSol}{\textcolor{inkSol}{GPT-6.1}}{\textcolor{inkSol}{Sol}} & \jStackT{\labicon{deepseek}}{DeepSeek}{V4.1 Flash} \\
\midrule
Stories & all (206) & 32.0 & 19.9 & 19.7 & 18.7 & 22.8 & 2.4 \\
 & not AI (191) & 34.6 & 21.5 & 21.3 & 20.2 & 24.6 & 2.6 \\
\addlinespace
Papers & all (120) & 46.2 & 34.7 & 31.9 & 31.7 & 22.5 & 7.6 \\
 & not AI (90) & 57.3 & 43.2 & 40.4 & 40.0 & 28.9 & 10.1 \\
\addlinespace
Public reviews & all (1,559) & 21.4 & 15.2 & 18.8 & 16.9 & 13.0 & 5.7 \\
 & not AI (789) & 39.3 & 28.4 & 34.5 & 30.7 & 24.1 & 9.9 \\
\addlinespace
Private reviews & all (398) & 56.0 & 41.2 & 40.3 & 40.7 & 9.3 & 2.5 \\
 & not AI (396) & 56.3 & 41.4 & 40.5 & 40.9 & 9.3 & 2.5 \\
\addlinespace
ICML 2020 reviews & all (15,136) & 55.5 & 48.8 & -- & 37.7 & 22.3 & 5.4 \\
\addlinespace
Reddit & all (1,258) & 31.1 & 16.8 & 12.0 & 10.0 & 5.2 & 5.3 \\
 & not AI (1,230) & 31.8 & 17.2 & 12.3 & 10.2 & 5.4 & 5.4 \\
\bottomrule
\end{tabular}
\end{table}

\begin{table}[h]
\centering\footnotesize
\providecommand{\jStackT}[3]{\begin{tabular}[b]{@{}c@{}}#1\\[-1pt]{\scriptsize #2}\\[-2pt]{\scriptsize #3}\end{tabular}}
\caption{Every detector names fewer human authors on texts that  Pangram~4.1 labels AI or mixed. \S\ref{sec:human_attribution} dataset accuracies (\%) by the Pangram~4.1 label of the text.} %
\label{tab:atrey_rq3_pangram}
\setlength{\tabcolsep}{4pt}
\begin{tabular}{l l r rrrrrr}
\toprule
Domain & Pangram label & Texts & \jStackT{\iconFable}{\textcolor{inkFable}{Fable}}{\textcolor{inkFable}{5.2}} & \jStackT{\iconOpus}{\textcolor{inkOpus}{Opus}}{\textcolor{inkOpus}{5.5}} & \jStackT{\iconCarbon}{\textcolor{inkCarbon}{Gemini 4}}{\textcolor{inkCarbon}{Carbon}} & \jStackT{\iconArgon}{\textcolor{inkArgon}{Gemini 4}}{\textcolor{inkArgon}{Argon}} & \jStackT{\iconSol}{\textcolor{inkSol}{GPT-6.1}}{\textcolor{inkSol}{Sol}} & \jStackT{\labicon{deepseek}}{DeepSeek}{V4.1 Flash} \\
\midrule
Public reviews & Human & 571 & 45.4 & 32.4 & 40.4 & 36.1 & 28.2 & 10.3 \\
 & Mixed & 218 & 23.8 & 17.9 & 19.0 & 16.5 & 13.3 & 8.8 \\
 & AI & 770 & 3.5 & 2.0 & 2.9 & 2.7 & 1.6 & 1.3 \\
\addlinespace
Papers & Human & 34 & 84.8 & 72.7 & 61.8 & 64.7 & 35.3 & 17.6 \\
 & Mixed & 56 & 41.1 & 25.5 & 27.3 & 25.0 & 25.0 & 5.5 \\
 & AI & 30 & 13.3 & 10.0 & 6.7 & 6.7 & 3.3 & 0.0 \\
\bottomrule
\end{tabular}
\end{table}

\autoref{tab:rq3-private-meta} breaks accuracy on private reviews down by review type (meta-reviews vs.\ standard peer reviews). Meta-reviews are harder to attribute.

\begin{table}[!htbp]
\centering\footnotesize
\providecommand{\jStackT}[3]{\begin{tabular}[b]{@{}c@{}}#1\\[-1pt]{\scriptsize #2}\\[-2pt]{\scriptsize #3}\end{tabular}}
\setlength{\tabcolsep}{4pt}
\caption{Accuracy (\%) on the 398 private reviews from \S\ref{sec:human_attribution}, with and without meta-reviews. 132 of the 398 reviews are meta-reviews written by two of the six writers. Pangram~4.1 labels none of the meta-reviews AI.}
\label{tab:rq3-private-meta}
\begin{tabular}{l rrrrrr}
\toprule
Texts & \jStackT{\iconFable}{\textcolor{inkFable}{Fable}}{\textcolor{inkFable}{5.2}} & \jStackT{\iconOpus}{\textcolor{inkOpus}{Opus}}{\textcolor{inkOpus}{5.5}} & \jStackT{\iconCarbon}{\textcolor{inkCarbon}{Gemini 4}}{\textcolor{inkCarbon}{Carbon}} & \jStackT{\iconArgon}{\textcolor{inkArgon}{Gemini 4}}{\textcolor{inkArgon}{Argon}} & \jStackT{\iconSol}{\textcolor{inkSol}{GPT-6.1}}{\textcolor{inkSol}{Sol}} & \jStackT{\labicon{deepseek}}{DeepSeek}{V4.1 Flash} \\
\midrule
\multicolumn{7}{l}{\textbf{All private reviews}} \\
\quad All (398) & 56.0 & 41.2 & 40.3 & 40.7 & 9.3 & 2.5 \\
\quad Reviews (266) & 68.4 & 51.1 & 50.4 & 51.1 & 13.2 & 3.8 \\
\quad Meta-reviews (132) & 31.1 & 21.2 & 19.2 & 19.7 & 1.5 & 0.0 \\
\bottomrule
\end{tabular}
\end{table}

\subsubsection{Author-level accuracy}
\label{app:rq3_author_level}
All results in \S\ref{sec:human_attribution} report accuracy in terms of the number of texts which are correctly attributed. Since there is an imbalance in the number of texts per author (especially for private reviews where there are only 6 authors), we also report the accuracy at the author level. We do this by calculating the accuracy for each author's texts and then averaging all authors' individual accuracies (\autoref{tab:rq3_author_level}). Accuracy averaged over authors is lower than over texts on Reddit (24\% against 31\% for Fable~5.2), because the most active users are both easier to name and contribute more comments. Fable~5.2 names at least one text's author for 26\% of Reddit users, 29\% of story authors and 22\% of public reviewers.

\begin{table}[!htbp]
\centering\footnotesize
\caption{Accuracy by text and by author for human author attribution (\S\ref{sec:human_attribution}). Accuracy over answered texts (\%); author-macro is the mean of per-author accuracy; identified is the percent of authors named for at least one of their texts (public reviews include the 90 Nature-family reviews). Authors counts only authors with at least one answered text (e.g. 860 of the 870 Reddit users for Fable~5.2, it refused on the rest).}
\label{tab:rq3_author_level}
\setlength{\tabcolsep}{3.5pt}
\begin{tabular}{llrrrrr}
\toprule
Detector & Domain & Answered & Authors & Accuracy & Author-macro & Identified \\
\midrule
\jFable{Fable~5.2} & Stories & 206 & 187 & 32.0 & 28.1 & 28.9 \\
 & Reddit & 1,244 & 860 & 31.1 & 24.1 & 26.2 \\
 & Public reviews & 1,530 & 1,467 & 21.4 & 21.5 & 21.7 \\
 & Private reviews & 398 & 6 & 56.0 & 49.1 & 100.0 \\
\midrule
\jOpus{Opus~5.5} & Stories & 206 & 187 & 19.9 & 16.6 & 16.6 \\
 & Reddit & 1,237 & 859 & 16.8 & 12.6 & 14.6 \\
 & Public reviews & 1,521 & 1,457 & 15.2 & 15.3 & 15.5 \\
 & Private reviews & 398 & 6 & 41.2 & 29.5 & 100.0 \\
\bottomrule
\end{tabular}
\end{table}

\subsubsection{Accuracy by author visibility}
\label{app:rq3_visibility}
\autoref{tab:rq3_visibility_reddit} and \autoref{tab:rq3_visibility_stories} give the accuracies plotted in \autoref{fig:rq3_fame}, binned by Reddit activity and number of Goodreads ratings.

\begin{table}[!htbp]
\centering\footnotesize
\caption{Reddit author attribution accuracy (\%) by the account's activity: its posts plus comments in the same subreddit from June 2025 to May 2026. Comments is the number in each bin.}
\label{tab:rq3_visibility_reddit}
\setlength{\tabcolsep}{3.5pt}
\begin{tabular}{lrrrrr}
\toprule
Activity & Comments & \multicolumn{1}{c}{\jFable{Fable~5.2}} & \multicolumn{1}{c}{\jOpus{Opus~5.5}} & \multicolumn{1}{c}{\jArgon{Gemini~4 Argon}} & \multicolumn{1}{c}{\jSol{GPT-6.1 Sol}} \\
\midrule
0 & 47 & 2.1 & 0.0 & 0.0 & 2.1 \\
1--50 & 569 & 12.8 & 7.2 & 4.6 & 1.9 \\
51--200 & 254 & 33.1 & 18.4 & 10.2 & 4.7 \\
201--500 & 223 & 53.6 & 29.0 & 18.8 & 9.0 \\
$>$500 & 165 & 68.9 & 36.0 & 19.4 & 13.3 \\
\bottomrule
\end{tabular}
\end{table}

\begin{table}[!htbp]
\centering\footnotesize
\caption{Story author attribution accuracy (\%) by the total number of Goodreads ratings across the author's books. The 206 stories have 187 unique authors. None means the author has no Goodreads profile. Authors and Stories are the number of each in the bin.}
\label{tab:rq3_visibility_stories}
\setlength{\tabcolsep}{3.5pt}
\begin{tabular}{lrrrrrr}
\toprule
Ratings & Authors & Stories & \multicolumn{1}{c}{\jFable{Fable~5.2}} & \multicolumn{1}{c}{\jOpus{Opus~5.5}} & \multicolumn{1}{c}{\jArgon{Gemini~4 Argon}} & \multicolumn{1}{c}{\jSol{GPT-6.1 Sol}} \\
\midrule
None & 69 & 71 & 4.2 & 0.0 & 0.0 & 0.0 \\
$<$100 & 35 & 39 & 17.9 & 2.6 & 0.0 & 10.3 \\
100--999 & 31 & 34 & 23.5 & 8.8 & 2.9 & 8.8 \\
1k--10k & 28 & 28 & 50.0 & 21.4 & 29.6 & 35.7 \\
10k--100k & 10 & 10 & 100.0 & 80.0 & 80.0 & 80.0 \\
100k--1M & 9 & 9 & 100.0 & 88.9 & 77.8 & 88.9 \\
$\geq$1M & 5 & 15 & 100.0 & 100.0 & 93.3 & 93.3 \\
\bottomrule
\end{tabular}
\end{table}

\subsubsection{Stated confidence}
\label{app:rq3_confidence}
\begin{figure}[!htbp]
\centering
\includegraphics{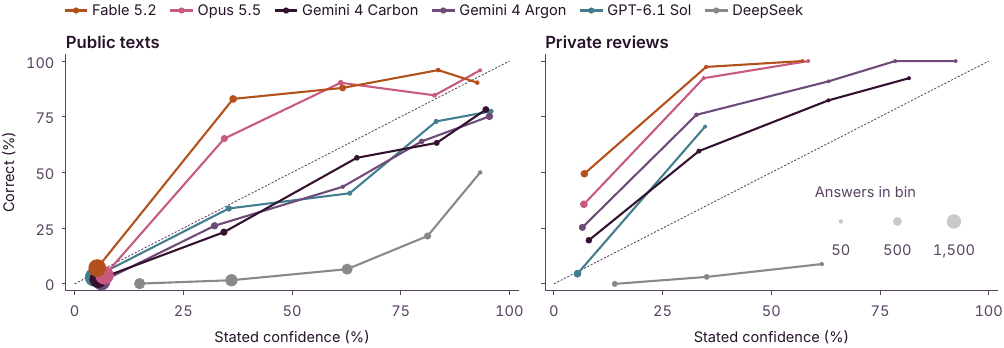}
\caption{The Claude models understate their confidence, and DeepSeek overstates it. Percent of RQ3 answers that name the right person against the mean confidence the model stated, in bins of confidence levels (0--25, 25--50, 50--75, 75--90 and 90--100\%). Public texts pool stories, papers, Reddit comments and all 1,559 public reviews, including Nature-family reviews; the private panel uses all 398 private reviews. The dashed line is perfect calibration; points above it are underconfident.}
\label{fig:rq3_confidence}
\end{figure}

Each model is requested to report a confidence level in their prediction. \autoref{tab:rq3_confidence} relates it to accuracy on pooled public texts: stories, papers, Reddit comments and all 1,559 public reviews, including the 90 Nature-family reviews. ICML 2020 reviews are not included. Confident mistakes are rare for Fable~5.2 and Opus~5.5 (0.3\% and 0.4\% of all answers name the wrong person at a stated confidence of 80\% or more), but more common for Carbon and Argon (3.1\% and 3.3\%) and for DeepSeek~V4.1 Flash (5.1\%), whose mean stated confidence (43\%) far exceeds its accuracy (5\%). \autoref{fig:rq3_confidence} plots accuracy against stated confidence for every detector, on the same pooled public texts and, separately, the private reviews.

\begin{table}[!htbp]
\centering\footnotesize
\caption{Stated confidence and accuracy on pooled public RQ3 texts: stories, papers, Reddit comments and all 1,559 public reviews, including the 90 Nature-family reviews, in \%. This is the public-text cohort in \autoref{fig:rq3_confidence}. Accuracy and percents are over answered texts; refusals and other invalid replies are excluded. The last column counts answers stated at 80\% or more that name the wrong person, as a percent of all answers. ICML 2020 reviews are excluded.}
\label{tab:rq3_confidence}
\setlength{\tabcolsep}{4pt}
\begin{tabular}{lrrrrr}
\toprule
 & & & \multicolumn{2}{c}{Stated 25\% or more} & Stated 80\% or more \\
\cmidrule(lr){4-5}
Detector & Accuracy & Mean confidence & percent & Accuracy & Wrong \\
\midrule
\jFable{Fable~5.2} & 27.0 & 17.5 & 25.0 & 86.6 & 0.3 \\
\jOpus{Opus~5.5} & 16.9 & 14.9 & 18.1 & 76.2 & 0.4 \\
\jCarbon{Gemini~4 Carbon} & 16.7 & 23.0 & 28.7 & 52.8 & 3.1 \\
\jArgon{Gemini~4 Argon} & 14.8 & 22.3 & 26.7 & 52.6 & 3.3 \\
\jSol{GPT-6.1 Sol} & 10.9 & 13.9 & 16.0 & 52.2 & 1.3 \\
\jDeepSeek{DeepSeek~V4.1 Flash} & 5.4 & 42.8 & 72.0 & 7.4 & 5.1 \\
\bottomrule
\end{tabular}
\end{table}

\subsection{Additional experiments}
\label{app:rq3_analysis}

\subsubsection{Self-reported familiarity}
\label{app:rq3-familiarity}
\autoref{tab:rq3_familiarity_ratings} shows attribution accuracy by each model's own familiarity rating across domains.
\begin{table}[H]
\centering\footnotesize
\caption{Human author attribution accuracy (\%, correct/answered) by the model's name-only familiarity rating. Known means a mean rating of at least 1.5 over two calls on a 0--3 scale; a paper takes its best-known listed author.} %
\label{tab:rq3_familiarity_ratings}
\begin{tabular}{l rr r}
\toprule
Domain & Not known & Known & Hits from known (\%) \\
\midrule
\multicolumn{4}{l}{\textbf{\jOpus{Opus~5.5}}} \\
\quad Stories & 0.7 (1/135) & 56.3 (40/71) & 97.6 \\
\quad Papers & 17.6 (3/17) & 37.6 (38/101) & 92.7 \\
\quad Public reviews & 2.5 (22/890) & 29.4 (161/547) & 88.0 \\
\quad Nature-family reviews & 37.5 (6/16) & 61.8 (42/68) & 87.5 \\
\quad Private reviews & N/A (0) & 41.2 (164/398) & 100.0 \\
\quad Reddit & 4.2 (35/840) & 43.6 (173/397) & 83.2 \\
\midrule
\multicolumn{4}{l}{\textbf{\jArgon{Gemini~4 Argon}}} \\
\quad Stories & 0.0 (0/133) & 54.3 (38/70) & 100.0 \\
\quad Papers & 14.3 (3/21) & 35.4 (35/99) & 92.1 \\
\quad Public reviews & 4.4 (47/1,060) & 38.6 (157/407) & 77.0 \\
\quad Nature-family reviews & 29.4 (5/17) & 74.0 (54/73) & 91.5 \\
\quad Private reviews & 6.9 (5/72) & 48.2 (157/326) & 96.9 \\
\quad Reddit & 6.0 (70/1,173) & 65.9 (56/85) & 44.4 \\
\bottomrule
\end{tabular}
\end{table}

\autoref{tab:rq3_familiarity_cross} compares each model's attribution accuracy with the other model's name-only familiarity rating.

\begin{table}[!htbp]
\centering\footnotesize
\caption{Attribution accuracy (\%, correct/answered) grouped by the other model's familiarity with the writer. A writer is known if the mean of two name-only ratings is at least 1.5 on a 0--3 scale.}
\label{tab:rq3_familiarity_cross}
\begin{tabular}{llrr}
\toprule
Attribution model & Rating model & Not known & Known \\
\midrule
\multicolumn{4}{l}{\textbf{Stories}} \\
\quad \jOpus{Opus~5.5} & \jArgon{Gemini~4 Argon} & 0.7 (1/135) & 56.3 (40/71) \\
\quad \jArgon{Gemini~4 Argon} & \jOpus{Opus~5.5} & 0.8 (1/133) & 52.9 (37/70) \\
\midrule
\multicolumn{4}{l}{\textbf{Public reviews}} \\
\quad \jOpus{Opus~5.5} & \jArgon{Gemini~4 Argon} & 4.5 (47/1,049) & 35.1 (136/388) \\
\quad \jArgon{Gemini~4 Argon} & \jOpus{Opus~5.5} & 2.9 (26/900) & 31.4 (178/567) \\

\midrule
\multicolumn{4}{l}{\textbf{Nature-family reviews}} \\
\quad \jOpus{Opus~5.5} & \jArgon{Gemini~4 Argon} & 31.2 (5/16) & 63.2 (43/68) \\
\quad \jArgon{Gemini~4 Argon} & \jOpus{Opus~5.5} & 38.9 (7/18) & 72.2 (52/72) \\
\bottomrule
\end{tabular}
\end{table}

\subsubsection{Refusals and answers that name no one}
\label{app:rq3_refusal_types}
The model could fail to name a specific human author for multiple reasons. Sometimes the model genuinely cannot tell who wrote the text due to lack of clues, or it can explicitly decline to name a writer due to policy or ethical concerns. To separate the two, and to measure how much the prompt and the access route change them, we run Fable~5.2 (high effort) on 500 public peer reviews and 500 Reddit comments with two prompts (our regular and a simple one) through the API and CLI (\autoref{tab:rq3_refusal_types}). The public reviews include all 90 Nature-family reviews, plus a random sample of 410 of the other 1,469. The Reddit comments are all 72 comments that Fable~5.2 rates very familiar and a random sample of 428 of the other 1,186, in proportion to familiarity bin. Note that both samples over-represent writers which are familiar to Fable~5.2.

The \emph{holistic} prompt is the one behind our main results (\S\ref{sec:human_attribution}). The \emph{simple} prompt asks ``Who wrote this text? Name the author as precisely as you can.'' (for Reddit: ``Which Reddit user wrote this comment? Give their exact username.''), followed by a confidence line and an author line. The \emph{Claude Code} route runs the Claude Code CLI in an isolated Docker container. We replace Claude Code's system prompt with ours, disable all tools and remove the account's email from the context. %

We use Opus~5.5 (medium effort, with the same Claude Code setup) to label every response that doesn't name a specific human writer. It categorizes the reasons the response gives (misattribution harm, anonymity, privacy, fabrication, policy, or none), quotes the sentence it relies on, and records whether the response says the text was written or assisted by AI.

\begin{table}[!htbp]
\centering\footnotesize
\caption{Fable~5.2 declines far more often through Claude Code than through the API. \emph{Named}: response names a human author; \emph{Correct}: it names the correct author. A \emph{refusal in answer} declines for a privacy, ethical or policy reason or for no stated reason (e.g., ``I'd rather not guess''); a \emph{can't identify} reply says that any name would be a complete guess or fabrication. \emph{Other or blocked}: number of times the provider rejected the query (API), or Claude Code's safety classifier flagged it.}
\label{tab:rq3_refusal_types}
\setlength{\tabcolsep}{3pt}
\begin{adjustbox}{max width=\linewidth}
\begin{tabular}{lllrrrrrrrrrrr}
\toprule
 & & & \multicolumn{2}{c}{Named} & \multicolumn{7}{c}{Refusal in answer, by reason cited} & & \\
\cmidrule(lr){4-5}\cmidrule(lr){6-12}
Domain & Prompt & Route & All & Correct & All & Misattr. & Anon. & Privacy & Fabr. & Policy & None & \shortstack[r]{Can't\\identify} & \shortstack[r]{Other or\\blocked} \\
\midrule
Public reviews & Simple & Claude Code & 64 & 47 & 205 & 113 & 78 & 38 & 4 & 1 & 32 & 218 & 13 \\
 & & API & 294 & 128 & 22 & 3 & 1 & 2 & 1 & 0 & 16 & 168 & 16 \\
 & Holistic & Claude Code & 184 & 97 & 268 & 221 & 135 & 76 & 49 & 1 & 9 & 32 & 16 \\
 & & API & 442 & 134 & 14 & 10 & 0 & 1 & 3 & 0 & 3 & 28 & 16 \\
\midrule
Reddit & Simple & Claude Code & 34 & 22 & 443 & 422 & 2 & 6 & 34 & 19 & 13 & 23 & 0 \\
 & & API & 296 & 157 & 77 & 68 & 0 & 0 & 35 & 1 & 3 & 126 & 1 \\
 & Holistic & Claude Code & 291 & 145 & 199 & 185 & 19 & 42 & 97 & 9 & 3 & 9 & 1 \\
 & & API & 480 & 181 & 16 & 13 & 2 & 3 & 4 & 2 & 1 & 2 & 2 \\
\bottomrule
\end{tabular}
\end{adjustbox}
\end{table}

\begin{figure}[!htbp]
\centering
\includegraphics[width=\textwidth]{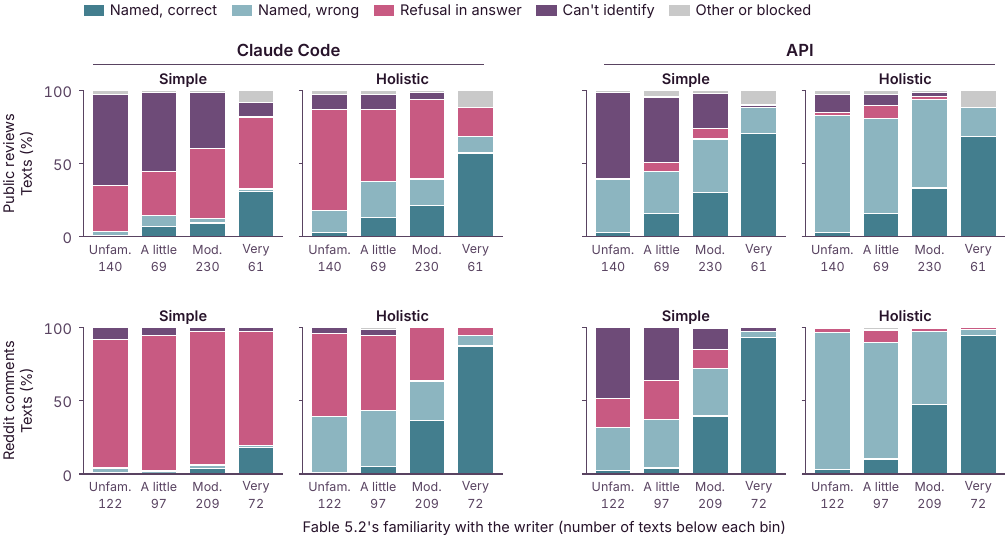}
\caption{We show the refusal types from \autoref{tab:rq3_refusal_types} based on Fable~5.2's familiarity rating of the author.}
\label{fig:rq3_refusal_familiarity}
\end{figure}

\textbf{The route (API vs. CLI) affects refusal rate more than the prompt.} That being said, the simple prompt does result in more refusals especially when run through the CLI, especially on Reddit comments (\autoref{fig:rq3_refusal_familiarity}). Through the API, most replies that name no one say that the model cannot identify the writer and these replies are concentrated on authors with lower familiarity ratings (\S\ref{app:rq3-familiarity}). When the same text and prompt is run through Claude Code, many of the replies become refusals even for familiar writers, and Fable~5.2 has much lower accuracy overall (\autoref{tab:rq3_refusal_types}). 

\textbf{Misattribution harm is the most common reason for refusal.} Refusals on public reviews sometimes mention the reviewer's anonymity and privacy, which are very rarely mentioned for Reddit comments.

\textbf{Replies that call a review AI-written agree with Pangram~4.1.} About half of the public-review replies that name no one say that the review was written or assisted by AI, against about 1\% on Reddit. Pangram~4.1 labels nearly all of these texts AI (93\%), and only 4 of 530 fully Human.

\subsubsection{Topic-only baseline experiments}
\label{app:rq3_topic_only}
It is possible that a model could name an author because it knows who works on a given topic, not because it recognizes how they write (i.e. use content of the text instead of stylistic features). To check this, we give the model only the topic of a text (keywords, or a title and abstract) and ask who works on it or reviewed it. We then compare that with attribution from the full text. We find that providing the full text gives higher accuracy on both papers and public reviews (\autoref{tab:rq3_topic_only} and \autoref{tab:rq3_topic_only_reviews}).

\begin{table}[!htbp]
\centering\small
\caption{We test a topic-only baseline for paper author attribution. We show the percentage of 34 recent cs.CL papers (labelled human by Pangram~4.1) for which Opus~5.5 names a co-author. We compare with our regular full main text version which asks who wrote the paper. The topic-only prompts ask for one early-career and one senior researcher who work on the problem; the full-text prompt asks for the most likely junior and senior author. A name counts if the person is any author of the paper, whatever role Opus~5.5 gives them.}
\label{tab:rq3_topic_only}
\begin{tabular}{llrr}
\toprule
Input & Question & Senior researcher (\%) & Either person (\%) \\
\midrule
Five topic keywords & Who works on this? & 9 & 12 \\
Title and abstract & Who works on this? & 41 & 53 \\
Full main text & Who wrote this? & 65 & 76 \\
\bottomrule
\end{tabular}
\end{table}

\begin{table}[!htbp]
\centering\small
\caption{We test a topic-only baseline for public reviews. Opus~5.5 gives one name per reviewed paper, so from the title and abstract at most 87.3\% of reviews can be named. Reviews are the 789 public reviews of \autoref{fig:rq3_domains} that Pangram~4.1 does not label AI.} %
\label{tab:rq3_topic_only_reviews}
\begin{tabular}{llr}
\toprule
Input & Question & Reviewer named (\%) \\
\midrule
Title and abstract of the reviewed paper & {Who reviewed this?} & {2.0} \\
{Full review} & {Who wrote this?} & {28.4} \\
\bottomrule
\end{tabular}
\end{table}

\subsubsection{Stability across repeated judgments and paraphrasing}
\label{app:rq3_stability}
\paragraph{Overall accuracy over repeated judgments varies little, though individual predictions can change.} In five independent Fable~5.2 judgments of Reddit comments, overall accuracy ranges from 29.9--31.1\% (\autoref{tab:rq3_stability}, top). Only 32.6\% of comments receive the same normalized answer in all five runs, whereas 25.0\% are correctly attributed in every run and 63.2\% are incorrect in every run. The remaining 11.8\% are correct in one to four runs. Earlier same-text repeats on a 200-document, five-domain sample also show relatively small accuracy ranges for Argon and Opus~5.5 (\autoref{tab:rq3_stability}, bottom).

\begin{table}[!htbp]
\centering\footnotesize
\setlength{\tabcolsep}{5pt}
\caption{Attribution stability on unchanged texts. Top: consistency among the 1,240 Reddit comments with five valid answers.} 
\label{tab:rq3_stability}
\begin{tabular}{lrrrr}
\toprule
\multicolumn{5}{l}{\textbf{Fable~5.2: five-run consistency ($n=1{,}240$)}} \\
Pattern & \multicolumn{2}{r}{Comments} & \multicolumn{2}{r}{Share} \\
\midrule
Same answer in all five & \multicolumn{2}{r}{404} & \multicolumn{2}{r}{32.6} \\
Correct in all five & \multicolumn{2}{r}{310} & \multicolumn{2}{r}{25.0} \\
Incorrect in all five & \multicolumn{2}{r}{784} & \multicolumn{2}{r}{63.2} \\
Correct in one to four & \multicolumn{2}{r}{146} & \multicolumn{2}{r}{11.8} \\
\midrule
\multicolumn{5}{l}{\textbf{200 Document Sample (all domains), three runs}} \\
Model & Documents & Original & Repeat 2 & Repeat 3 \\
\midrule
Gemini~4 Argon & 194 & 20.6 & 21.6 & 21.1 \\
Opus~5.5 & 176 & 21.6 & 23.3 & 25.6 \\
\bottomrule
\end{tabular}
\end{table}

\begin{table}[!tb]
\centering\footnotesize
\caption{Attribution accuracy before and after rewriting on the Figure~\ref{fig:rq3_domains} dataset. Columns show the accuracy for the original and two independent Argon rewrites, judged by Argon or Opus~5.5 with the same holistic prompt. Each row uses the same $n$ texts with valid judgments of all three versions; original accuracies can differ from Figure~\ref{fig:rq3_domains}, which uses all valid originals. Reddit requires the exact username, papers accept any listed author, reviews and stories use the shared name matching rule. ICML 2020 reviews have no rewrite judgments and are excluded. \emph{All texts} includes every text regardless of its Pangram~4.1 label; \emph{Not AI} keeps the texts that Pangram~4.1 labels human or mixed, as in \autoref{tab:rq3_full}.}
\label{tab:rq3-rewrite-accuracy}
\setlength{\tabcolsep}{3pt}
\begin{tabular}{lrrrr@{\hspace{10pt}}rrrr}
\toprule
 & \multicolumn{4}{c}{All texts} & \multicolumn{4}{c}{Not AI} \\
\cmidrule(lr){2-5}\cmidrule(lr){6-9}
Detector & \textit{n} & Original & Rewrite 1 & Rewrite 2 & \textit{n} & Original & Rewrite 1 & Rewrite 2 \\
\midrule
\multicolumn{9}{l}{\textbf{Private reviews}} \\
\quad \jArgon{Gemini~4 Argon} & 398 & 40.7 & 14.8 & 15.1 & 396 & 40.9 & 14.9 & 15.2 \\
\quad \jOpus{Opus~5.5} & 398 & 41.2 & 16.1 & 15.1 & 396 & 41.4 & 16.2 & 15.2 \\
\addlinespace
\multicolumn{9}{l}{\textbf{Reddit}} \\
\quad \jArgon{Gemini~4 Argon} & 1,253 & 10.0 & 7.0 & 6.7 & 1,225 & 10.2 & 7.2 & 6.9 \\
\quad \jOpus{Opus~5.5} & 1,236 & 16.8 & 10.3 & 10.5 & 1,208 & 17.2 & 10.5 & 10.8 \\
\addlinespace
\multicolumn{9}{l}{\textbf{Public reviews}} \\
\quad \jArgon{Gemini~4 Argon} & 702 & 23.5 & 15.1 & 15.4 & 395 & 38.7 & 25.1 & 25.8 \\
\quad \jOpus{Opus~5.5} & 672 & 20.8 & 15.0 & 14.6 & 371 & 35.8 & 25.3 & 24.0 \\
\addlinespace
\multicolumn{9}{l}{\textbf{Stories}} \\
\quad \jArgon{Gemini~4 Argon} & 199 & 18.6 & 15.6 & 16.1 & 184 & 20.1 & 16.8 & 17.4 \\
\quad \jOpus{Opus~5.5} & 199 & 20.1 & 16.1 & 16.6 & 184 & 21.7 & 17.4 & 17.9 \\
\addlinespace
\multicolumn{9}{l}{\textbf{Papers}} \\
\quad \jArgon{Gemini~4 Argon} & 120 & 31.7 & 23.3 & 26.7 & 90 & 40.0 & 28.9 & 32.2 \\
\quad \jOpus{Opus~5.5} & 118 & 34.7 & 31.4 & 31.4 & 88 & 43.2 & 39.8 & 40.9 \\
\bottomrule
\end{tabular}
\par\smallskip
\end{table}

\subsubsection{Topics in reviewer-attribution explanations}
\label{app:rq3-topics}
\autoref{tab:deniz_agent_rq3_topic_desc} gives TopicGPT's definition of topics referenced in \autoref{fig:rq3-topics}. We generate separate topic lists for ICML 2020 reviews and Reddit comments, so the table has one block for each. We use GPT-6.1~Sol to process 5,000 ICML reviews and 1,200 Reddit comments to propose 154 and 97 potential topics respectively. GPT-6.1~Sol is then asked to refine the topic lists by removing rare topics and merging similar ones. Our final list contains 77 topics for ICML explanations and 45 for Reddit. Each explanation is assigned 9--10 distinct topics on average. The topics label what a model wrote in its explanation, not the text itself. Evidence topics describe what the model noticed in the text, and inference topics describe what it concluded about the writer. The most frequent topics in the explanations include ``Peer-review format" (the text is identified as a peer review) and ``Writing style'' on ICML, and ``Tone and stance'' and ``Subject focus'' on Reddit.

\begin{table*}[t]
\centering\footnotesize
\setlength{\tabcolsep}{4pt}
\renewcommand{\arraystretch}{1.12}
\caption{Definitions, as written by TopicGPT, of the topics in \autoref{fig:rq3-topics}, of the structure topics it groups in one row, and of the topics named in \S\ref{sec:ha_analysis}. ICML 2020 reviews and Reddit comments have separate codebooks, generated independently, so a topic in one block does not correspond to a similarly named topic in the other.}
\label{tab:deniz_agent_rq3_topic_desc}
\begin{tabularx}{\textwidth}{@{}>{\raggedright\arraybackslash}p{0.27\textwidth} >{\raggedright\arraybackslash}p{0.075\textwidth} >{\raggedright\arraybackslash}X@{}}
\toprule
Topic & Type & TopicGPT description \\
\midrule
\multicolumn{3}{@{}l}{\textbf{ICML 2020 reviews (77-topic codebook)}} \\
\addlinespace[3pt]
\multicolumn{3}{@{}l}{\textit{Shown in \autoref{fig:rq3-topics}}} \\
Technical understanding and reasoning & Evidence & The reviewer demonstrates understanding of relevant concepts, technical distinctions, or precise reasoning. \\
Prior-work references & Evidence & The reviewer cites specific prior work or distinguishes its attribution and historical relationship to other work. \\
Claimed publication experience & Evidence & The author explicitly claims publication experience in the area under review. \\
Linguistic background & Inference & Language patterns are explicitly linked to the author's native language or language background. \\
Researcher role & Inference & The explanation explicitly identifies the author as a researcher in a relevant area acting as a reviewer. \\
Anonymous reviewer role & Inference & The explanation explicitly identifies the author as an anonymous reviewer for a publication venue. \\
\addlinespace[3pt]
\multicolumn{3}{@{}l}{\textit{Structure topics (one row in \autoref{fig:rq3-topics})}} \\
Peer-review format & Evidence & The text is identified as a peer review through its structure, context, or explicit description. \\
Review-platform provenance & Inference & The explanation explicitly attributes elements of the review to a reviewing platform or its template. \\
Standardized review language & Evidence & The explanation identifies wording supplied by a standard review form rather than distinctive author-written prose. \\
Review ratings & Evidence & The explanation identifies numeric scores or rating fields in the review as an observed clue. \\
\addlinespace[3pt]
\multicolumn{3}{@{}l}{\textit{Named in \S\ref{sec:ha_analysis}}} \\
Writing style & Evidence & The explanation describes the prose's concision, directness, tone, punctuation, or organization. \\
Language mechanics errors & Evidence & The text contains grammatical irregularities or spelling errors without necessarily attributing them to a language background. \\
Non-native language patterns & Evidence & The text contains spelling, grammatical, or syntactic patterns described as characteristic of non-native English. \\
\midrule
\multicolumn{3}{@{}l}{\textbf{Reddit comments (45-topic codebook)}} \\
\addlinespace[3pt]
Tone and stance & Evidence & The explanation identifies the writer's tone, attitude, or evaluative posture. \\
Subject focus & Evidence & The explanation identifies the subject matter around which the discussion centers. \\
Punctuation conventions & Evidence & The writing exhibits an identified punctuation practice. \\
Text formatting & Evidence & The writing uses identifiable visual formatting or markup conventions. \\
Sentence structure & Evidence & The explanation identifies characteristic sentence lengths or syntactic constructions. \\
Capitalization conventions & Evidence & The writing exhibits an identified practice of capitalizing words or terms. \\
Language errors & Evidence & The writing contains grammatical slips, typos, run-on sentences, or other unpolished language. \\
\bottomrule
\end{tabularx}
\end{table*}

\subsubsection{Factuality of claims in model explanations from ICML 2020 reviews}
\label{app:rq3_grounding}
We check whether the external facts that judges state in their explanations are true, following the claim-extraction approach of VeriScore~\citep{song2024veriscore}. We use the explanations of Opus~5.5, Argon, Sol and DeepSeek for the same 5,000 ICML 2020 reviews as the topic analysis (\S\ref{app:rq3-topics}); Fable~5.2's explanations were generated after this check and are not included. Gemini~4 Argon extracts 8,448 candidate external claims from the 107,645 observations in these 20,000 explanations, 0.42 per explanation (Opus~5.5 3,073, Argon 2,596, Sol 561, DeepSeek 2,218). Most observations describe the review itself or speculate about the reviewer, so few contain a claim that can be checked outside the review. Argon then checks 5,477 of these claims, from 2,000 reviews, with Google Search grounding and labels each one supported, contradicted, insufficient evidence or not assessable. It labels 5,192 claims supported, 161 contradicted, 61 insufficient evidence and 63 not assessable. 95.9\% of the 5,414 assessable claims (94.8\% of all checked claims) are supported.

\section{Cost estimate}
\label{app:cost}
\begin{table}[H]
\centering\footnotesize
\caption{Estimated cost (USD) of the experiments reported in this paper. \emph{API}: API charges. \emph{CLI}: runs through Claude Code or Codex subscriptions, priced at API list rates for the tokens used.}
\label{tab:project_costs}
\begin{tabular}{@{}lrrr@{}}
\toprule
 & API & CLI & Total \\
\midrule
\S\ref{sec:ai_vs_human}: human or AI & 4,983 & 5,649 & 10,632 \\
\S\ref{sec:model_detection}: which model & 169 & 831 & 1,000 \\
\S\ref{sec:human_attribution}: which human & 2,199 & 510 & 2,709 \\
\midrule
Total & 7,351 & 6,990 & 14,341 \\
\bottomrule
\end{tabular}
\end{table}

\textbf{Limitations of the cost estimate.}
These are estimates. They leave out runs with no saved usage, the feature pipeline of \S\ref{app:feat-analysis}, the Reddit familiarity, imitation and Fable~5.2 and Argon reasoning-effort runs. Costs of our joint-prompt variant judgments are not counted.

\end{document}